\documentclass[10pt,journal,compsoc]{IEEEtran}

\ifCLASSOPTIONcompsoc
  \usepackage[nocompress]{cite}
\else
  \usepackage{cite}
\fi

\ifCLASSINFOpdf

\else

\fi

\usepackage{epsfig}
\usepackage{graphicx}
\usepackage{amsmath}
\usepackage{amssymb}
\usepackage{comment}
\usepackage{multirow}

\usepackage{booktabs}
\usepackage{array, caption, threeparttable}
\usepackage{caption}

\usepackage{algorithm}
\usepackage{listings}
\usepackage{color}

\usepackage{mathtools}

\usepackage{todonotes}
\usepackage{gensymb}

\usepackage{hyperref}

\begin{document}

\title{Deep Learning for Face Anti-Spoofing: A Survey}

\author{Zitong Yu,~\IEEEmembership{Member,~IEEE}, Yunxiao Qin, Xiaobai Li,~\IEEEmembership{Member,~IEEE}, Chenxu Zhao, \\
Zhen Lei,~\IEEEmembership{Senior Member,~IEEE} and Guoying Zhao,~\IEEEmembership{Fellow,~IEEE}


\IEEEcompsocitemizethanks{\IEEEcompsocthanksitem Z. Yu, X. Li and G. Zhao are with Center for Machine Vision and Signal Analysis, University of Oulu, Oulu 90014, Finland. 

E-mail: \{zitong.yu, xiaobai.li, guoying.zhao\}@oulu.fi

\IEEEcompsocthanksitem Y. Qin is with Communication University of China, Beijing 100024, China.
E-mail: qinyunxiao@cuc.edu.cn

\IEEEcompsocthanksitem C. Zhao is with SailYond Technology, Beijing 100000, China.

E-mail: zhaochenxu@sailyond.com

\IEEEcompsocthanksitem Z. Lei is with the National Laboratory of Pattern Recognition (NLPR), Center for Biometrics and Security Research (CBSR), Institute of Automation, Chinese Academy of Sciences (CASIA), Beijing 100190, China, also with the School of Artificial Intelligence, University of Chinese Academy of Sciences (UCAS), Beijing 100049, China, and also with the Centre for Artificial Intelligence and Robotics, Hong Kong Institute of Science \& Innovation, Chinese Academy of Sciences, Hong Kong, SAR.

E-mail: zlei@nlpr.ia.ac.cn



}
\thanks{Manuscript received June 26, 2021; revised May 15 and September 2, 2022. (Corresponding author: Guoying Zhao)}}

\markboth{IEEE Transactions on on Pattern Analysis and Machine Intelligence}%
{Shell \MakeLowercase{\textit{et al.}}: Bare Advanced Demo of IEEEtran.cls for IEEE Computer Society Journals}

\IEEEtitleabstractindextext{%
\begin{abstract}

Face anti-spoofing (FAS) has lately attracted increasing attention due to its vital role in securing face recognition systems from presentation attacks (PAs). As more and more realistic PAs with novel types spring up, early-stage FAS methods based on handcrafted features become unreliable due to their limited representation capacity. With the emergence of large-scale academic datasets in the recent decade, deep learning based FAS achieves remarkable performance and dominates this area. However, existing reviews in this field mainly focus on the handcrafted features, which are outdated and uninspiring for the progress of FAS community. In this paper, to stimulate future research, we present the first comprehensive review of recent advances in deep learning based FAS. It covers several novel and insightful components: 1) besides supervision with binary label (e.g., ‘0’ for bonafide vs. ‘1’ for PAs), we also investigate recent methods with pixel-wise supervision (e.g., pseudo depth map); 2) in addition to traditional intra-dataset evaluation, we collect and analyze the latest methods specially designed for domain generalization and open-set FAS; and 3) besides commercial RGB camera, we summarize the deep learning applications under multi-modal (e.g., depth and infrared) or specialized (e.g., light field and flash) sensors. We conclude this survey by emphasizing current open issues and highlighting potential prospects. 



\end{abstract}

}

\maketitle

\IEEEdisplaynontitleabstractindextext

%
\IEEEpeerreviewmaketitle

\ifCLASSOPTIONcompsoc
\IEEEraisesectionheading{\section{Introduction}\label{sec:introduction}}
\else
\section{Introduction}
\label{sec:introduction}
\fi

\begin{figure}
\centering
\includegraphics[scale=0.5]{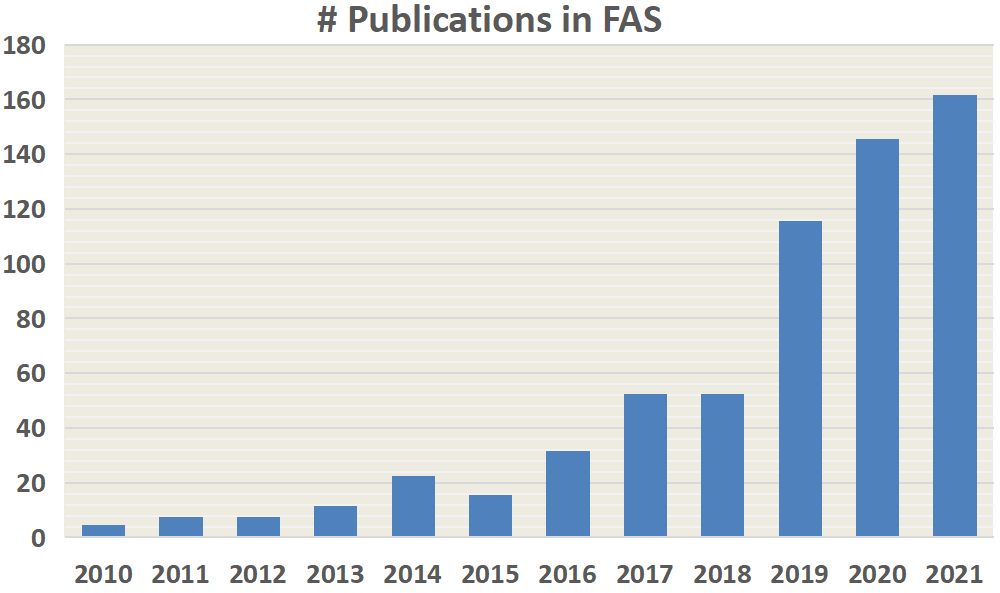}
\vspace{-1.4em}
  \caption{ 
  The increasing research interest in the FAS field, obtained through Google scholar search with key-words: allintitle: “face anti-spoofing”, “face presentation attack detection”, and “face liveness detection”.}
\label{fig:Figure1}
\vspace{-1.2em}
\end{figure}

\IEEEPARstart{D}{ue} to its convenience and remarkable accuracy, face recognition technology~\cite{guo2020learning} has been applied in a few interactive intelligent applications such as checking-in and mobile payment. However, existing face recognition systems are vulnerable to presentation attacks (PAs) ranging from print, replay, makeup, 3D-mask, etc. Therefore, both academia and industry have paid extensive attention to developing face anti-spoofing (FAS) technology for securing the face recognition system. As illustrated in Fig.~\ref{fig:Figure1}, FAS (namely `face presentation attack detection' or `face liveness detection') is an active research topic in computer vision and has received an increasing number of publications in recent years.

In the early stage, plenty of traditional handcrafted feature~\cite{pan2007eye,li2016generalized,Pereira2012LBP,Komulainen2014Context,Patel2016Secure} based methods have been proposed for presentation attack detection (PAD). Most traditional algorithms are designed based on human liveness cues and handcrafted features, which need rich task-aware prior knowledge for design. In term of the methods based on the liveness cues, eye-blinking~\cite{pan2007eye,jee2006liveness,li2008eye}, face and head movement~\cite{wang2009face,bao2009liveness} (e.g., nodding and smiling), gaze tracking~\cite{bigun2004assuring,ali2012liveness} and remote physiological signals (e.g., rPPG~\cite{li2016generalized,Liu2018Learning,lin2019face,yu2019remote1}) are explored for dynamic discrimination. However, these physiological liveness cues are usually captured from long-term interactive face videos, which is inconvenient for practical deployment. Furthermore, the liveness cues are easily mimicked by video attacks, making them less reliable. On the other hand, classical handcrafted descriptors (e.g., LBP~\cite{boulkenafet2015face,Pereira2012LBP}, 
SIFT~\cite{Patel2016Secure}, SURF~\cite{boulkenafet2016face2}, HOG~\cite{Komulainen2014Context} and DoG~\cite{tan2010face}) are designed for extracting effective spoofing patterns from various color spaces (RGB, HSV, and YCbCr). It can be seen from Table-A 1 (in Appendix) that the FAS surveys before 2018 mainly focus on this category. 


\begin{figure*}[t]
\centering
\includegraphics[scale=0.39]{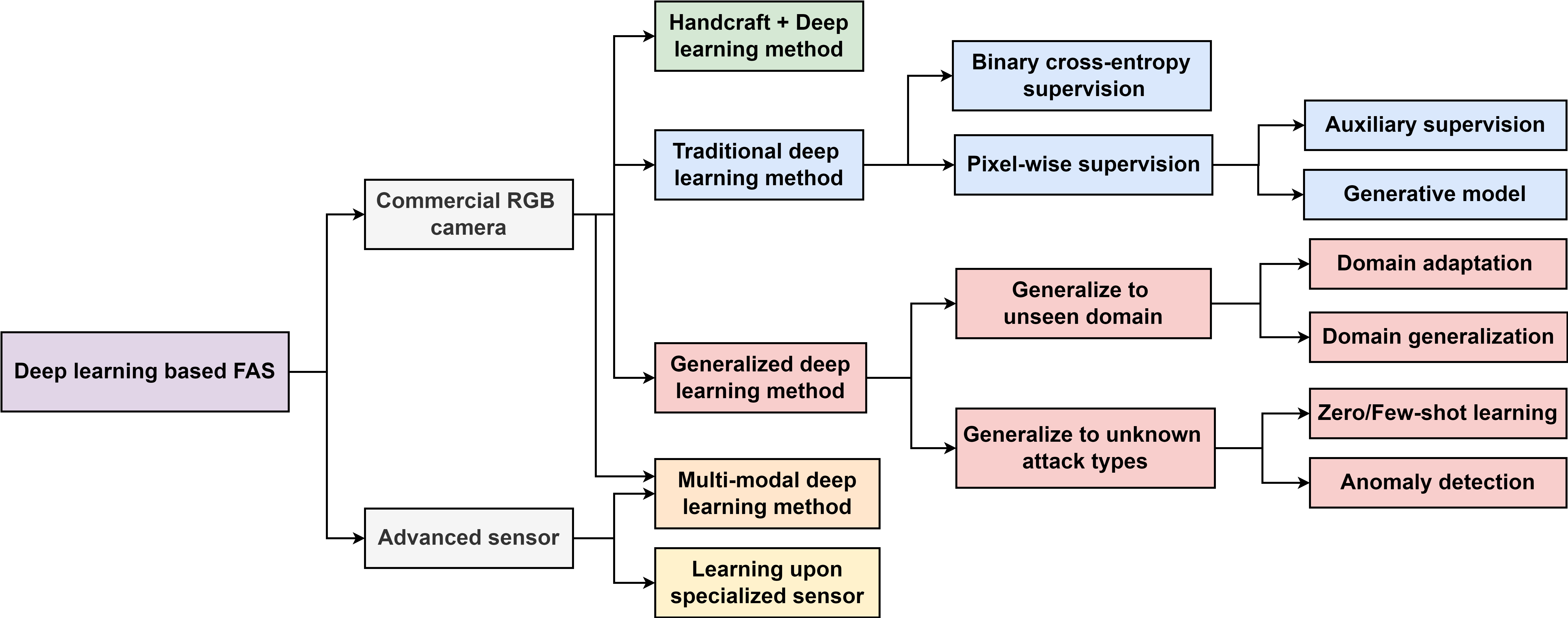}
  \caption{ 
  Topology of the deep learning based FAS methods.}
\label{fig:topology}
\vspace{-1.0em}
\end{figure*}

\newcommand{\tabincell}[2]{\begin{tabular}{@{}#1@{}}#2\end{tabular}}

Subsequently, a few hybrid (handcrafted+deep learning)~\cite{song2019discriminative,asim2017cnn,rehman2020enhancing,khammari2019robust} and end-to-end deep learning based methods~\cite{yu2020searching,yu2020face,Liu2018Learning,yang2019face,Atoum2018Face,yu2020multi,zhang2020casia} are proposed for both static and dynamic face PAD. Most works\cite{yang2014learn,Li2017An,Patel2016Cross,george2019deep,jourabloo2018face,jia20203d,li2020compactnet} treat FAS as a binary classification problem (e.g., `0' for live while `1' for spoofing faces, or vice versa) thus supervised by a simple binary cross-entropy loss. Different from other binary vision tasks, the FAS is a self-evolving problem (i.e., attack vs. defense develop iteratively), which makes it more challenging. Furthermore, other binary vision tasks (e.g., human gender classification) highly rely on the obvious appearance-based semantic clues (e.g., hair style, wearing, facial shape) while the intrinsic features (e.g., material and geometry) in FAS are usually content-irrelevant (e.g., not related to facial attribute and ID), subtle and with fine-grained details, which are very challenging to distinguish by even human eyes. Thus, convolutional neural networks (CNNs) with single binary loss might reasonably mine different kinds of semantic features for binary vision tasks like gender classification but discover arbitrary and unfaithful clues (e.g., screen bezel) for spoofing patterns. Fortunately, such intrinsic live/spoof clues are usually closely related with some position-aware auxiliary tasks. For instance, the face surface of print/replay and transparent mask attacks are usually with irregular/limited geometric depth distribution and abnormal reflection, respectively. Based on these physical evidences, recently, pixel-wise supervision~\cite{Atoum2018Face,Liu2018Learning,yu2020face,kim2019basn,george2019deep,yu2020fas2} attracts more attention as it provides more fine-grained context-aware supervision signals, which is beneficial for deep models learning intrinsic spoofing cues. On one hand, pseudo depth labels~\cite{Atoum2018Face,Liu2018Learning}, reflection maps~\cite{yu2020face,kim2019basn}, binary mask label~\cite{george2019deep,liu2019deep,sun2020face} and 3D point cloud maps~\cite{li3dpc} are typical pixel-wise auxiliary supervisions, which describe the local live/spoof cues in pixel/patch level. On the other hand, besides physical-guided auxiliary signals, a few generative deep FAS methods model the intrinsic spoofing patterns via relaxed pixel-wise reconstruction constraints~\cite{jourabloo2018face,feng2020learning,liu2020physics,qin2021meta}. As shown in Table-A 1 (in Appendix), the latest FAS surveys from 2018 to 2020 investigate limited numbers ($\textless$50) of deep learning based methods, which hardly provide comprehensive elaborations for the community researchers. Note that most data-driven methods introduced in previous surveys are supervised by traditional binary loss, and there is still a blank for summarizing the arisen pixel-wise supervision methods.

Meanwhile, the emergence of large-scale public FAS datasets with rich attack types and recorded sensors also greatly boosts the research community. First, the datasets with vast samples and subjects have been released. For instance, CelebA-Spoof~\cite{zhang2020celeba}, recorded from 10177 subjects, contains 156384 and 469153 face images for bonafide and PAs, respectively. Second, besides the common PA types (e.g., print and replay attacks), some up-to-date datasets contain richer challenging PA types (e.g., SiW-M~\cite{liu2019deep} and WMCA~\cite{george2019biometric} with more than 10 PA types). However, we can find from Table-A 1 (in Appendix) that existing surveys only investigate a handful of ($\textless$15) old and small-scale FAS datasets, which cannot provide fair benchmarks for deep learning based methods. Third, in terms of modality and hardware for recording, besides commercial visible RGB camera, numerous multimodal and specialized sensors benefit the FAS task. For example, CASIA-SURF~\cite{zhang2020casia} and WMCA~\cite{george2019biometric} show the effectiveness of PAD via fusing RGB/depth/NIR information while dedicated systems with multispectral SWIR~\cite{steiner2016reliable} and four-directional polarized~\cite{tian2020face} cameras significantly benefit for spoofing material perception. However, previous surveys mostly focus on single RGB modality using a commercial visible camera, and neglect the deep learning applications on the multimodal and specialized systems for high-security scenarios.

From the perspective of evaluation protocols, traditional `intra-dataset intra-type' and `cross-dataset intra-type' protocols are widely investigated in previous FAS surveys (see Table-A 1 in Appendix). As FAS is actually an open-set problem in practice, the uncertain gaps (e.g., environments and attack types) between training and testing conditions should be considered. However, no existing reviews consider the issues about unseen domain generalization~\cite{shao2019multi,shao2019regularized,wang2020cross,jia2020single} and unknown PAD~\cite{arashloo2017anomaly,liu2019deep,qin2019learning,qin2020one}. Most reviewed FAS methods design or train the FAS model on predefined scenarios and PAs. Thus, the trained models easily overfit on several specific domains and attack types, and are vulnerable to unseen domains and unknown attacks. To bridge the gaps between academic research and real-world applications, in this paper, we fully investigate deep learning based methods under four FAS protocols, including challenging domain generalization and open-set PAD situations. 
Compared with existing literatures, the major contributions of this work can be summarized as follows:

\begin{figure*}
\centering
\includegraphics[scale=0.4]{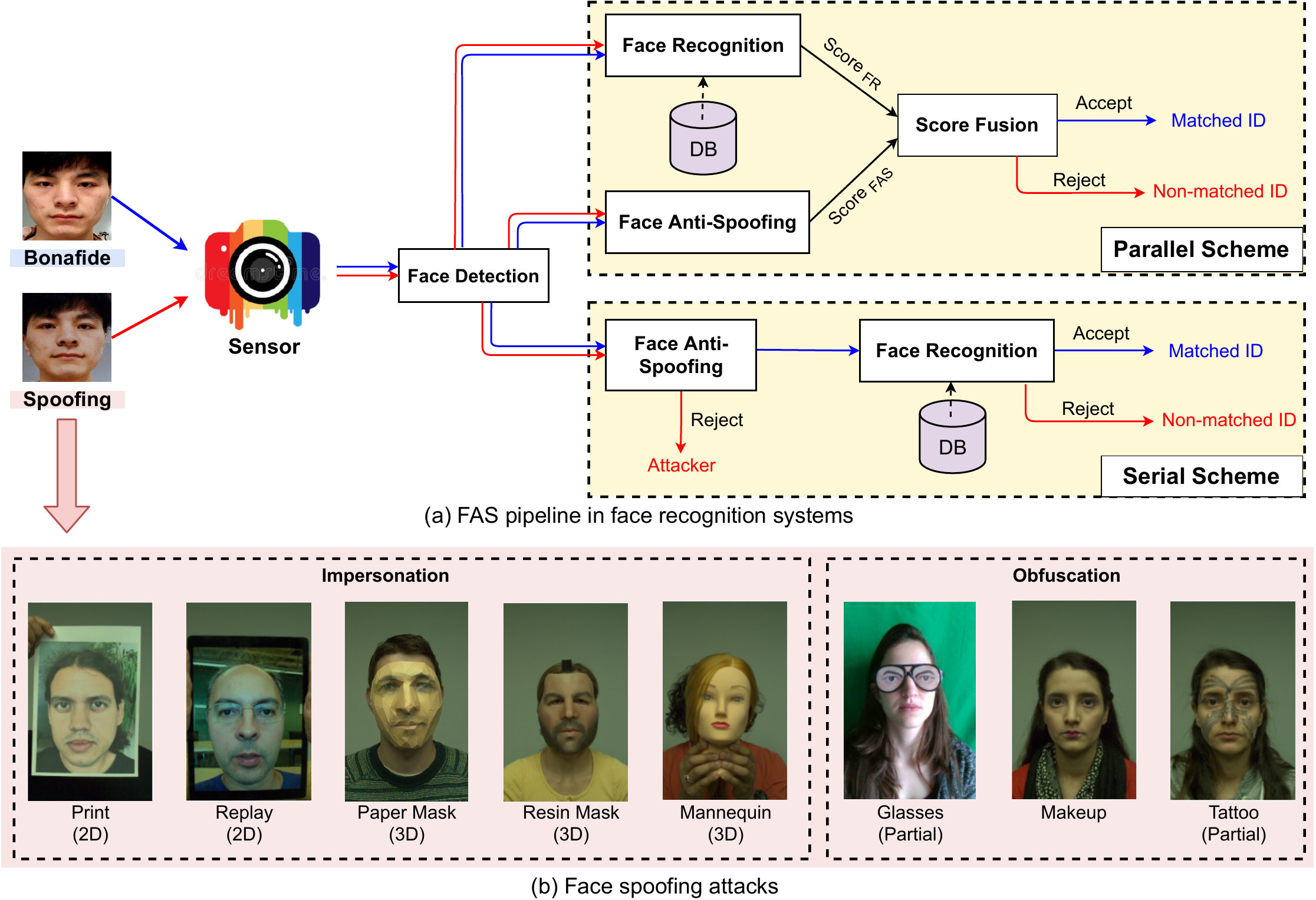}
\vspace{-0.3em}
  \caption{ 
  Typical face spoofing attacks and face anti-spoofing pipeline. (a) FAS could be integrated with face recognition systems with paralled or serial scheme for reliable face ID matching. (b) Visualization of several classical face spoofing attack types~\cite{heusch2020deep} in terms of impersonation/obfuscation, 2D/3D, and whole/partial evidences.
  }
  \vspace{-0.8em}
\label{fig:pipeline}
\end{figure*}

\begin{itemize}
    
    \item To the best of our knowledge, this is the first survey paper to comprehensively cover ($\textgreater$100) deep learning methods for both single- and multi-modal FAS with generalized protocols. Compared with previous surveys only considering the methods with binary loss supervision, we also elaborate on those with auxiliary/generative pixel-wise supervision. 
    
    \item  As opposed to existing reviews~\cite{pereira2020rise,jia2020survey,el2020deep} with only limited numbers ($\textless$15) of small-scale datasts, we show detailed comparisons among past-to-present 35 public datasets including various kinds of PAs as well as advanced recording sensors.
    
   \item  This paper covers the most recent and advanced progress of deep learning on four practical FAS protocols (i.e., intra-dataset intra-type, cross-dataset intra-type, intra-dataset cross-type, and cross-dataset cross-type testings). Therefore, it provides the readers with state-of-the-art methods with different application scenarios (e.g., unseen domain generalization and unknown attack detection).
   
   \item  Comprehensive comparisons of existing deep FAS methods with insightful taxonomy are provided in Tables-A 5, 6, 7, 8, 9, 10, and 11 (in Appendix), with brief summaries and discussions being presented.
 
\end{itemize}

We summarize the topology of deep learning based FAS methods with the commercial monocular RGB camera and advanced sensors in Fig.~\ref{fig:topology}. On one hand, as commercial RGB camera is widely used in many real-world applicational scenarios (e.g., access control system and mobile device unlocking), there are richer research works based on this branch. It includes three main categories: 1) hybrid learning methods combining both handcrafted and deep learning features; 2) traditional end-to-end supervised deep learning based methods; and 3) generalized deep learning methods to both unseen domain and unknown attack types. 
Besides the commercial RGB camera, researchers have also developed sensor-aware deep learning methods for efficient FAS using specialized sensors/hardwares.
Meanwhile, as multi-spectrum imaging systems with acceptable costs are increasingly used in real-world applications, multi-modal deep learning based methods become hot and active in the FAS research community.

The structure of this paper is as follows. Section~\ref{sec:background} introduces the research background, including presentation attacks, datasets, evaluation metrics, and protocols for the FAS task. Section~\ref{sec:RGB} reviews the methods for visible RGB based FAS according to two kinds of supervision signals (i.e., binary loss and pixel-wise loss) as well as generalized learning for unseen domains and unknown attacks. Section~\ref{sec:multimodal} gives a comparison about the recording sensors as well as modalities, and then presents the methods for specific recorded inputs. Section~\ref{sec:discussion} discusses the current issues of deep FAS, and indicates the future directions. Finally, conclusions are given in Section~\ref{sec:conclusion}. Researchers can track an up-to-date list at \href{https://github.com/ZitongYu/DeepFAS}{https://github.com/ZitongYu/DeepFAS}.

\begin{figure*}
\centering
\includegraphics[scale=0.33]{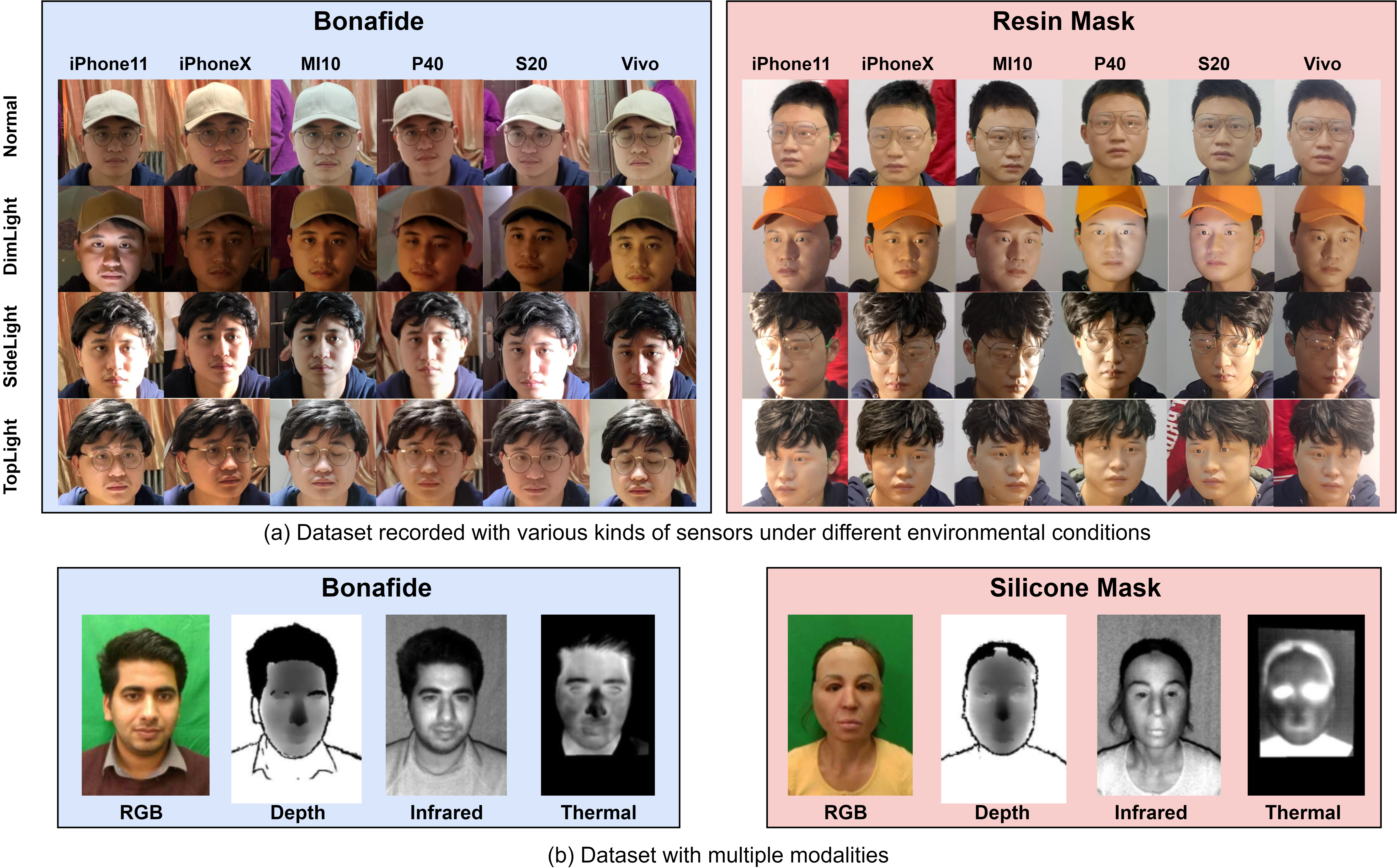}
\vspace{-0.3em}
  \caption{ 
  Visualization of the bonafide and spoofing samples from the HiFiMask dataset~\cite{liu2021contrastive} (a) with various cameras under different lighting conditions~\cite{liu2021contrastive}; and the WMCA dataset~\cite{george2019biometric} (b) with multiple modalities such as visible RGB, depth, infrared, and thermal.
  }
\label{fig:dataset}
\vspace{-0.8em}
\end{figure*}

\vspace{-1.0em}
\section{background} \label{sec:background}

In this section, we will introduce the common face spoofing attacks first, and then investigate the existing FAS datasets as well as their evaluation metrics and protocols.

\vspace{-1.0em}
\subsection{Face Spoofing Attacks}
Attacks on automatic face recognition (AFR) system usually divide into two categories: digital manipulation~\cite{tolosana2020deepfakes,goswami2019detecting} and physical presentation attacks~\cite{liu2021cross}. The former one fools the face system via imperceptibly visual manipulation in the digital virtual domain, while the latter usually misleads the real-world AFR systems via presenting face upon physical mediums in front of the imaging sensors. In this paper, we focus on the detection of physical face presentation attacks, whose pipeline is illustrated in Fig.~\ref{fig:pipeline}(a). It can be seen that there are two kinds of schemes~\cite{hernandez2019introduction} for integrating FAS with AFR systems: 1) \textit{parallel} fusion~\cite{de2013can} with the predicted scores from FAS and AFR systems. The combined new final score is used to determine if the sample comes from a genuine user or not; and 2) \textit{serial} scheme~\cite{li2018face} for early face PAs detection and spoof rejection, thus avoiding the spoof face accessing the subsequent face recognition phase.

In Fig.~\ref{fig:pipeline}(b), some representative spoofing attack types are illustrated. According to the attackers' intention, face PAs~\cite{marcel2019handbook} can be divided into two typical cases: 1)  \textit{impersonation}, which entails the use of spoof to be recognized as someone else via copying a genuine user’s facial attributes to special mediums such as photo, electronic screen, and 3D mask; and 2) \textit{obfuscation}, which entails
the use to hide or remove the attacker’s own identity using various methods such as glasses, makeup, wig, and disguised face.

Based on the geometry property, PAs are broadly classified into \textit{2D} and \textit{3D} attacks. 2D PAs~\cite{ramachandra2017presentation} are carried out by presenting facial attributes using photo or video to the sensor. Flat/wrapped printed photos, eye/mouth-cut photos, and digital replay of videos are common 2D attack variants. With the maturity of 3D printing technology, face 3D mask~\cite{jia2020survey} has become a new type of PA to threaten AFR systems. Compared with traditional 2D PAs, face masks are more realistic in terms of color, texture, and
geometry structure. 3D masks are made of different materials, e.g., hard/rigid masks can be made from paper, resin, plaster, or plastic while flexible soft masks are usually composed of silicon or latex. 

In consideration of the facial region covering, PAs can be also separated to \textit{whole} or \textit{partial} attacks. As shown in Fig.~\ref{fig:pipeline}(b), compared with common PAs (e.g., print photo, video replay, and 3D mask) covering the whole face region, a few partial attacks only placed upon specific facial regions (e.g., part-cut print photo, funny eyeglass worn in the eyes region and partial tattoo on the cheek region), which are more obscure and challenging to detect.

\begin{table*}
\centering
\caption{A summary of \textbf{public available datasets} for face anti-spoofing. The upper part of the table lists the datasets recorded via \textit{commercial RGB camera} while the half bottom investigates the datasets with \textit{multiple modalities or specialized sensors}. In the column `\#Live/Spoof', `I' and `V' denotes `images' and `videos', respectively. `\#Sub.' is short for Subjects.} \label{tab:dataset}
\resizebox{0.96\textwidth}{!} {\begin{tabular}{c c c c c c c} 
 \toprule[1pt]
 Dataset \& Reference & Year & \#Live/Spoof & \#Sub. & M\&H & Setup & Attack Types \\
 \midrule
 NUAA~\cite{tan2010face} & 2010 & 5105/7509(I) & 15  & VIS & N/R & Print(flat, wrapped)\\

 \midrule
 YALE\_Recaptured~\cite{peixoto2011face} & 2011 & 640/1920(I) & 10  & VIS & 50cm-distance from 3 LCD minitors & Print(flat)\\

 \midrule
 CASIA-MFSD
~\cite{Zhang2012A} & 2012 & 150/450(V) & 50 & VIS & 7 scenarios and 3 image quality & Print(flat, wrapped, cut), Replay(tablet)\\

 \midrule
REPLAY-ATTACK
~\cite{ReplayAttack} & 2012 & 200/1000(V) & 50 & VIS &  Lighting and holding  & Print(flat), Replay(tablet, phone)\\

 \midrule
Kose and Dugelay
~\cite{kose2013shape} & 2013 & 200/198(I) & 20 & VIS &  N/R  & Mask(hard resin)\\

 \midrule
MSU-MFSD
~\cite{wen2015face} & 2014 & 70/210(V) & 35 & VIS &  Indoor scenario; 2 types of cameras & Print(flat), Replay(tablet, phone)\\

 \midrule
UVAD
~\cite{pinto2015using} & 2015 & 808/16268(V) & 404 & VIS &  \tabincell{c}{Different lighting, background \\and places in two sections}  & Replay(monitor)\\

 \midrule
REPLAY-Mobile
~\cite{costa2016replay} & 2016 & 390/640(V) & 40 & VIS &  5 lighting conditions & Print(flat), Replay(monitor)\\

 \midrule
HKBU-MARs V2
~\cite{liu20163d} & 2016 & 504/504(V) & 12 & VIS &  \tabincell{c}{ 7 cameras from stationary and mobile \\devices and 6 lighting settings} & \tabincell{c}{Mask(hard resin) from \\ Thatsmyface and REAL-f}\\

 \midrule
MSU USSA
~\cite{Patel2016Secure} & 2016 & 1140/9120(I) & 1140 & VIS &  Uncontrolled; 2 types of cameras & Print(flat), Replay(laptop, tablet, phone)\\

 \midrule
SMAD
~\cite{manjani2017detecting} & 2017 & 65/65(V) & - & VIS &  Color images from online resources & Mask(silicone)\\

 \midrule
OULU-NPU
~\cite{Boulkenafet2017OULU} & 2017 &  720/2880(V)  & 55 & VIS &  Lighting \& background in 3 sections  & Print(flat), Replay(phone)\\

 \midrule
Rose-Youtu
~\cite{li2018unsupervised} & 2018 &  500/2850(V) & 20 & VIS &  \tabincell{c}{5 front-facing phone camera; \\5 different illumination conditions}  & \tabincell{c}{Print(flat), Replay(monitor, laptop), \\Mask(paper, crop-paper)}\\

 \midrule
SiW
~\cite{Liu2018Learning} & 2018 &  1320/3300(V)  & 165 & VIS &  \tabincell{c}{4 sessions with variations of distance, \\pose, illumination and expression}  & \tabincell{c}{Print(flat, wrapped), \\Replay(phone, tablet, monitor)}\\

 \midrule
WFFD
~\cite{jia20203d} & 2019 & \tabincell{c}{2300/2300(I)\\140/145(V)} & 745 & VIS &  \tabincell{c}{Collected online; super-realistic; \\ removed low-quality faces}  & Waxworks(wax)\\

 \midrule
SiW-M
~\cite{liu2019deep} & 2019 & 660/968(V) & 493 &  VIS &  \tabincell{c}{Indoor environment with pose, \\lighting and expression variations}  & \tabincell{c}{Print(flat), Replay, Mask(hard resin, \\plastic, silicone, paper, Mannequin),\\ Makeup(cosmetics, impersonation, \\Obfuscation), Partial(glasses, cut paper)}\\

 \midrule
Swax
~\cite{vareto2020swax} & 2020 &  \tabincell{c}{Total 1812(I)\\110(V) } & 55 & VIS &   \tabincell{c}{Collected online; captured \\under uncontrolled scenarios} & Waxworks(wax)\\

 \midrule
CelebA-Spoof
~\cite{zhang2020celeba} & 2020 &  \tabincell{c}{156384/\\469153(I)} & 10177 & VIS &   \tabincell{c}{4 illumination conditions; \\indoor \& outdoor; rich annotations} & \tabincell{c}{Print(flat, wrapped), Replay(monitor, \\tablet, phone), Mask(paper)}\\

 \midrule
\tabincell{c}{RECOD-\\Mtablet}
~\cite{almeida2020detecting} & 2020 &  450/1800(V) & 45 & VIS &   \tabincell{c}{Outdoor environment and\\ low-light \& dynamic sessions} & Print(flat), Replay(monitor)\\

 \midrule
\tabincell{c}{CASIA-SURF\\3DMask}
~\cite{yu2020fas2} & 2020 &  288/864(V) & 48 & VIS &   \tabincell{c}{High-quality identity-preserved; \\ 3 decorations and 6 environments} & Mask(mannequin with 3D print)\\

 \midrule
\tabincell{c}{HiFiMask}
~\cite{liu2021contrastive} & 2021 &  13650/40950(V) & 75 & VIS &   \tabincell{c}{three mask decorations; 7 recording\\ devices; 6 lighting conditions\\ (periodic/random); 6 scenes} & Mask(transparent, plaster, resin)\\

 \midrule[1pt]
 \midrule[1pt]
 
3DMAD
~\cite{erdogmus2014spoofing} & 2013 & 170/85(V) & 17 & VIS, Depth &  3 sessions (2 weeks
interval)  & Mask(paper, hard resin)\\

 \midrule
GUC-LiFFAD
~\cite{raghavendra2015presentation} & 2015 & 1798/3028(V) & 80 & Light field &  \tabincell{c}{Distance of 1.5$\sim$2 m in \\constrained conditions}  & \tabincell{c}{Print(Inkjet paper, Laserjet paper),\\ Replay(tablet)}\\

\midrule
3DFS-DB
~\cite{galbally2016three} & 2016 & 260/260(V) & 26 & VIS, Depth &  Head movement with rich angles & Mask(plastic)\\

 \midrule
\tabincell{c}{BRSU \\ Skin/Face/Spoof}
~\cite{steiner2016reliable} & 2016 & 102/404(I) & 137 & VIS, SWIR &  \tabincell{c}{ multispectral SWIR with 4 wavebands \\935nm, 1060nm, 1300nm and 1550nm} & Mask(silicon, plastic, resin, latex)\\

 \midrule
Msspoof
~\cite{chingovska2016face} & 2016 & 1470/3024(I) & 21 & VIS, NIR &  7 environment conditions & Black\&white Print(flat)\\

 \midrule
MLFP
~\cite{agarwal2017face} & 2017 & 150/1200(V) & 10 & \tabincell{c}{VIS, NIR, \\Thermal} &  \tabincell{c}{Indoor and outdoor with fixed\\ and random backgrounds} & Mask(latex, paper)\\

 \midrule
ERPA
~\cite{bhattacharjee2017you} & 2017 & Total 86(V) & 5 & \tabincell{c}{VIS, Depth, \\NIR, Thermal} &  \tabincell{c}{Subject positioned close (0.3$\sim$0.5m)\\ to the 2 types of cameras} & \tabincell{c}{Print(flat), Replay(monitor), \\ Mask(resin, silicone)}\\

 \midrule
LF-SAD
~\cite{liu2019light} & 2018 &  328/596(I)  & 50 & Light field &  \tabincell{c}{Indoor fix background, captured\\ by Lytro ILLUM camera}  & Print(flat, wrapped), Replay(monitor)\\

 \midrule
CSMAD
~\cite{bhattacharjee2018spoofing} & 2018 &  104/159(V+I)  & 14 & \tabincell{c}{VIS, Depth, \\NIR, Thermal} &  4 lighting conditions  & Mask(custom silicone)\\

 \midrule
3DMA
~\cite{xiao20193dma} & 2019 & 536/384(V) & 67 & VIS, NIR &  \tabincell{c}{48 masks with different ID; 2 illumi-\\ nation \& 4 capturing distances}  & Mask(plastics)\\

 \midrule
CASIA-SURF
~\cite{casiasurf} & 2019 &  \tabincell{c}{3000/\\18000(V)} & 1000 &  \tabincell{c}{VIS, Depth,\\ NIR} &  \tabincell{c}{Background removed; Randomly \\cut  eyes, nose or mouth areas}  & Print(flat, wrapped, cut)\\

 \midrule
WMCA
~\cite{george2019biometric} & 2019 & 347/1332(V) & 72 &  \tabincell{c}{VIS, Depth,\\ NIR, Thermal}  &  \tabincell{c}{6 sessions with different \\backgrounds and illumination; \\pulse data for bonafide recordings}  & \tabincell{c}{Print(flat), Replay(tablet),\\ Partial(glasses), Mask(plastic, \\silicone, and paper, Mannequin)}\\

 \midrule
CeFA
~\cite{li2020casia} & 2020 &   \tabincell{c}{6300/\\27900(V)} & 1607 & \tabincell{c}{VIS, Depth,\\NIR} &   \tabincell{c}{3 ethnicities; outdoor \& indoor;\\ decoration with wig and glasses} & \tabincell{c}{Print(flat, wrapped), Replay, \\Mask(3D print, silica gel)}\\

 \midrule
HQ-WMCA
~\cite{heusch2020deep} & 2020 &  555/2349(V) & 51 & \tabincell{c}{VIS, Depth, \\NIR, SWIR,\\Thermal} &   \tabincell{c}{Indoor; 14 `modalities', including \\4
NIR and 7 SWIR wavelengths; \\masks and mannequins were \\heated up to reach body temperature} & \tabincell{c}{Laser or inkjet  Print(flat), \\Replay(tablet, phone), Mask(plastic, \\silicon, paper, mannequin), Makeup,\\ Partial(glasses, wigs, tatoo)}\\

 \midrule
PADISI-Face
~\cite{rostami2021detection} & 2021 & 1105/924(V) & 360 &  \tabincell{c}{VIS, Depth,\\ NIR, SWIR, \\Thermal}  &  \tabincell{c}{Indoor, fixed green background,  \\60-frame sequence of \\ 1984 × 1264 pixel images}  & \tabincell{c}{Print(flat), Replay(tablet, phone),\\ Partial(glasses,funny eye), Mask(plastic, \\silicone, transparent, Mannequin)}\\

 \bottomrule[1pt]
 \end{tabular}}
\end{table*}

\begin{figure*}
\centering
\includegraphics[scale=0.28]{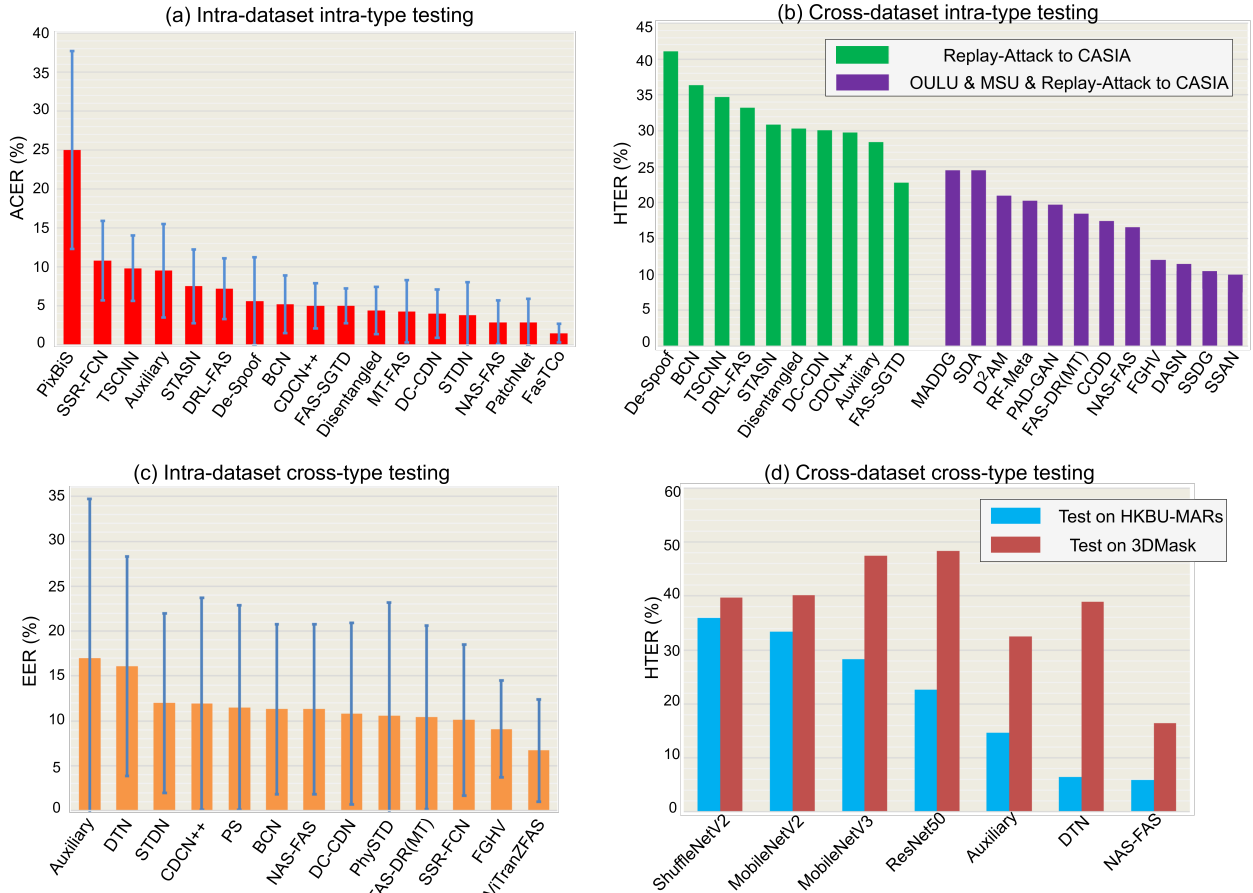}
\vspace{-1.0em}
  \caption{ 
  The performance of deep FAS approaches on four mainstream evaluation protocols. The lower ACER, HTER and EER, the better performance. (a) Intra-dataset intra-type testing on the Protocol-4 of OULU-NPU. (b) Cross-dataset intra-type testing on CASIA-MFSD when training on single Replay-Attack dataset (see green columns) or multiple datasets including OULU-NPU, MSU-MFSD, and Replay-Attack (see purple columns). (c) Intra-dataset cross-type testing on SiW-M with leave-one-type-out setting. (d) Cross-dataset cross-type testing on 3D mask FAS datasets including HKBU-MARs~\cite{liu20163d} and CASIA-SURF 3DMask when training on OULU-NPU and SiW datasets with only 2D attacks. }
\label{fig:protocols}
\end{figure*}

\subsection{Datasets for Face Anti-Spoofing}

Large-scale and diverse datasets are pivotal for deep learning based methods during both training and evaluating phases. We summarize prevailing public FAS datasets in Table~\ref{tab:dataset} in terms of data amount, subject numbers, modality/sensor, environmental setup, and attack types. We also visualize some samples under different environmental conditions and modalities in Fig.~\ref{fig:dataset}(a) and (b), respectively.

It can be seen from Table~\ref{tab:dataset} that most datasets~\cite{tan2010face,peixoto2011face,Zhang2012A,ReplayAttack,kose2013shape,wen2015face,pinto2015using} contain only a few attack types (e.g., print and replay attacks) under simple recording conditions (e.g., indoor scene) from the early stage (i.e., year 2010-2015), which have limited variations in samples for generalized FAS training and evaluation. Subsequently, there are three main trends for dataset progress: 1) \textit{large scale data amount}. For example, the recently released datasets CelebA-Spoof~\cite{zhang2020celeba} and HiFiMask~\cite{liu2021contrastive} contain more than 600000 images and 50000 videos, respectively, where most of them are with PAs; 2) \textit{diverse data distribution}. Besides common print and replay attacks recorded in controllable indoor scenario, more and more novel attack types as well as complex recording conditions are considered in recent FAS datasets. For example, there are 13 fine-grained attack types in SiW-M~\cite{liu2019deep} while HiFiMask~\cite{liu2021contrastive} consists of 3D masks attacks with three kinds of materials (transparent, plaster, resin) recorded under six lighting conditions and six indoor/outdoor scenes; and 3) \textit{multiple modalities and specialized sensors}. Apart from traditional visible RGB camera, some recent datasets also consider various modalities (e.g., NIR~\cite{heusch2020deep,li2020casia,casiasurf,george2019biometric}, Depth~\cite{heusch2020deep,li2020casia,casiasurf,george2019biometric}, Thermal~\cite{heusch2020deep,george2019biometric}, and SWIR~\cite{heusch2020deep}) and other specialized sensors (e.g., Light field camera~\cite{raghavendra2015presentation,liu2019light}). All these advanced factors facilitate the area of FAS in both academic research and industrial deployment.

\begin{figure*}
\centering
\includegraphics[scale=0.44]{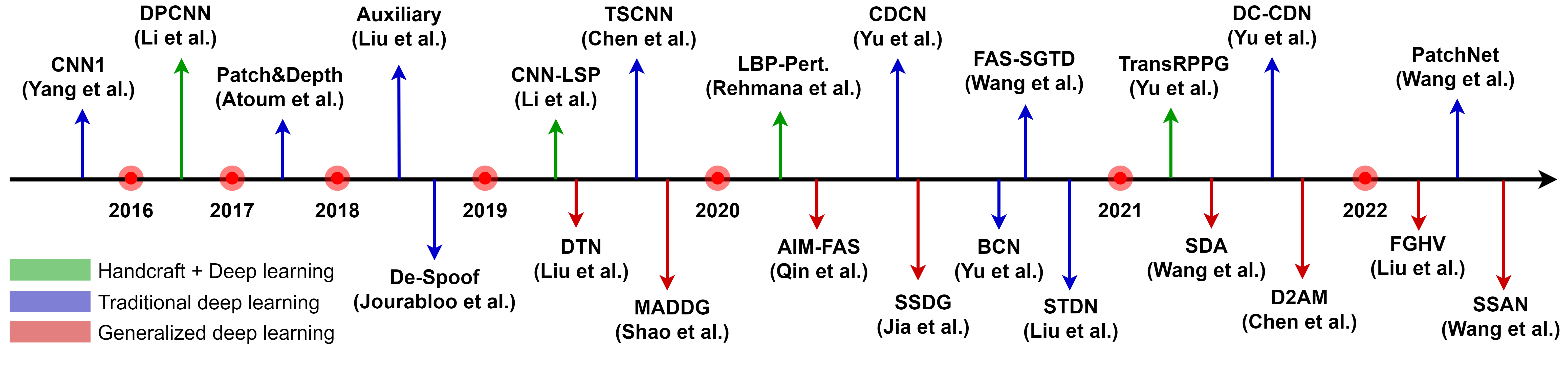}
\vspace{-1.0em}
  \caption{ 
  Chronological overview of the milestone deep learning based FAS methods using commercial RGB camera.}
  
\label{fig:milestone}
\vspace{-1.0em}
\end{figure*}

\vspace{-0.8em}

\subsection{Evaluation Metrics}

As FAS systems usually focus on the concept of bonafide and PA acceptance and rejection, two basic metrics False Rejection Rate (FRR) and False Acceptance Rate (FAR)~\cite{galbally2012high} are widely used. The ratio of incorrectly accepted spoofing attacks defines FAR, whereas FRR stands for the ratio of incorrectly rejected live accesses~\cite{chingovska2014biometrics}. FAS follows ISO/IEC DIS 30107- 3:2017~\cite{iso2017information} standards to evaluate the performance of the FAS systems under different scenarios. The most commonly used metrics in both intra- and cross-testing scenarios is Half Total Error Rate (\textit{HTER})~\cite{chingovska2014biometrics}, Equal Error Rate (\textit{EER})~\cite{ramachandra2017presentation}, and Area Under the Curve (\textit{AUC}). HTER is found out by calculating the average of FRR (ratio of incorrectly rejected bonafide score) and FAR (ratio of incorrectly accepted PA). EER is a specific value of HTER at which FAR and FRR have equal values. AUC represents the degree of separability between bonafide and spoofings. 

Recently, Attack Presentation Classification Error Rate (\textit{APCER}), Bonafide Presentation Classification Error Rate (\textit{BPCER}) and Average Classification Error Rate (\textit{ACER}) suggested in ISO standard~\cite{iso2017information} are also used for intra-dataset testings~\cite{Boulkenafet2017OULU,Liu2018Learning}. BPCER and APCER measure bonafide and attack classification error rates, respectively. ACER is calculated as the mean of BPCER and APCER, evaluating the reliability of intra-dataset performance.

\subsection{Evaluation Protocols}
\label{sec:protocols}

To evaluate the discrimination and generalization capacities of the deep FAS models, various protocols have been established. We summarize the development of deep FAS approaches on four representative protocols in Fig.~\ref{fig:protocols} and Tables-A 2, 3 and 4 (in Appendix).


\vspace{0.3em}

\noindent\textbf{Intra-Dataset Intra-Type Protocol.}\quad 
Intra-dataset intra-type protocol has been widely used in most FAS datasets to evaluate the model's discrimination ability for spoofing detection under scenarios with slight domain shift. As the training and testing data are sampled from the same datasets, they share similar domain distribution in terms of the recording environment, subject behavior, etc. (see Fig.~\ref{fig:dataset}(a) for examples). The most classical intra-dataset intra-type testing is the Protocol-4 of OULU-NPU dataset~\cite{Boulkenafet2017OULU}, and the performance comparison of recent deep FAS methods on this protocol is shown in Fig.~\ref{fig:protocols}(a). Due to the strong discriminative feature representation ability via deep learning, many methods (e.g., CDCN++~\cite{yu2020searching}, FAS-SGTD~\cite{wang2020deep}, Disentangled~\cite{zhang2020face}, MT-FAS~\cite{qin2021meta}, DC-CDN~\cite{yu2021dual}, STDN~\cite{liu2020disentangling}, NAS-FAS~\cite{yu2020fas2}, FasTCo~\cite{xu2020improving} and PatchNet~\cite{wang2022patchnet}) have reached satisfied performance ($\textless$5\% ACER) under small domain shifts in terms of external environment, attack mediums and recording camera variation. More intra-dataset intra-type results on OULU-NPU (4 sub-protocols) and SiW (3 sub-protocols) datasets are listed in Table-A 2 (in Appendix).

\vspace{0.3em}

\noindent\textbf{Cross-Dataset Intra-Type Protocol.}\quad 
This protocol focuses on cross-dataset level domain generalization ability measurement, which usually trains models on one or several datasets (source domains) and then tests on unseen datasets (shifted target domain). We summarize recent deep FAS approaches on two favorite cross-dataset testings~\cite{yu2020searching,shao2019multi} in Fig.~\ref{fig:protocols}(b). It can be seen from green columns that, when trained on Replay-Attack and tested on CASIA-MFSD, most deep models perform poorly ($\textgreater$20\% HTER) due to the serious lighting and camera resolution variations. In contrast, when trained on multiple source datasets (i.e., OULU-NPU, MSU-MFSD, and Replay-Attack), domain generalization based methods achieve acceptable performance (especially SSDG~\cite{jia2020single} and SSAN~\cite{wang2022domain} with 10.44\% and 10.00\% HTER, respectively). In real-world cross-testing cases, small amount of target domain data are easily obtained, which can also be utilized for domain adaptation~\cite{jia2021unified} to mitigate domain shifts further. More cross-dataset intra-type testings among OULU-NPU, CASIA-MFSD, Replay-Attack, and MSU-MFSD datasets with different numbers of source domains for training can be found in Table-A 3 (in Appendix).  

\vspace{0.3em}

\noindent\textbf{Intra-Dataset Cross-Type Protocol.}\quad  The protocol adopts `leave one attack type out' to validate the model's generalization for unknown attack types, i.e., one kind of attack type only appears in the testing stage. Considering the rich (13 kinds) attack types, SiW-M~\cite{liu2019deep} is investigated in this protocol, and the corresponding results are illustrated in Fig.~\ref{fig:protocols}(c). Most of the deep models achieve around 10\% EER and with large standard deviations among all attack types, which indicates the huge challenges in this protocol. Benefited from the large-scale pretraining, ViTranZFAS~\cite{liu2019deep} achieves surprising 6.7\% EER, implying the promising usage of transfer learning for unknown attack type detection. Detailed intra-dataset cross-type testing results on SiW-M with the leave-one-type-out setting are shown in Table-A 4 (in Appendix).

\vspace{0.4em}

\noindent\textbf{Cross-Dataset Cross-Type Protocol.}\quad Although the three protocols mentioned above mimic most factors in real-world applications, they do not consider the most challenging case, i.e., cross-dataset cross-type testing.~\cite{yu2020fas2} proposes a 'Cross-Dataset Cross-Type Protocol' to measure the FAS model's generalization on both unseen domain and unknown attack types.
In this protocol, OULU-NPU and SiW (with 2D attacks) are mixed for training, while HKBU-MARs and 3DMask (with 3D attacks) are used for testing. It can be seen from Fig.~\ref{fig:protocols}(d) that recent deep models (DTN~\cite{liu2019deep} and NAS-FAS~\cite{yu2020fas2}) hold good generalization for lab-controlled low-fidelity 3D mask detection on HKBU-MARs but still cannot satisfactorily detect unrestricted high fidelity masks on 3DMask.

Besides these four mainstream evaluation protocols, more new trends about practical protocol settings (e.g., semi-/un-supervised, real-world open-set, and dynamic multimodality) will be discussed in Section~\ref{sec:discussion}.

\section{Deep FAS with Commercial RGB Camera} \label{sec:RGB}
As commercial RGB camera is widely used in many real-world applicational scenarios (e.g., access control system and mobile device unlocking), in this section, we will review existing commercial RGB camera based FAS methods. Several
milestone deep FAS methods are illustrated in Fig.~\ref{fig:milestone}. 


\subsection{Hybrid (Handcraft + Deep Learning) Method}
Although deep learning and CNNs have achieved great success in many computer vision tasks (e.g., image classification~\cite{He2015Deep,huang2017densely}, semantic segmentation~\cite{long2015fully}, and object detection~\cite{ren2016faster}), they suffer from the overfitting problem for the FAS task due to the limited amount and diversity of the training data. As handcrafted features (e.g., LBP~\cite{ahonen2006face}, HOG~\cite{dalal2005histograms} descriptors, image quality~\cite{galbally2014face}, optical flow motion~\cite{brox2010large}, and rPPG clues~\cite{niu2020video}) have been proven to be discriminative to distinguish bonafide from PAs, some recent \textit{hybrid} works combine handcrafted features with deep features for FAS. Typical properties of these hybrid methods are summarized in Table-A 5 (in Appendix).

\begin{figure}
\centering
\includegraphics[scale=0.47]{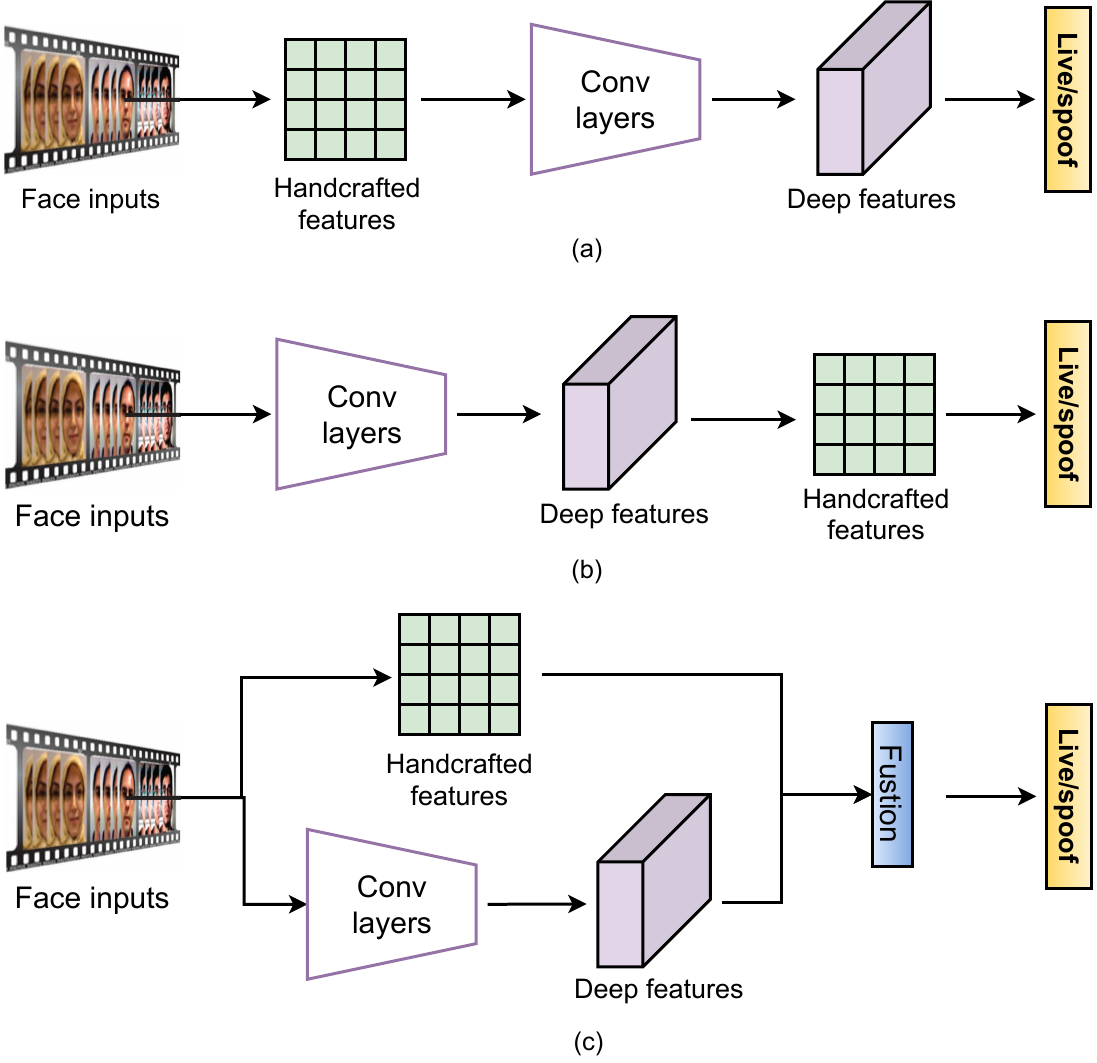}
  \caption{ 
   Hybrid frameworks for FAS. (a) Deep features from handcrafted features. (b) Handcrafted features from deep features. (c) Fused handcrafted and deep features.
  }
\label{fig:hybrid}
\end{figure}

Some FAS approaches firstly extract handcrafted features from face inputs, and then employ CNNs for semantic feature representation (see Fig.~\ref{fig:hybrid}(a) for paradigm). On one hand, color texture based static features are extracted from each frame, and then are feed into the deep model. Based on the rich low-level texture features, deep model is able to mine texture-aware semantic clues. To this end, Cai and Chen~\cite{cai2019learning} adopt multi-scale color LBP features as local texture descriptors, then a random forest is cascaded for semantic representation. Similarly, Khammari~\cite{khammari2019robust} extracts LBP and Weber local descriptor encoded CNN features, which are combined to preservate the local intensity and edge orientation information. However, compared with the original face input, local descriptor based features lose pixel-level details thus limiting the model performance. On the other hand, dynamic features (e.g., motion, illumination changes, physiological signals) across temporal frames are also effective CNN inputs. Feng et al.~\cite{feng2016integration} propose to train a multi-layer perceptron from the extracted dense optical flow-based motions, which reveal anomalies in print attacks. Moreover, Yu et al.~\cite{yu2021transrppg} construct spatio-temporal rPPG maps from face videos, and use a vision transformer to capture the periodic heartbeat liveness features for the bonafide. However, head motions and rPPG signals are easily imitated in the replay attack, making such dynamic clues less reliable. Basing on the fact that replay attacks usually have abnormal reflection changes, Li et al.~\cite{li20203d} propose to capture such illumination changes using a 1D CNN with inputs of the intensity difference histograms from reflectance images.

Several other hybrid FAS methods extract handcrafted features from deep convolutional features, which follow the hybrid framework in Fig.~\ref{fig:hybrid}(b). To reduce the FAS-unrelated redundancy, Li et al.~\cite{Li2017An} utilize the block principal component analysis (PCA) to filter out the irrelevant deep features from pretrained VGG-face model. Besides PCA-based dimension reduction, Agarwal et al.~\cite{agarwal2019chif} explicitly extract the color LBP descriptor from the shallow convolutional features, which contains richer low-level statistics. In addition to static spoof patterns, some works also explore handcrafted dynamic temporal clues from well-trained deep models. Asim et al.~\cite{asim2017cnn} and Shao et al.~\cite{shao2018joint} extract deep dynamic textures and motion features using LBP-TOP~\cite{zhao2007dynamic} and optical flow from the sequential convolutional features, respectively. One limitation of this hybrid framework is that the handcrafted features are highly dependent on the well-trained convolutional features, and it is still unknown whether shallow or deep convolutional features are more suitable for different kinds of handcrafted features.

As handcrafted and deep convolutional features hold different properties, another favorite hybrid framework (see Fig.~\ref{fig:hybrid}(c)) fuses them for more generic representation. To make more reliable predictions, Sharifi~\cite{sharifi2019score} proposes to fuse the predicted scores from both handcrafted LBP features and deep VGG16 model. However, it is challenging to determine the optimal score weights for these two kinds of features. Besides score-level fusion, Rehmana et al.~\cite{rehman2019perturbing,rehman2020enhancing} propose to utilize HOG and LBP maps to perturb and modulate the low-level convolutional features. Despite the fact that local prior knowledge from handcrafted features enhances discriminative capacity, the overall model suffers from semantic representation degradation. In terms of the temporal methods, to leverage the dynamic discrepancy between the bonafide and PAs, Li et al.~\cite{li2019replayed} extract intensity variation features via 1D CNN, which are fused with the motion blur features from motion magnified face videos for replay attack detection.

Overall, benefitted from the explicit expert-designed feature extraction, hybrid methods are able to represent particular non-texture clues (e.g., temporal rPPG and motion blur), which are hard to capture via end-to-end texture-based FAS models. However, the shortcomings are also obvious: 1) handcrafted features highly rely on the expert knowledge and not learnable, which are inefficient once enough training data are available; and 2) there might be feature gaps/incompatibility between handcrafted and deep features, resulting in performance saturation.

\subsection{Traditional Deep Learning Method}
\begin{figure}
\centering
\includegraphics[scale=0.47]{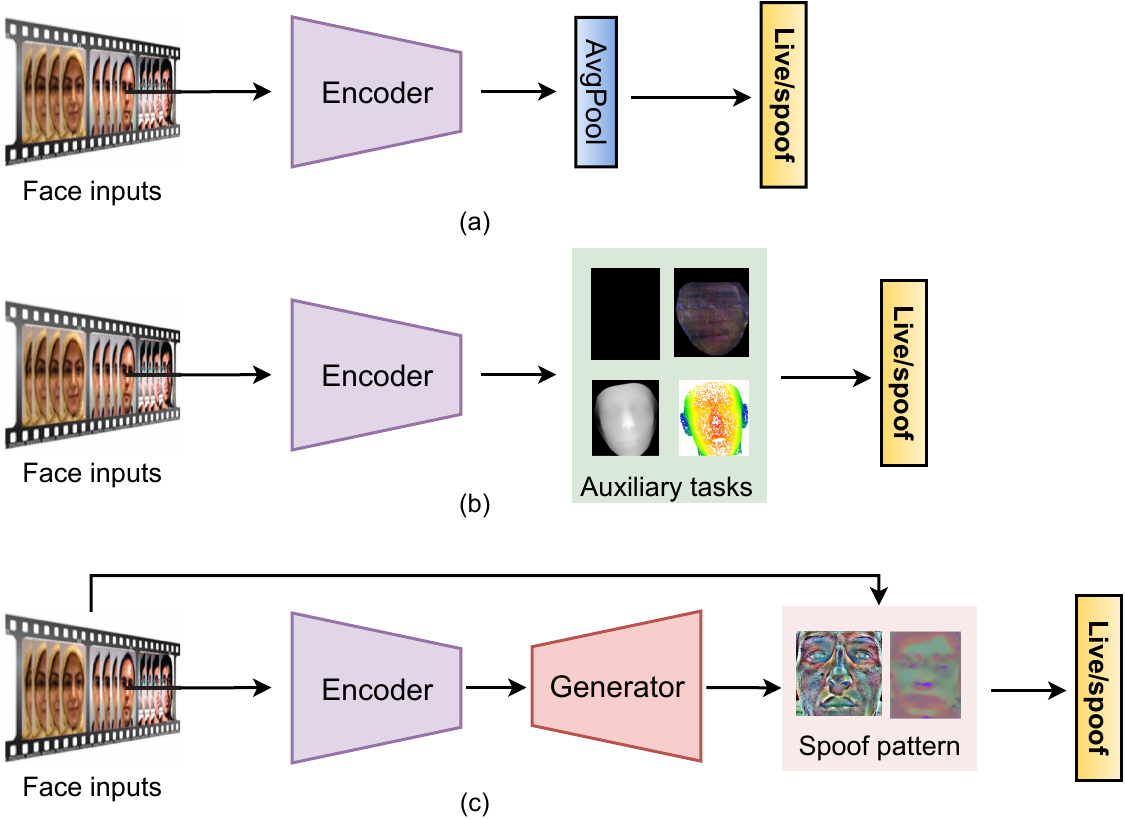}
  \caption{ 
   Traditional end-to-end deep learning frameworks for FAS. (a) Direct supervision with binary cross entropy loss. (b) Pixel-wise supervision with auxiliary tasks. (c) Pixel-wise supervision with generative model for implicit spoof pattern representation.
  }
\label{fig:pixelwise}
\end{figure}

Benefited from the development of the advanced CNN architectures~\cite{huang2017densely,ronneberger2015u} and regularization~\cite{ioffe2015batch,srivastava2014dropout} techniques as well as the recent released large-scale FAS datasets~\cite{Boulkenafet2017OULU,zhang2020celeba,liu2021contrastive}, end-to-end deep learning based methods attract more and more attention, and dominate the field of FAS. Different from the hybrid methods which integrate parts of handcrafted features without learnable parameters, \textit{Traditional} deep learning based methods directly learn the mapping functions from face inputs to spoof detection. Traditional deep learning frameworks usually include: 1) direct supervision with binary cross-entropy loss (see Fig.~\ref{fig:pixelwise}(a)); and 2) pixel-wise supervision with auxiliary tasks (see Fig.~\ref{fig:pixelwise}(b)) or generative models (see Fig.~\ref{fig:pixelwise}(c)).

\subsubsection{Direct Supervision with Binary Cross Entropy Loss}
As FAS can be intuitively treated as a binary (bonafide vs. PA) classification task, numerous end-to-end deep learning methods are directly supervised with binary cross-entropy (CE) loss as well as extended losses (e.g., triplet loss~\cite{hermans2017defense}), which are summarized in Table-A 6 (in Appendix). 

On one side, researchers have proposed various network architectures supervised by binary CE loss for FAS. Yang et al.~\cite{yang2014learn} propose the first end-to-end deep FAS method using 8-layer shallow CNN for feature representation. However, due to the limited scale and diversity of datasets, CNN-based models easily overfit in the FAS task. To alleviate this issue, some works~\cite{lucena2017transfer,chen2019attention,george2020effectiveness} finetune the ImageNet-pretrained models (e.g., VGG16, ResNet18 and  vision transformer) for FAS. Towards moblie-level FAS applications, Heusch et al.~\cite{heusch2020deep} consider using the lightweight MobileNetV2~\cite{sandler2018mobilenetv2} for efficient FAS. The above-mentioned generic backbones usually focus on high-level semantic representation while neglect low-level features, which are also important for spoof pattern mining. To better leverage the multi-scale features for FAS, Deb and Jain~\cite{deb2020look} propose to use a shallow fully convolutional network (FCN) to learn local discriminative cues from face images in a self-supervised manner. Besides the single-frame-based appearance features, several works~\cite{Xu2016Learning,muhammad2019face,yang2019face,ge2020face} consider the temporal discrepancy between bonafide and PAs, and cascade multi-frame-based CNN features with LSTM~\cite{hochreiter1997long} for robust dynamic clues propagation.

On the other side, considering the weak intra- and inter-class constraints from binary CE, a few works modify binary CE loss to provide CNNs more discriminative supervision signals. Instead of binary constraints, Xu et al.~\cite{xu2020improving} rephrase FAS as a fine-grained classification problem, and the type labels (e.g., bonafide, print, and replay) are used for multi-class supervision. In this way, the particular properties (e.g., materials) of PAs could be represented. However, FAS models supervised with multi-class CE loss still have confused live/spoof distributions especially on hard live/spoof samples. For instance, high-fidelity PAs have similar appearance clues as the corresponding bonafide. On one hand, to learn a compact space with small intra-class distances but large inter-class distances, Hao~\cite{hao2019face} and Almeida et al.~\cite{almeida2020detecting} introduce contrastive loss and triplet loss, respectively. However, different from vision retrieval tasks, the bonafide and PAs in FAS task usually hold asymmetric intra-distributions (more compact and diverse, respectively). Based on this evidence, Wang et al.~\cite{wang2022patchnet} propose to supervise the FAS patch models via an asymmetric angular-margin softmax loss to relax the intra-class constraints among PAs. On the other hand, to provide more confident predictions on hard samples, Chen et al.~\cite{chen2021camera} adopt the binary focal loss to guide the model to enlarge the margin between live/spoof samples and achieve strong discrimination for hard samples.

Overall, both binary CE loss and its extended losses are easy and efficient to use, which supervise deep FAS models to fastly converge. However, these supervision signals only provide global (spatial/temporal) constraints for live/spoof embedding learning, which may causes FAS models to easily overfit to unfaithful patterns. Furthermore, FAS models with binary supervision are usually black-box and the characteristic of their learned features  are hard to understand.

\subsubsection{Pixel-wise Supervision}

Deep models directly supervised by binary loss might easily learn unfaithful patterns (e.g., screen bezel). In contrast, pixel-wise supervision can provide more fine-grained and contextual task-related clues for better intrinsic feature learning. On one hand, based on the physical clues and discriminative design philosophy, auxiliary supervision signals such as pseudo depth labels~\cite{Atoum2018Face,Liu2018Learning}, binary mask label~\cite{george2019deep,liu2019deep,sun2020face} and reflection maps~\cite{yu2020face,kim2019basn} are developed for local live/spoof clues description. On the other hand, generative models with explicit pixel-wise supervision (e.g., original face input reconstruction~\cite{mohammadi2020improving,liu2020physics}) are recently utilized for generic spoof pattern estimation. We summarize the representative pixel-wise supervision methods in Table-A 7 (in Appendix).

\begin{figure*}
\centering
\includegraphics[scale=0.45]{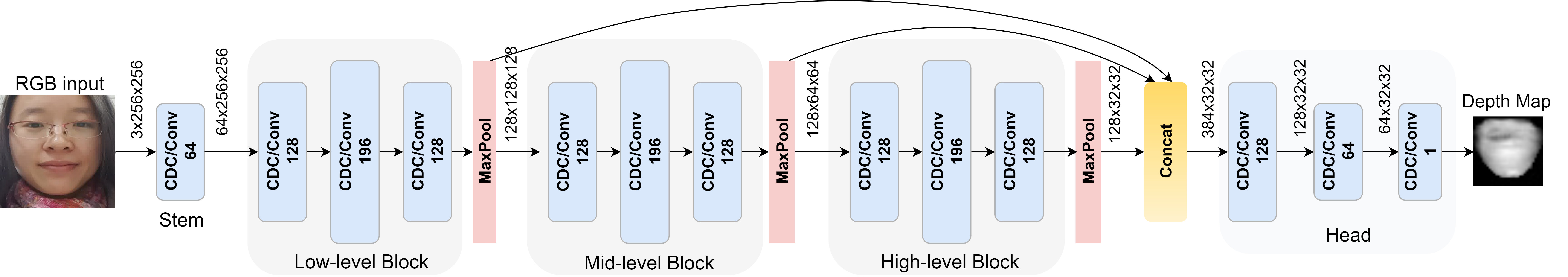}
  \caption{ 
   The shared architecture of CDCN~\cite{yu2020searching}/DepthNet~\cite{Liu2018Learning}. Inside the blue block are the convolutional filters with 3x3 kernel size and their feature dimensionalities. `CDC' and `Conv' suggest central difference convolution adopted in CDCN and vanilla convolution adopted in DepthNet, respectively.}
  
\label{fig:cdcn}
\end{figure*}

\begin{figure*}
\centering
\includegraphics[scale=0.5]{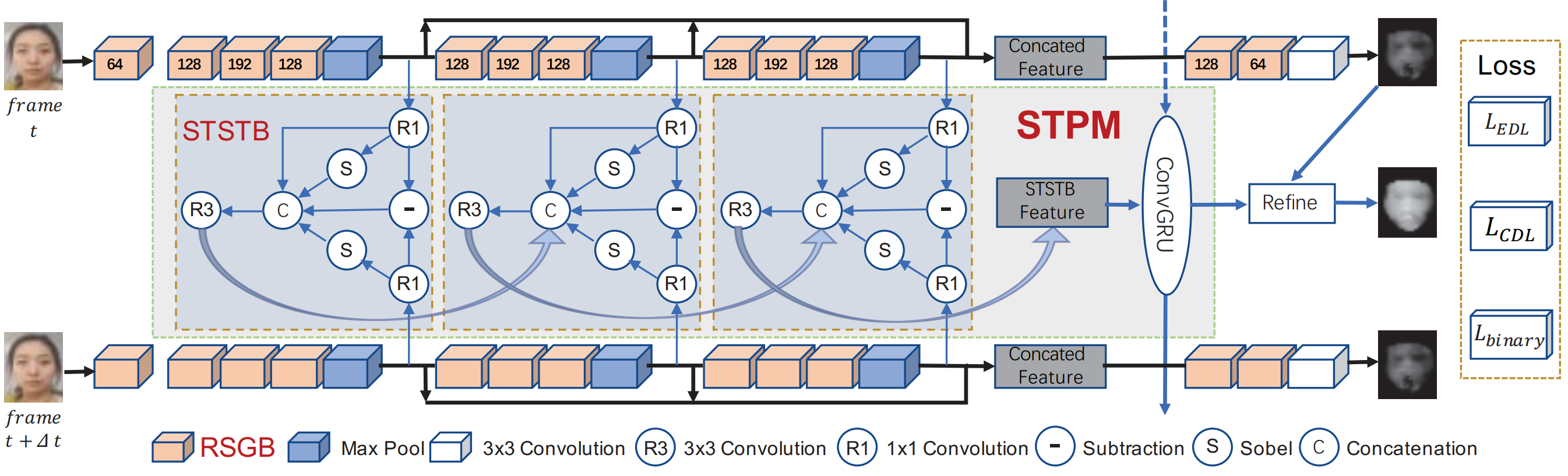}
  \caption{ 
   The network structure of FAS-SGTD~\cite{wang2020deep}. The inputs are consecutive frames with a fixed interval. Each frame is processed by cascaded Residual Spatial Gradient Block (RSGB) with a shared backbone which generates a corresponding coarse depth map. The number in RSGB cubes denotes the output channels. Spatio-Temporal Propagation Module (STPM) is plugged between frames for estimating the temporal depth and refining the corresponding coarse depth map.}
\label{fig:SGTD}
\end{figure*}

\vspace{0.4em}
\noindent\textbf{Pixel-wise supervision with Auxiliary Task.}\quad
According to the human prior knowledge of FAS, most PAs (e.g., plain printed paper and electronic screen) merely have no genuine facial depth information, which could be utilized as discriminative supervision signals. As a result, some recent works~\cite{Atoum2018Face,peng2020ts,yu2020searching,wang2020deep} adopt pixel-wise \textit{pseudo depth} labels to guide the deep models, enforcing them predict the genuine depth for live samples while zero maps for the spoof ones. Atoum et al.~\cite{Atoum2018Face} first leverage pseudo depth labels to guide the multi-scale FCN (namely `DepthNet' for simplicity). Thus, the well-trained DepthNet is able to predict holistic depth maps as decision evidence. To further improve the fine-grained intrinsic feature representation capacity, Yu et al.~\cite{yu2020searching} replace vanilla convolution in DepthNet with central difference convolution (CDC) to form the CDCN architecture (see Fig.~\ref{fig:cdcn} for detailed structures). In terms of static architectures, DepthNet and CDCN are favorite and widely used in the deep FAS community due to their compactness and excellent performance. Many recent variants~\cite{yu2020fas2,yu2021dual,wu2021dual} are established based on the DepthNet/CDCN. As for the temporal architectures, FAS-SGTD~\cite{wang2020deep} is classical and well-known for its excellent short- and long-term micro-motion estimation, which can be utilized for accurate facial depth prediction. The detailed architecture of FAS-SGTD is illustrated in Fig.~\ref{fig:SGTD}, which is later modified and extended in a transformer counterpart~\cite{wang2022learning}.


Synthesizing 3D shape labels for every training sample is costly and not accurate enough, and also lacks the reasonability for the PAs with real depth (e.g., 3D mask and Mannequin). In contrast, binary mask label~\cite{liu2019deep,george2019deep,hossaindeeppixbis,yu2020auto2,liu2020disentangling} is easier to be generated and more generalizable to all PAs. Specifically, binary supervision would be provided for the deep embedding features in each spatial position. In other words, through the binary mask label, we can find whether PAs occur in the corresponding patches, which is attack-type-agnostic and spatially interpretable. George and Marcel~\cite{george2019deep} are the first to introduce deep pixel-wise binary supervision to predict the intermediate confidence map for the cascaded final binary classification. With sufficient pixel-wise supervision, the backbone DenseNet121 converges well and is able to provide patch-wise live/spoof predictions. As subtle spoof clues (e.g., moiré pattern) usually exist in different spatial regions with different intensity, vanilla pixel-wise binary supervision treats all patches with equal contributions, which might lead to biased feature representation. To tackle this issue, Hossaind et al.~\cite{hossaindeeppixbis} propose to add a learnable attention module for feature refinement before calculating the deep pixel-wise binary loss, which benefits the salient information propagation. Though flexible and easy to use, current binary mask labels usually assume all pixels in the face region have the same live/spoof distributions thus generate all `one' and `zero' maps for bonafide and PAs, respectively. However, such labels are inaccurate and noisy to learn when encountering partial attacks (e.g., FunnyEye).




 Besides the mainstream depth map and binary mask labels, there are several informative auxiliary supervisions (e.g., pseudo reflection map~\cite{kim2019basn,yu2020face,zhang2020celeba}, 3D point cloud map~\cite{li3dpc}, ternary map~\cite{sun2020face}, and Fourier spectra~\cite{roy2021bi}). According to the discrepancy of facial material-related albedo between the live skin and spoof mediums, Kim et al.~\cite{kim2019basn} propose to supervise deep models with both depth and reflection labels. Moreover, to further enhance the type-agnostic generalization, binary mask maps are introduced in~\cite{yu2020face} to train the bilateral convolutional networks with all these three pixel-wise supervisions simultaneously. Unlike binary mask labels considering all spatial positions including live/spoof-unrelated background, Sun et al.~\cite{sun2020face} remove the face-unrelated parts and leave the entire face regions as a refined binary mask called `ternary map', which eliminates the noise outside the face and benefits the facial spoof clue mining. Based on the rich texture and geometry discrepancy between the bonafide and PAs, deep models with other auxiliary supervisions from the Fourier map~\cite{jourabloo2018face,roy2021bi}, LBP texture map~\cite{zhang2020face}, and sparse 3D point cloud map~\cite{li3dpc}, also show their excellent representation capability.

Overall, pixel-wise auxiliary supervision benefits the physically meaningful and explainable live/spoof feature learning (e.g., reflection and depth supervisions for material and geometry representation, respectively). Moreover, a reliable and generalized FAS model can be supervised with multiple complementary auxiliary supervisions (e.g., depth, reflection, and albedo) in a multi-task learning fashion~\cite{yu2020face}. However, two limitations of auxiliary supervision should be mentioned: 1) pixel-wise supervision usually relies on the high-quality (e.g., high-resolution) training data for fine-grained spoof clue mining, and is harder to provide effective supervision signals when training data are too noisy and with low quality; and 2) the pseudo auxiliary labels are either human-designed or generated by other off-the-shelf algorithms, which are not always trustworthy.

\vspace{0.4em}
\noindent\textbf{Pixel-wise Supervision with Generative Model.}\quad   Despite the fine-grained supervision signal in the auxiliary task, it is still hard to understand whether the deep black-box models represent intrinsic FAS features. Recently, one hot trend is to mine the visual spoof patterns existing in the spoof samples, aiming to provide a more intuitive interpretation of the sample’s spoofness. We summarize such kind of generative models with pixel-wise supervision in the lower part of Table-A 7 (in Appendix). In consideration of the strong physical-inspired constraints of auxiliary pixel-wise supervision, several works relax such explicit supervision signals and provide a broader space for implicit spoof clues mining. Jourabloo et al.~\cite{jourabloo2018face} rephrase FAS as a spoof noise modeling problem, and design an encoder-decoder architecture to estimate the underlying spoof patterns with relaxed pixel-wise supervisions (e.g., zero-noise map for live faces). With such unilateral constraint on the bonafide, the models are able to mine the spoof clues flexibly for PAs. Similarly, Feng et al.~\cite{feng2020learning} design a spoof cue generator to minimize the spoof cues of live samples while imposes no explicit constraints on those of spoof samples. Unlike above-mentioned works forcing strict constraints on live samples, Mohammadi et al.~\cite{mohammadi2020improving} use the reconstruction-error maps computed from a live-face-pretrained autoencoder for spoofing detection. As such error maps are generated from the residual noises of reconstructed live faces without human-defined elements, they are robust under domain shift with knowledge clue change. However, the low-quality reconstructed faces from autoencoder may lead to noisy residual error maps.

Besides direct spoof pattern generation, Qin et al.~\cite{qin2021meta} propose to automatically generate pixel-wise labels via a meta-teacher framework, which is able to provide better-suited supervision for the student FAS models to learn sufficient and intrinsic spoofing cues. However, only the learnable spoof supervision is generated in~\cite{qin2021meta}. Therefore, how to generate the optimal pixel-wise signals automatically for both live and spoof samples is still worth exploring.


Overall, pixel-wise supervision with generative model usually relaxes the expert-designed hard constraints (e.g., auxiliary tasks), and leaves the decoder to reconstruct more natural spoof-related trace. Thus, the predicted spoof patterns are strongly data-driven and have explainable views. The generated spoof patterns are visually insightful, and are challenging to manually describe with human prior knowledge. However, such soft pixel-wise supervision might easily fall into the local optimum and overfit on unexpected interference (e.g., sensor noise), which would generalize poorly under real-world scenarios. Combining explicit auxiliary supervision with generative model based supervision for jointly training might alleviate this issue.

\subsection{Generalized Deep Learning Method}

Traditional end-to-end deep learning based FAS methods might generalize poorly on unseen dominant conditions (e.g., illumination, facial appearance, and camera quality) and unknown attack types (e.g., emerging high fidelity mask made of new materials). Thus, these methods are unreliable to be applied in practical applications with strong security needs. In light of this, more and more researchers focus on enhancing the generalization capacity of the deep FAS models. On one hand, domain adaptation and generalization techniques are leveraged for robust live/spoof classification under unlimited domain variations. On the other hand, zero/few-shot learning as well as anomaly detection frameworks are applied for unknown face PA types detection. In this paper, the unseen domains indicate the spoof-irrelated external changes (e.g., lighting and sensor noise) but actually influence the appearance quality. In contrast, the unknown spoofing attacks usually mean the novel attack types with intrinsic physical properties (e.g., material and geometry) which have not occurred in the training phase. Representative generalized deep FAS methods on unseen domains and unknown attack types are summarized in Tables-A 8 and 9 (in Appendix), respectively.

\subsubsection{Generalization to Unseen Domain}

\begin{figure}
\centering
\includegraphics[scale=0.4]{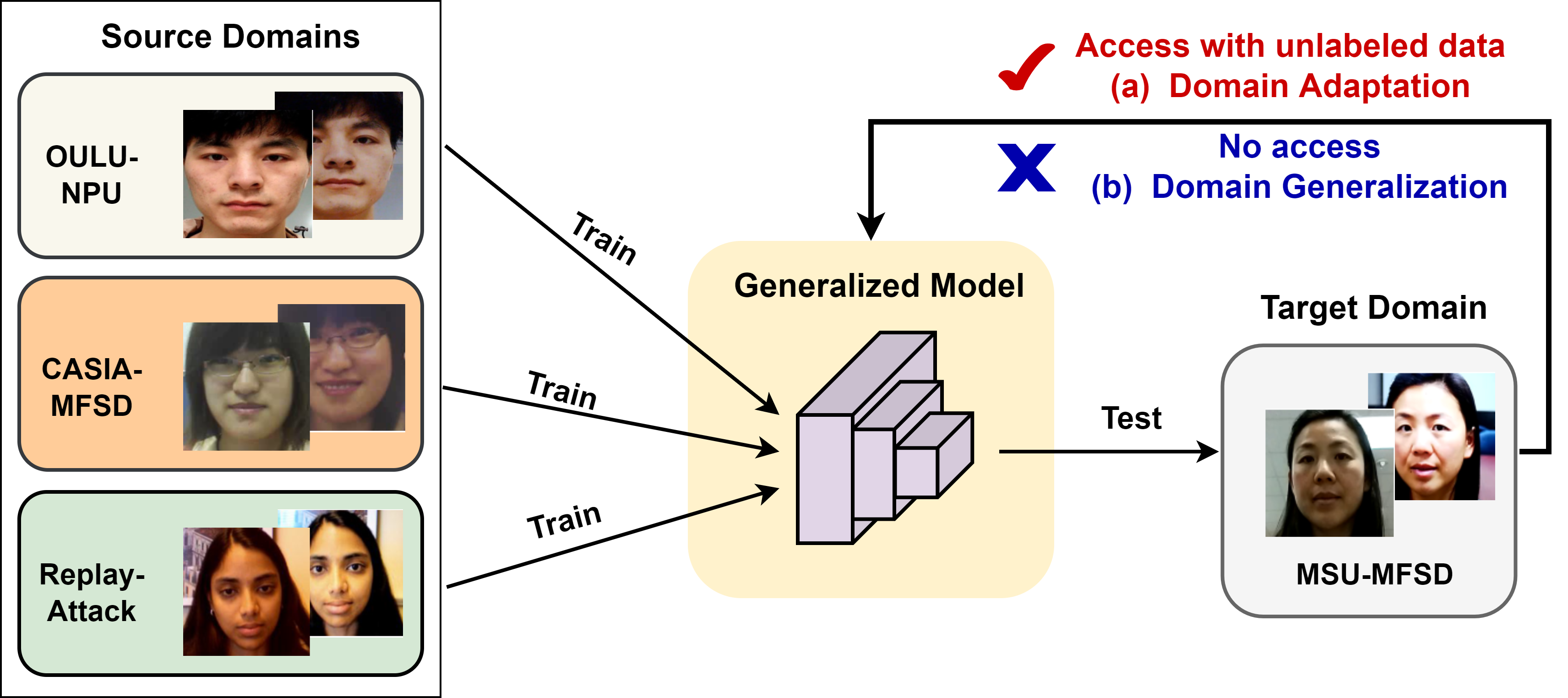}
  \caption{ 
Framework comparison among domain adaptation (DA) and domain generalization (DG). (a) The DA methods need the (unlabeled) target domain data to learn the model while (b) DG methods learn generalized model without knowledge from the target domain.}
\label{fig:domain}
\end{figure}

As shown in Fig.~\ref{fig:domain}, serious domain shifts exist among source domains and target domain, which easily leads to poor performance on biased target dataset (e.g., MSU-MFSD) when training deep models directly on sources datasets (e.g., OULU-NPU, CASIA-MFSD, and Replay-Attack). \textit{Domain adaptation} technique leverages the knowledge from target domain to bridge the gap between source and target domains. In contrast, \textit{domain generalization} helps the FAS model learn generalized feature representation from multiple source domains directly without any access to target data, which is more practical for real-world deployment. 


\vspace{0.4em}
\noindent\textbf{Domain Adaptation.}\quad    Domain adaptation technique alleviates the discrepancy between source and target domains. The distribution of source and target features are usually matched in a learned feature space. If the features share similar distributions, a classifier trained on features of the source samples can be used to classify the target samples.

To align the features space between source and target domain data, Li et al.~\cite{li2018unsupervised} propose the unsupervised domain adaptation to learn a mapping function to align the source-target embedded subspaces via minimizing their Maximum Mean Discrepancy (MMD)~\cite{gretton2012kernel}. To further enhance the generalization between both domains, UDA-Net~\cite{wang2019improving,wang2020unsupervised} is proposed with unsupervised adversarial domain adaptation in order to jointly optimize the source and target domain encoders. The domain-aware common feature space is learned when the features cannot be distinguished from both domains. The domain-invariant features constrained via MMD and adversarial domain adaptation are still with weak discrimination capacity because the label information in the target domain is inaccessible. To alleviate this problem, semi-supervised learning was introduced to domain adaptation by two works~\cite{jia2021unified,quan2021progressive}, where a few labeled data and a large amount unlabeled data in the target domain can be utilized. Jia et al.~\cite{jia2021unified} propose a unified unsupervised and semi-supervised domain adaptation network to represent the domain-invariant feature space, and find that leveraging few labeled target data (three to five) can significantly improve the performance on the target domain. Similarly, Quan et al.~\cite{quan2021progressive} propose a semi-supervised learning FAS method using only a few labeled training data for pretraining, and progressively adopt reliable unlabeled data during training to reduce the domain gap. Despite excellent adaptation, such semi-supervised methods heavily rely on class-balanced few-shot labeled data (i.e., with both live and spoof samples simultaneously), and performance degrade obviously when labeled spoof samples are unavailable.

Different from the above-mentioned methods only adapting the final classifier layer, there are a few works designing and adapting the whole FAS networks. As different deep layers share different granularities of domain information, Authors of~\cite{zhou2019face} consider multi-layer distribution adaptation on both the representation layers and the classifier layer with MMD loss. Despite efficient adaptation via multi-level clues, the architecture might be redundant and have limited generalization capacity per se. To obtain more generalized architectures, Mohammadi et al.~\cite{mohammadi2020domain} propose to prune the filters with high feature divergence that do not generalize well from one dataset to another, thus the performance of the network on target domain can be improved. Different from the network pruning in specific filters/layers, Li et al.~\cite{li2020face2} propose to distill the whole FAS model for the application-specific domain from a well-trained teacher network, which is regularized with feature MMD and pair-wise similarity embedding from both domains. In this way, lightweight yet generalized FAS models could be discovered but with weaker discrimination capacities compared with teacher FAS networks.

Although domain adaptation benefits to minimize the distribution discrepancy between the source and the target domain by utilizing unlabeled target data, in many real-world FAS scenarios, it is difficult and expensive to collect a lot of unlabeled target data (especially the spoofing attacks) for training. In addition, in consideration of the privacy issue, the source face data are usually inaccessible when deploying FAS models  on the target domain.

\vspace{0.4em}
\noindent\textbf{Domain Generalization.}\quad   
Domain generalization assumes that there exists a generalized feature space underlying the seen multiple source domains and the unseen but related target domain, on which the learned model from the seen source domains can generalize well to the unseen domain.

On one hand, a few works adopt domain-aware adversarial constraints to learn discriminative but domain-unrelated features. They assume that such features contain intrinsic clues across all seen domains thus would generalize well on unseen domain. Shao et al.~\cite{shao2019multi} are the first to propose to learn a generalized feature space shared by multiple source domains via a multi-adversarial discriminative domain generalization framework. Meanwhile, a domain generalization benchmark across four FAS datasets is also established in~\cite{shao2019multi}. However, there are two limitations: 1) such domain-independent features might still contain spoof-unrelated clues (e.g., subject ID and sensor noise); and 2) the discrimination of the domain generalized features is still unsatisfactory. To improve the first limitation, Wang et al.~\cite{wang2020cross} propose to disentangle generalized FAS features from subject discriminative and domain-dependent features. As for the second limitation, in consideration of the large distribution discrepancies among spoof faces of different domains, Jia et al.~\cite{jia2020single} propose to learn a discriminative and generalized feature space, where the feature distribution of the bonafide is compact while that of the PAs is dispersed among domains but is compact within each domain.

On the other hand, several representative works utilize domain-aware meta-learning to learn the domain generalized feature space. Specifically, faces from partial source domains are used as query set while those from remained non-overlap domains as support set. Based on this setting, Shao et al.~\cite{shao2019regularized} propose to regularize the FAS model by finding generalized learning directions in the fine-grained domain-aware meta-learning process. To alternatively force the meta-learner to perform well on support sets (domains), the learned models have robust generalization capacity. However, such domain-aware meta learning strictly needs the domain labels of the source data to construct the query and support sets but domain labels are not always available in real-world cases. Without using domain labels, Chen et al.~\cite{chen2021generalized} propose to train a generalized FAS model using the domain dynamic adjustment meta-learning, which iteratively divides mixture domains into clusters with pseudo domain labels. However, the spoof-discriminative and domain-aware features are disentangled via a simple channel attention module, making the domain features unreliable for pseudo domain label assignment. From the perspective of feature normalization, based on the evidence that instance normalization is effective to remove domain discrepancy, Liu et al.~\cite{liu2021adaptive} propose to adaptively aggregate batch and instance normalizations for generalized representation via meta-learning. Note that the adaptive tradeoffs between batch and instance normalizations might weaken the live/spoof discrimination capacity.

Overall, domain generalization for FAS is a new hot spot in recent three years, and some potential and exciting trends such as combining domain generalization with adaptation~\cite{wang2021self}, and learning without domain labels~\cite{chen2021generalized} are investigated. However, there still lacks of the works lifting the veil about discrimination and generalization capacities. In other words, domain generalization benefits FAS models to perform well in unseen domain, but it is still unknown whether it deteriorates the discrimination capability for spoofing detection under the seen scenarios.



\subsubsection{Generalization to Unknown Attack Types}

Besides domain shift issues, FAS models are vulnerable to emerging novel PAs in real-world practical applications. Most previous deep learning methods formulate FAS as a close-set problem to detect various pre-defined PAs, which need large-scale training data to cover as many attacks as possible. However, the trained model can easily overfit several common attacks (e.g., print and replay) and is still vulnerable to unknown attack types (e.g., mask and makeup). Recently, many researches focus on developing generalized FAS models for unknown spoofing attack detection. On one side, \textit{zero/few-shot learning} is employed for improving novel spoofing detection with very few or even none samples of target attack types. On the other side, FAS can also be treated as a \textit{one-class classification} task where the bonafide samples are clustered compactly, and \textit{anomaly detection} is used for detecting the out-of-distribution PA samples.

\vspace{0.4em}
\noindent\textbf{Zero/Few-Shot Learning.}\quad   
One straightforward way for novel attack detection is to finetune the FAS model with sufficient samples of the new attacks. However, collecting labeled data for every new attack is expensive and time-consuming since the spoofing keeps evolving. To overcome this challenge, several works~\cite{liu2019deep,qin2019learning,perez2020learning} propose to treat FAS as an open-set zero- and few-shot learning problem. \textit{Zero-shot learning} aims to learn generalized and discriminative features from the predefined PAs for unknown novel PA detection. \textit{Few-shot learning} aims to quickly adapt the FAS model to new attacks by learning from both the predefined PAs and the collected very few samples of the new attack.

Without any prior knowledge of the unknown spoof attacks, Liu et al.~\cite{liu2019deep} design a Deep Tree Network (DTN) to learn the semantic attributes of pre-defined attacks and partition the spoof samples into semantic sub-groups in an unsupervised fashion. Based on the similarity of the embedding features, DTN adaptively routes the known or unknown PAs to the corresponding spoof clusters. The live/spoof tree topology is constructed via DTN automatically, which is more semantic and generalized compared with the human-defined category relationship. However, without any prior knowledge of unknown attacks, the zero-shot DTN may fail to detect novel high-fidelity attacks. To alleviate this issue, two works adopt an open-set few-shot framework to introduce partial yet effective unknown attack knowledge for representation learning. Qin et al.~\cite{qin2019learning} unify the zero- and few-shot FAS tasks together by fusion training a meta-learner with an adaptive inner-update learning rate strategy. Training meta-learner on both zero- and few-shot tasks simultaneously enhances the discrimination and generalization capacities of FAS models from pre-defined PAs and few instances of the new PAs. However, directly using few-shot meta learning on novel attacks easily suffers from catastrophic forgetting about the pre-defined PAs. To tackle this issue, Perez-Cabo et al.~\cite{perez2020learning} propose a continual few-shot learning paradigm, which incrementally extends the acquired knowledge from the continuous stream of data, and detects new PAs using a small number of training samples via a meta-learning solution.

Although few-shot learning benefits the FAS models for unknown attack detection, the performance drops obviously when the data of the target attack types are unavailable for adaptation (i.e., zero-shot case). We observe that the failed detection usually occurs in the challenging attack types (e.g., transparent mask, funny eye, and makeup), which share similar appearance distribution with the bonafide.

\vspace{0.4em}
\noindent\textbf{Anomaly Detection.}\quad   
Anomaly detection based FAS assumes that the live samples are in a normal class as they share more similar and compact feature representation while features from the spoof samples have large distribution discrepancies in the anomalous sample space due to the high variance of attack types and materials. Based on the assumption, anomaly detection usually firstly trains a reliable one-class classifier to accurately cluster the live samples. Then any samples (e.g., unknown attacks) outside the margin of the live sample cluster would be detected as attacks.

Arashloo et al.~\cite{arashloo2017anomaly} is the first to evaluate one-class anomaly detection and traditional binary classification FAS systems on cross-type testing protocols. They find that anomaly-based methods using one-class SVM are not inferior compared to binary classification approaches using two-class SVM. To better represent the probability distribution of bonafide samples, Nikisins et al.~\cite{nikisins2018effectiveness} propose to replace traditional one-class SVM with Gaussian Mixture Model (GMM) as the anomaly detector. Besides one-class SVM and GMM, Xiong and AbdAlmageed~\cite{xiong2018unknown} also consider the autoencoder based outliers detector with LBP feature extractor for open-set unknown PAD. The above-mentioned works separate the feature extraction with the one-class classifier, which makes the bonafide representation learning challenging and sub-optimal. In contrast, Baweja et al.~\cite{baweja2020anomaly} present an end-to-end anomaly detection approach to train the one-class classifier and feature representations together. Moreover, to learn robust bonafide representation against out-of-distribution perturbations, they generate pseudo negative features to mimic the PA class and force the one-class classifier to be discriminative for PAD. However, the generated pseudo PA features cannot represent diverse real-world features, making the one-class anomaly detection system less reliable for real-world deployment.

Though reasonable, utilizing only live faces to train the classifier usually limits the anomaly detection model's generalization on new PA types. Instead of using only live samples, some works train the generalized anomaly detection systems with both live and spoof samples via metric learning. P{\'e}rez-Cabo et al.~\cite{perez2019deep} propose to regularize the FAS model by a triplet focal loss to learn discriminative feature representation, and then introduce a few-shot posteriori probability estimation as anomaly detector for unknown PA detection. Similarly, George and Marcel~\cite{george2020learning} design a pair-wise one-class contrastive loss (OCCL) to force the network to learn a compact embedding for bonafide class while being far from the representation of attacks. Then an one-class GMM is cascaded for unknown PA detection. Although discriminative embedding could be learned via triplet or contrastive loss, the works~\cite{perez2019deep,george2020learning} still need extra anomaly detectors (e.g., one-class GMM) cascaded after embedding features, which influences the end-to-end representation learning. In contrast, Li et al.~\cite{li2020unseen} propose to supervise deep FAS models with a novel hypersphere loss to keep the intra-class live compactness as well as inter-class live-spoof separation. The unknown attacks could be directly detected on learned feature space with no need of additional anomaly detection classifiers. One limitation is that the predicted live/spoof score is calculated from the square of L2-norm of the embedding features, which is hard to select a suitable predefined threshold for detecting different kinds of attack types.

Despite satisfactory generalization capacity for unknown attack detection, anomaly detection based FAS methods would suffer from discrimination degradation compared with conventional live/spoof classification in the real-world open-set scenarios (i.e., both known and unknown attacks).

\section{Deep FAS with Advanced Sensors} \label{sec:multimodal}


Commercial RGB camera-based FAS would be an excellent tradeoff solution in terms of security and hardware cost in daily face recognition applications. However, some high-security scenarios (face payment and vault entrance guard) require very low false acceptance errors. Recently, advanced sensors with various modalities are developed to facilitate the ultra-secure FAS. Merits and demerits of various sensors and hardware modules for FAS in terms of environmental conditions (lighting and distance) and attack types (print, replay, and 3D mask) are listed in Table~\ref{tab:sensorstemp}.

Compared with monocular visible RGB camera (VIS), stereo cameras (VIS-Stereo)~\cite{rehman2020slnet} benefits the 3D geometry information reconstruction for 2D spoofing detection. When assembling with dynamic flash light on the presentation face, VIS-Flash~\cite{liu2019AuroraGuard} is able to capture intrinsic reflection-based material clues to detect all three attack types.

Besides visible RGB modality, depth and NIR modalities are also widely used in practical FAS deployment with acceptable costs. Two kinds of depth sensors including Time of Flight (TOF)~\cite{wu2019review} and 3D Structured Light (SL)~\cite{connell2013fake} have been embedded in mainstream mobile phone platforms (e.g., Iphone, Sumsung, OPPO, and Huawei). They provide the accurate 3D depth distribution of the captured face for 2D spoofing detection. Compared with SL, TOF is more robust to environmental conditions such as lighting and distance. In contrast, NIR~\cite{sun2016context} modality is a complementary spectrum (900 to 1800nm) besides VIS, which effectively exploits reflection differences between live and spoof faces but is with poor imaging quality in long distance. In addition, the VIS-NIR integration hardware module is with a high performance-price ratio for many access control systems.

Meanwhile, several niche but effective sensors are introduced in FAS. Shortwave infrared (SWIR)~\cite{heusch2020deep} with the wavelengths of 940nm and 1450nm bands discriminates live skin material from non-skin pixels in face images via measuring water absorption, which is reliable for generic spoofing attacks detection. A thermal camera~\cite{seo2019face} is an alternative sensor for efficient FAS via face temperature estimation. However, it performs poorly when subjects wear transparent masks. Expensive Light Field camera~\cite{liu2019light} and four-directional Polarization sensor~\cite{tian2020face} are also explored for FAS according to their excellent representation for facial depth and reflection/refraction light, respectively.

\begin{table}[t]
\centering
\caption{Comparison with sensor/hardware for FAS under 2 environments (lighting condition and distance) and 3 attack types (print, replay and 3D mask). `TOF', `SL', `C', 'M', 'E', `P', 'G', 'VG' are short for `Time of Flight', `
Structured Light', `Cheap', `Medium', `Expensive', `Poor', `Good', `Very Good', respectively.}

\vspace{-0.5em}

\resizebox{0.49\textwidth}{!}{
\begin{tabular}{|c|c|c|c|c|c|c|}

\hline
\multirow{2}{*}{\textbf{Sensor}} & \multirow{2}{*}{\textbf{Cost}} &\multicolumn{2}{c|}{\textbf{Environment}} &\multicolumn{3}{c|}{\textbf{Attack Type}}\\
\cline{3-7} 
& & \tabincell{c}{Lighting} &\tabincell{c}{Distance} &\tabincell{c}{Print} &\tabincell{c}{Replay}&\tabincell{c}{Mask}\\

\hline

VIS & C & M & M & M & M & M \\
VIS-Stereo & M & M & M & VG & VG & M \\
VIS-Flash & C & M & M & G & G & M \\
Depth(TOF) & M & M & G & VG & VG & P \\
Depth(SL) & C & P & P & VG & VG & P \\
NIR & C & G & P & G & VG & M \\
VIS-NIR & M & G & M & G & VG & G \\
SWIR & E & G & M & VG & VG & G \\
Thermal & E & G & M & G & VG & M \\
Light Field & E & P & M & VG & VG & M \\
Polarization & E & G & M & VG & VG & G \\

\hline

\end{tabular}
}
\label{tab:sensorstemp}
\vspace{-0.5em}
\end{table}


\subsection{Uni-Modal Deep Learning upon Specialized Sensor}
Based on the specialized sensor/hardware for distinct imaging, researchers have developed sensor-aware deep learning methods for efficient FAS, which are summarized in Table-A 10 (in Appendix). Seo and Chung~\cite{seo2019face} propose a lightweight Thermal Face-CNN to estimate the facial temperature from the thermal image, and detect the spoofing with abnormal temperature (e.g., out of scope from 36 to 37 degrees). They find that the thermal image is more suitable than the RGB image for replay attack detection. However, such thermal-based method is vulnerable to the transparent mask attack. In terms of stereo-based FAS, several works~\cite{rehman2020slnet,kang2021facial,wu2020single} prove that leveraging the estimated disparity or depth/normal maps from the stereo inputs (from stereo and dual pixel (DP) sensors) via CNN could achieve remarkable performance on 2D print and replay attack detection. However, it usually performs poorly on the 3D mask attack with similar geometric distribution of live faces. To further capture detailed 3D local patterns, Liu et al.~\cite{liu2019light} propose to extract the ray difference and microlens images from a single-shot light field camera, and then a shallow CNN is used for face PAD. Due to the rich 3D information in light field imaging, the method is potential to classify fine-grained spoofing types. Towards real-time and mobile-level deployment, Tian et al.~\cite{tian2020face} propose to use lightweight MobileNetV2 to extract efficient DOLP features from an on-chip integrated polarization imaging sensor. The above-mentioned methods aim at tackling specific PA types (e.g., replay and print), which cannot generalize well across all PA types. In contrast, Heusch et al.~\cite{heusch2020deep} propose to use a multi-channel CNN for deep material-related feature extraction from the selected SWIR-difference inputs, which is able to almost perfectly detect all impersonation attacks while ensuring low bonafide classification errors.

Apart from using specialized hardware such as infrared dot projectors and dedicated cameras, some deep FAS methods are developed based on visible cameras with extra environmental flash. In~\cite{liu2019AuroraGuard} and ~\cite{farrukh2020facerevelio}, dynamic flash from the smartphone screen is utilized to illuminate a user’s face from multiple directions, which enables the recovery of the face surface normals via photometric stereo. Such dynamic normal cues are then fed into CNN to predict facial depth and light CAPTCHA for PA detection. Similarly, Ebihara et al.~\cite{ebihara2019specular} design a novel descriptor to represent the specular and diffuse reflections leveraging the difference cues with and without flash, which outperforms the end-to-end ResNet with concatenated flash inputs. These methods are easy to deploy without extra hardware integration, and have been used in mobile verification and payment systems such as Alipay and WeChat Pay. However, dynamic flash is sensitive under outdoor environments and is not user-friendly due to the long temporal activation time.


\subsection{Multi-Modal Deep Learning}
Meanwhile, multi-modal learning based methods become hot and active in the FAS research community. Representative multi-modal fusion and cross-modal translation approaches for FAS are collected in Table-A 11 (in Appendix).


\vspace{0.4em}

\noindent\textbf{Multi-Modal Fusion.}\quad With multi-modal inputs, mainstream FAS methods extract complementary multi-modal features using feature-level fusion strategies. 
As there are redundancy across multi-modal features, direct feature concatenation easily results in high-dimensional features and overiftting. To alleviate this issue, Zhang et al.~\cite{zhang2020casia} propose the SD-Net using a feature re-weighting mechanism to select the informative and discard the redundant channel features among RGB, depth, and NIR modalities. However, the re-weighting fusion in SD-Net is only conducted on the high-level features but neglecting the multi-modal low-level clues. To further boost the multi-modal feature interaction at different levels, authors from~\cite{parkin2019recognizing} and~\cite{kuang2019multi} introduce a multi-modal multi-layer fusion branch to enhance the contextual clues among modalities. Despite advanced fusion strategies, multi-modal fusion is easily dominated by partial modalities (e.g., depth) thus performs poorly when these modalities are noisy or missing. To tackle this issue, Shen et al.~\cite{shen2019facebagnet} design a Modal Feature Erasing operation to randomly dropout partial-modal features to prevent modality-aware overfitting. In addition, George and Marcel~\cite{george2021cross} present a cross-modal focal loss to modulate the loss contribution of each modality, which benefits the model to learn complementary information among modalities. Overall, feature-level fusion is flexible and effective for multi-modal clue aggregation. However, modality features are usually extracted from separate branches with high computational cost. 


Besides feature-level fusion, there are a few works that consider input-level and decision-level fusions. Input-level fusion assumes that multi-modal inputs are already aligned spatially, and can be fused in the channel dimension directly. In~\cite{nikisins2019domain}, the composite image is fused from gray-scale, depth, and NIR modalities by stacking the normalized images, and then fed to deep PA detectors. Similarly, Liu et al.~\cite{liu2021data} composite VIS-NIR inputs via different fusion operators (i.e., stack, summation, and difference), and all fused face images are forwarded by a multi-modal FAS network for live/spoof prediction. These input-level fusion methods are efficient and with a little extra computational cost (mostly on fusion operator and the first network layer). However, fusion in too early stage easily vanishes multi-modal clues in the subsequent mid- and high-level spaces. In contrast, to tradeoff the individual modality bias and make reliable binary decision, some works adopt decision-level fusion based on the predicted score from each modality branch. On one hand, Yu et al.~\cite{yu2020multi} directly average the predicted binary scores of individual models from RGB, depth, and NIR modalities, which outperforms the input- and feature-level fusions on CeFA~\cite{li2020casia} dataset. On the other hand, Zhang et al.~\cite{zhang2019feathernets} design a decision-level fusion strategy to firstly aggregate scores from several models using depth modality, and then cascaded with the score from the IR model for final live/spoof classification. Despite reliable prediction, decision-level fusion is inefficient as it needs separate well-trained models for particular modalities. 

\vspace{0.4em}

\noindent\textbf{Cross-Modal Translation.}\quad
Multi-modal FAS system needs additional sensors for imaging face inputs with different modalities. However, in some conventional scenarios, only partial modalities (e.g., RGB) can be available. To tackle this modality missing issues at the inference stage, a few works adopt the cross-modal translation technique to generate the missing modal data for multi-modal FAS. To generate the corresponding NIR images from RGB face images, Jiang et al.~\cite{jiang2020face} first propose a novel multiple categories (live/spoof, genuine/synthetic) image translation cycle-GAN. Based on the generated NIR and original RGB inputs, the method is able to extract more robust fused features compared with using only the RGB images. However, the generated NIR images from raw cycle-GAN are with low quality, which limits the performance of the fused features. To generate high-fidelity target NIR modality, Liu et al.~\cite{liu2021face} design a novel subspace-based modality regularization in the cross-modal translation framework. Besides generating the NIR images, Mallat and Dugelay~\cite{mallatindirect2021} propose a visible-to-thermal conversion scheme to synthesize thermal attacks from RGB face images using a cascaded refinement network. Though effectiveness on intra-dataset testings, one main concern of these methods is that the domain shifts and unknown attacks might significantly influence the generated modality's quality, and the fused features would be unreliable using paired noisy modality data.

Despite a rising trend since 2019, the progress of sensor-based multi-modal FAS is still slow compared with RGB based unimodal methods. It is worth noting that multi-modal approaches also exist in deep FAS with commercial RGB camera. For instance, decision-level fusion of two RGB video based modalities (i.e., remote physiological signals and face visual image) has been explored in~\cite{lin2019face}. Therefore, to effectively fuse such natural modalities from commercial RGB camera with those from advanced sensors will be an interesting and valuable direction. Meanwhile, some advanced sensors (e.g., SWIR, light field, and polarization) are expensive and non-portable for real-world deployment. More efficient FAS-dedicated sensors as well as multi-modal approaches should be explored.

\vspace{-0.3em}

\section{Discussion and Future Directions} \label{sec:discussion}

Thanks to the recent advances in deep learning, FAS has achieved rapid improvement over the past few years. As can be seen from Fig.~\ref{fig:protocols}, recent deep FAS methods refresh the state of the arts and obtain satisfied performance (e.g., $\textless$5\% ACER, $\textless$15\% HTER, $\textless$10\% EER, and $\textless$20\% HTER) on four evaluation protocols, respectively. On one hand, advanced architectures (e.g., NAS-FAS~\cite{yu2020fas2} and FAS-SGTD~\cite{wang2020deep}) and pixel-wise supervision (e.g., pseudo depth and reflection maps) benefit the 2D attack detection as well as the fine-grained spoof material perception (e.g., silicone and transparent 3D masks). On the other hand, domain and attack generalization based methods (e.g., SSDG~\cite{jia2020single}, FGHV~\cite{liu2022feature}, and SSAN~\cite{wang2022domain}) mine the intrinsic live/spoof clues across multiple source domains and attack types, which can generalize well even on unseen domains and unknown attacks. These generalized deep learning based methods usually detect different kinds of attacks (2D \& 3D) under diverse scenarios more stably (with lower standard deviation errors) under leave-one-out cross-testing protocols. Furthermore, some insightful conclusions could be drawn from Tables-A 2, 3, and 4 (in Appendix): 1) Advanced architectures (e.g., DC-CDN~\cite{yu2021dual}) with elaborate supervisions (e.g., pseudo depth supervision) dominate the testing performance when training on single source domain. In contrast, when training on multiple (three) domains, generalized learning strategies play more important roles. 2) Transfer learning from large-scale pre-trained models (e.g., SSAN~\cite{wang2022domain} and ViTranZFAS~\cite{liu2019deep} using ResNet18 and vision transformer pretrained from ImageNet1K and ImageNet21K, respectively) alleviates the overfitting issue caused by limited-scale live/spoof data, thus improves the generalization capacity and benefits cross-dataset and cross-type testings.

However, FAS is still an unsolved problem due to the challenges such as subtle spoof pattern representation, complex real-world domain gaps, and rapidly iterative novel attacks. We conclude the limitations of the current development as follows: 1) \textit{Limited live/spoof representation capacity with sub-optimal deep architectures, supervisions, and learning strategies}. Learning discriminative and generalized live/spoof features is vital for deep FAS. Until now, it is still hard to find the best-suited architectures as well as the supervisions across all different evaluation benchmarks. For example, CDCN with pixel-wise supervision achieves excellent and poor performance on intra-dataset and multi-source-domain cross-dataset testings, respectively, while ResNet with binary cross-entropy loss performs inversely. 2) \textit{Evaluation under saturating and unpractical testing benchmarks and protocols}. For example, for intra-testing on the OULU-NPU dataset, ACER of 0.4\% and 0.8\% might make slight difference and indicate the performance saturation on such small-scale and monotonous test set. And the cross-domain testings are still far from real-world scenarios as only limited sorts of attack types are considered. 3) \textit{Isolating the anti-spoofing task on only the face area and physical attacks}. Besides physical presentation attacks in the face area, spoofing in more general applications (e.g., commodity and document) and even digital attacks via stronger and stronger face swapping and generative models should be considered. These tasks might share partial intrinsic knowledge and benefit the representation learning. 4) \textit{Insufficient consideration about the interpretability and privacy issues}. Most existing FAS researches devote to developing novel algorithms 
against state-of-the-art performance but rarely think about the interpretability behind. Such black-box methods are hard to make reliable decisions in real-world cases. In addition, most existing works train and adapt deep FAS models with huge stored source face data, and neglect the privacy and biometric sensitivity issue. According to the discussion above, we summarize some solutions and potential research directions in the following subsections.


\vspace{-0.3em}
\subsection{Architecture, Supervision and Interpretability}
As can be seen from Sections~\ref{sec:RGB} and~\ref{sec:multimodal}, most of the researchers choose the off-the-shelf network architectures as well as handcrafted supervision signals for deep FAS, which might be sub-optimal and hard to leverage the large-scale training data adequately. Although several recent works have applied AutoML in FAS for searching well-suited architecture~\cite{yu2020searching,yu2020fas2}, loss function~\cite{qin2020one}, and auxiliary supervision~\cite{qin2021meta}, they focus on uni-modality and single-frame configuration while neglecting the temporal or multi-modal situation. Hence, one promising direction is to automatically search and find the best-suited temporal architectures especially for multi-modal usage. In this way, more reasonable fusion strategies would be discovered among modalities instead of coarse handcrafted design. In addition, rich temporal context should be considered in dynamic supervision design instead of static binary or pixel-wise supervision.  

On the other hand, to design \textit{efficient} network architecture is vital for real-time FAS in mobile devices. Over the past years, most research focuses on tackling the accuracy and generalization issues in FAS while only a few works consider lightweight~\cite{yu2020auto2} or distilled~\cite{li2020face2} CNNs for efficient deployment. Besides CNN with strong inductive bias, researchers should also rethink the usage of some flexible architectures (e.g., vision transformer~\cite{yu2021transrppg,george2020effectiveness}) in terms of efficiency and computational cost.

Recently, great efforts have been achieved on \textit{interpretable} FAS~\cite{sequeiraexploratory}. Some methods try to localize the spoof regions according to the feature activation using visual interpretability tools (e.g., Grad-CAM~\cite{selvaraju2017grad}) or soft-gating strategy~\cite{deb2020look}. In addition, auxiliary supervised~\cite{Liu2018Learning,yu2020face} and generative~\cite{jourabloo2018face,liu2020physics} FAS models devote to estimating the underlying spoof maps. Besides the visual activation maps, natural language~\cite{mirzaalian2021explaining} has been introduced for explaining the FAS predictions with meaningful sentence-level descriptions. All these trials help researchers understand and localize the spoof patterns, and convince the FAS decision. However, due to the lack of precious pixel-level spoof annotation, the estimated spoof maps are still coarse and easily influenced by unfaithful clues (e.g., hands). More advanced feature visualization manners and fine-grained pixel-wise spoof segmentation should be developed for interpretable FAS.

\vspace{-0.3em}
\subsection{Representation Learning}
Learning discriminative and intrinsic feature representation is the key to reliable FAS. A handful of previous researches have proven the effectiveness of transfer learning~\cite{lucena2017transfer,parkin2019recognizing} and disentangled learning~\cite{zhang2020face,liu2020physics} for FAS. The former leverages the pre-trained semantic features from other large-scale datasets to alleviate the overfitting issue, while the latter aims to disentangle the intrinsic spoofing clues from the noisy representation. To learn discriminative embedding spaces with compact distributions among live faces and distinguishable distances between live/spoof faces, deep metric learning is used for training FAS models. However, the uncertainty of the model prediction is still high in the extreme/noisy scenario (e.g., presenting with very high-quality spoof and low-quality live samples). More advanced metric learning techniques (e.g., on hyperbolic manifold space) could be explored in the future for mining subtle spoof patterns. Moreover, rephrasing FAS as a fine-grained recognition~\cite{yu2020face,wang2022patchnet} problem to learn type-discriminative representation is worth exploring, which is inspired by the fact that humans could detect spoofing via recognizing the specific attack types.  

Researchers should also get hung up on fully exploiting the live/spoof training data with or without labels for representation enhancement. On one side, self-supervised on large-scale combined datasets might reduce the risk of overfitting, and actively mine the intrinsic knowledge (e.g., high similarity among intra face patches). On the other side, in real-world scenarios, daily unlabeled face data are collected from various face recognition terminals continuously, which could be utilized for semi-supervised learning~\cite{quan2021progressive}. One challenge is how to make full use of the unlabeled imbalanced (i.e., live $\gg$ spoof) data, avoiding unexpected performance drop. In addition, suitable data augmentation strategies~\cite{yu2021dual} for FAS are rarely investigated. Adversarial learning might be a good choice for adaptive data augmentation in more diverse domains.

\vspace{-0.5em}
\subsection{Real-World Open-Set FAS}
As discussed in Section~\ref{sec:protocols}, traditional FAS evaluation protocols usually consider intra-domain~\cite{Boulkenafet2017OULU}, cross-domain~\cite{shao2019multi}, and cross-type~\cite{liu2019deep} testings within one or several small-scale datasets. The state-of-the-art methods in such protocols cannot guarantee consistently good performance in practical scenarios because 1) the data amount (especially testing set) is relatively small thus the high performance is not very convincing; and 2) the protocols focus on a single factor (e.g., seen/unseen domains or known/unknown attack types), which cannot satisfy the need of complex real-world scenarios. Recently, more practical protocols such as GrandTest~\cite{perez2020learning} and open-set~\cite{liu2021contrastive,liu2020physics} are proposed. GrandTest contains large-scale mixed-domain data, while open-set testing considers models' discrimination and generalization capacities on both known and unknown attack types. However, real-world open-set situations with simultaneous domains and attack types are still neglected. More comprehensive protocols (e.g., domain- and type-aware open-set) should be explored for fair and practical evaluation to bridge the gap between academia and industry.

As for the multi-modal protocols, training data with multiple modalities are assumed available, and two testing settings are widely used: 1) with corresponding multiple modalities~\cite{liu2020cross}; and 2) only single modality~\cite{george2021cross,liu2021face} (usually RGB). However, there are various kinds of modality combinations~\cite{yu2022flexible} (e.g., RGB-NIR, RGB-D, NIR-D, and RGB-D-NIR) in real-world deployment according to different user terminal devices. Therefore, it is pretty costly and inefficient to train individual models for each multi-modal combination. Although pseudo modalities could be generated via cross-modality translation~\cite{jiang2020face,liu2021face}, their fidelity and stability are still weaker compared with modalities from real-world sensors. To design a dynamic multi-modal framework to propagate the learned multi-modal knowledge to various modality combinations might be a possible direction for unlimited multi-modal deployment.

\vspace{-0.5em}
\subsection{Generic and Unified PA Detection}
Understanding the intrinsic property of face PAD with other related tasks (e.g., generic PAD, and digital face attack detection) is important for explainable FAS. On one hand, `generic' assumes that both face and other object presentation attacks might have independent content but share intrinsic spoofing patterns~\cite{stehouwer2020noise}. For instance, replay attacks about different objects (e.g., a face and a football) are made of the same glass material~\cite{yu2020face}, and with abnormal reflection clues. Thus, generic PAD and material recognition datasets could be introduced in face PAD for common live/spoof feature representation in a multi-task learning fashion.

\begin{figure}
\centering
\includegraphics[scale=0.48]{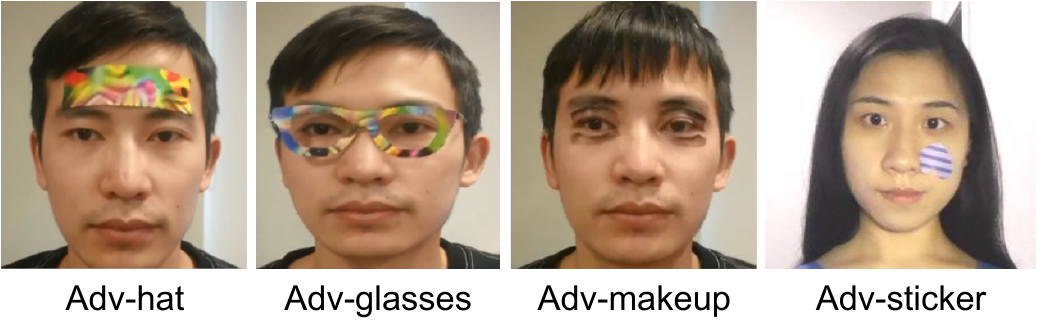}
\vspace{-0.6em}
  \caption{ 
   Illustration of the physical adversarial faces generated by Adv-glasses~\cite{sharif2019general}, Adv-hat~\cite{komkov2021advhat}, Adv-makeup~\cite{yin2021adv}, and Adv-sticker~\cite{guo2021meaningful}.
  }
  \vspace{-0.9em}
\label{fig:adv}
\end{figure}

Apart from common PAs, two kinds of physical adversarial attacks (AFR-aware and FAS-aware) should be considered for generic PAD. As illustrated in Fig.~\ref{fig:adv}, physical  eyeglass~\cite{sharif2019general} and hat~\cite{komkov2021advhat} achieved from adversarial generators, or
special stickers~\cite{guo2021meaningful} containing feature patterns proved to be effective against deep learning based AFR systems can be printed out and wore by attackers to spoof such systems. Moreover, imperceptible makeup~\cite{yin2021adv} nearby the eye regions have been verified for attacking commercial AFR systems. Besides AFR-aware adversarial attacks, adversarial print/replay attacks~\cite{zhang2019attacking} with perturbation before physical broadcast are developed to fool the FAS system. Therefore, it is expected and necessary to establish large-scale FAS datasets with diverse physical adversarial attacks as well as annotated attack localization labels.

On the other hand, besides physical face presentation attacks, there are many vicious digital manipulation attacks (e.g., Deepfake~\cite{ciftci2020fakecatcher}) and morphing attacks (e.g., via generative model StyleGAN\cite{sarkar2022gan}) on face videos. As generative models become stronger and stronger, these direct digital attacks from generative models become bigger threats. Despite different generation manners with diverse attack traces and visual qualities, parts of these attacks might still have coherent properties. In~\cite{deb2021unified,yu2022benchmarking}, a unified digital and physical face attack detection framework is proposed to learn joint representations for coherent attacks. However, there are serious imbalanced numbers among digital and physical attack types due to data collection costs. In other words, large-scale digital attacks are easier to generate compared with high-cost presentation attacks. Such imbalanced distribution might harm the intrinsic representation during the multi-task learning, which needs to think about in the future. 


\vspace{-0.5em}
\subsection{ Privacy-Preserved Training}

Leveraging large-scale live/spoof face data, deep learning based FAS has achieved huge breakthroughs. However, the legal and privacy issues of the face data attract more and more attention. For example, the GDPR (General Data Protection Regulation)~\cite{voigt2017eu}, came into effect in May 2018, brings the importance of preserving the privacy of personal information (e.g., face images) to the forefront. Therefore, a noteworthy direction is to alleviate the privacy issue (i.e., storing/sharing large-scale users' face data) but maintaining satisfied performance for deep FAS models. 

On one hand, the live/spoof face training data are usually not directly shared between data owners (domains). To tackle this challenge, federate learning~\cite{mcmahan2017communication}, a distributed and privacy-preserving machine learning technique, is introduced in FAS to simultaneously take advantage of rich live/spoof information available at different data owners while maintaining data privacy. To be specific, each data center/owner locally trains its own FAS model. Then a server learns a global FAS model by iteratively aggregating model updates from all data centers without accessing original private data in each of them. Finally, the converged global FAS model would be utilized for inference. To enhance the generalization ability of the server model, in~\cite{shao2021federated}, a federated domain disentanglement strategy is introduced, which treats each data center as one domain and decomposes the FAS model into domain-invariant and domain-specific parts in each data center. Overall, the existing federated learning based FAS usually focuses on the privacy problem of \textit{data} sets but neglects the privacy issues in the \textit{model} level. Thus, the training of the global model needs multiple teams to share their own local models, which might harm the commercial competition.

On the other hand, due to privacy and security concerns of human faces, source data are usually inaccessible during adaptation for practical deployment. Specifically, in a source-free~\cite{kundu2020universal} setting, a FAS model is first pre-trained on the (large-scale) source data and is released for deployment. In the deployment phase, the source data cannot be shared for adapting the pre-trained model to the target data, as they contain sensitive biometric information. Lv et al.~\cite{lv2021combining} benchmark the source-free setting for FAS via directly applying a self-training approach, which easily obtains noisy target pseudo labels due to the challenges in the FAS task (e.g., the intra-class distance between live faces of different identities probably exceeds the inter-class distance between live and spoof faces of the same identity). Thus, the performance gain (1.9\% HTER reduction on average) by adaptation is quite limited. To efficiently and accurately adapt the source knowledge without accessing source data is worth exploring in the future.

\vspace{-0.5em}
\section{Conclusion} \label{sec:conclusion}
This paper has presented a contemporary survey of the deep learning based methods, datasets as well as protocols for face anti-spoofing (FAS). A comprehensive taxonomy of these methods have been presented. Merits and demerits of various methods and sensors for FAS are also covered, with potential research directions being listed.

\vspace{0.5em}

\noindent\textbf{Acknowledgments} \quad This work was supported by the Academy of Finland (Academy Professor project EmotionAI with grant numbers 336116 and 345122, and ICT2023 project with grant number 345948), the National Natural Science Foundation of China (No. 61876178, 61976229, and 62106264), and Beijing Academy of Artificial Intelligence (BAAI).



\ifCLASSOPTIONcaptionsoff
  \newpage
\fi



%
\bibliographystyle{IEEEtran}
\bibliography{IEEEabrv,reference}

\begin{thebibliography}{100}
\providecommand{\url}[1]{#1}
\csname url@samestyle\endcsname
\providecommand{\newblock}{\relax}
\providecommand{\bibinfo}[2]{#2}
\providecommand{\BIBentrySTDinterwordspacing}{\spaceskip=0pt\relax}
\providecommand{\BIBentryALTinterwordstretchfactor}{4}
\providecommand{\BIBentryALTinterwordspacing}{\spaceskip=\fontdimen2\font plus
\BIBentryALTinterwordstretchfactor\fontdimen3\font minus
  \fontdimen4\font\relax}
\providecommand{\BIBforeignlanguage}[2]{{%
\expandafter\ifx\csname l@#1\endcsname\relax
\typeout{** WARNING: IEEEtran.bst: No hyphenation pattern has been}%
\typeout{** loaded for the language `#1'. Using the pattern for}%
\typeout{** the default language instead.}%
\else
\language=\csname l@#1\endcsname
\fi
#2}}
\providecommand{\BIBdecl}{\relax}
\BIBdecl

\bibitem{guo2020learning}
J.~Guo, X.~Zhu, C.~Zhao, D.~Cao, Z.~Lei, and S.~Z. Li, ``Learning meta face
  recognition in unseen domains,'' in \emph{CVPR}, 2020.

\bibitem{pan2007eye}
G.~Pan, L.~Sun, Z.~Wu, and S.~Lao, ``Eyeblink-based anti-spoofing in face
  recognition from a generic webcamera,'' in \emph{ICCV}, 2007.

\bibitem{li2016generalized}
X.~Li, J.~Komulainen, G.~Zhao, P.-C. Yuen, and M.~Pietik{\"a}inen,
  ``Generalized face anti-spoofing by detecting pulse from face videos,'' in
  \emph{ICPR}, 2016.

\bibitem{Pereira2012LBP}
T.~de~Freitas~Pereira, A.~Anjos, J.~M. De~Martino, and S.~Marcel, ``Lbp- top
  based countermeasure against face spoofing attacks,'' in \emph{ACCV}, 2012.

\bibitem{Komulainen2014Context}
J.~Komulainen, A.~Hadid, and M.~Pietikainen, ``Context based face
  anti-spoofing,'' in \emph{BTAS}, 2013.

\bibitem{Patel2016Secure}
K.~Patel, H.~Han, and A.~K. Jain, ``Secure face unlock: Spoof detection on
  smartphones,'' \emph{TIFS}, 2016.

\bibitem{jee2006liveness}
H.-K. Jee, S.-U. Jung, and J.-H. Yoo, ``Liveness detection for embedded face
  recognition system,'' \emph{International Journal of Biological and Medical
  Sciences}, 2006.

\bibitem{li2008eye}
J.-W. Li, ``Eye blink detection based on multiple gabor response waves,'' in
  \emph{ICMLC}, vol.~5.\hskip 1em plus 0.5em minus 0.4em\relax IEEE, 2008, pp.
  2852--2856.

\bibitem{wang2009face}
L.~Wang, X.~Ding, and C.~Fang, ``Face live detection method based on
  physiological motion analysis,'' \emph{Tsinghua Science \& Technology},
  vol.~14, no.~6, pp. 685--690, 2009.

\bibitem{bao2009liveness}
W.~Bao, H.~Li, N.~Li, and W.~Jiang, ``A liveness detection method for face
  recognition based on optical flow field,'' in \emph{ICASSP}, 2009.

\bibitem{bigun2004assuring}
J.~Bigun, H.~Fronthaler, and K.~Kollreider, ``Assuring liveness in biometric
  identity authentication by real-time face tracking,'' in \emph{CIHSPS}.\hskip
  1em plus 0.5em minus 0.4em\relax IEEE, 2004.

\bibitem{ali2012liveness}
A.~Ali, F.~Deravi, and S.~Hoque, ``Liveness detection using gaze
  collinearity,'' in \emph{ICEST}.\hskip 1em plus 0.5em minus 0.4em\relax IEEE,
  2012.

\bibitem{Liu2018Learning}
Y.~Liu, A.~Jourabloo, and X.~Liu, ``Learning deep models for face
  anti-spoofing: Binary or auxiliary supervision,'' in \emph{CVPR}, 2018.

\bibitem{lin2019face}
B.~Lin, X.~Li, Z.~Yu, and G.~Zhao, ``Face liveness detection by rppg features
  and contextual patch-based cnn,'' in \emph{ICBEA}.\hskip 1em plus 0.5em minus
  0.4em\relax ACM, 2019.

\bibitem{yu2019remote1}
Z.~Yu, W.~Peng, X.~Li, X.~Hong, and G.~Zhao, ``Remote heart rate measurement
  from highly compressed facial videos: an end-to-end deep learning solution
  with video enhancement,'' in \emph{ICCV}, 2019.

\bibitem{boulkenafet2015face}
Z.~Boulkenafet, J.~Komulainen, and A.~Hadid, ``Face anti-spoofing based on
  color texture analysis,'' in \emph{ICIP}, 2015.

\bibitem{boulkenafet2016face2}
{Boulkenafet, Zinelabidine and Komulainen, Jukka and Hadid, Abdenour}, ``Face
  antispoofing using speeded-up robust features and fisher vector encoding,''
  \emph{SPL}, vol.~24, no.~2, pp. 141--145, 2016.

\bibitem{tan2010face}
X.~Tan, Y.~Li, J.~Liu, and L.~Jiang, ``Face liveness detection from a single
  image with sparse low rank bilinear discriminative model,'' in
  \emph{ECCV}.\hskip 1em plus 0.5em minus 0.4em\relax Springer, 2010.

\bibitem{song2019discriminative}
X.~Song, X.~Zhao, L.~Fang, and T.~Lin, ``Discriminative representation
  combinations for accurate face spoofing detection,'' \emph{Pattern
  Recognition}, 2019.

\bibitem{asim2017cnn}
M.~Asim, Z.~Ming, and M.~Y. Javed, ``Cnn based spatio-temporal feature
  extraction for face anti-spoofing,'' in \emph{ICIVC}.\hskip 1em plus 0.5em
  minus 0.4em\relax IEEE, 2017.

\bibitem{rehman2020enhancing}
Y.~A.~U. Rehman, L.-M. Po, and J.~Komulainen, ``Enhancing deep discriminative
  feature maps via perturbation for face presentation attack detection,''
  \emph{Image and Vision Computing}, vol.~94, p. 103858, 2020.

\bibitem{khammari2019robust}
M.~Khammari, ``Robust face anti-spoofing using cnn with lbp and wld,''
  \emph{IET Image Processing}, 2019.

\bibitem{yu2020searching}
Z.~Yu, C.~Zhao, Z.~Wang, Y.~Qin, Z.~Su, X.~Li, F.~Zhou, and G.~Zhao,
  ``Searching central difference convolutional networks for face
  anti-spoofing,'' in \emph{CVPR}, 2020.

\bibitem{yu2020face}
Z.~Yu, X.~Li, X.~Niu, J.~Shi, and G.~Zhao, ``Face anti-spoofing with human
  material perception,'' in \emph{ECCV}, 2020.

\bibitem{yang2019face}
X.~Yang, W.~Luo, L.~Bao, Y.~Gao, D.~Gong, S.~Zheng, Z.~Li, and W.~Liu, ``Face
  anti-spoofing: Model matters, so does data,'' in \emph{CVPR}, 2019.

\bibitem{Atoum2018Face}
Y.~Atoum, Y.~Liu, A.~Jourabloo, and X.~Liu, ``Face anti-spoofing using patch
  and depth-based cnns,'' in \emph{IJCB}, 2017.

\bibitem{yu2020multi}
Z.~Yu, Y.~Qin, X.~Li, Z.~Wang, C.~Zhao, Z.~Lei, and G.~Zhao, ``Multi-modal face
  anti-spoofing based on central difference networks,'' in \emph{CVPRW}, 2020.

\bibitem{zhang2020casia}
S.~Zhang, A.~Liu, J.~Wan, Y.~Liang, G.~Guo, S.~Escalera, H.~J. Escalante, and
  S.~Z. Li, ``Casia-surf: A large-scale multi-modal benchmark for face
  anti-spoofing,'' \emph{TBIOM}, vol.~2, no.~2, pp. 182--193, 2020.

\bibitem{yang2014learn}
J.~Yang, Z.~Lei, and S.~Z. Li, ``Learn convolutional neural network for face
  anti-spoofing,'' \emph{arXiv preprint arXiv:1408.5601}, 2014.

\bibitem{Li2017An}
L.~Li, X.~Feng, Z.~Boulkenafet, Z.~Xia, M.~Li, and A.~Hadid, ``An original face
  anti-spoofing approach using partial convolutional neural network,'' in
  \emph{IPTA}, 2016.

\bibitem{Patel2016Cross}
K.~Patel, H.~Han, and A.~K. Jain, ``Cross-database face antispoofing with
  robust feature representation,'' in \emph{CCBR}, 2016.

\bibitem{george2019deep}
A.~George and S.~Marcel, ``Deep pixel-wise binary supervision for face
  presentation attack detection,'' in \emph{ICB}, no. CONF, 2019.

\bibitem{jourabloo2018face}
A.~Jourabloo, Y.~Liu, and X.~Liu, ``Face de-spoofing: Anti-spoofing via noise
  modeling,'' in \emph{ECCV}, 2018.

\bibitem{jia20203d}
S.~Jia, X.~Li, C.~Hu, G.~Guo, and Z.~Xu, ``3d face anti-spoofing with
  factorized bilinear coding,'' \emph{arXiv preprint arXiv:2005.06514}, 2020.

\bibitem{li2020compactnet}
L.~Li, Z.~Xia, X.~Jiang, F.~Roli, and X.~Feng, ``Compactnet: learning a compact
  space for face presentation attack detection,'' \emph{Neurocomputing}, 2020.

\bibitem{kim2019basn}
T.~Kim, Y.~Kim, I.~Kim, and D.~Kim, ``Basn: Enriching feature representation
  using bipartite auxiliary supervisions for face anti-spoofing,'' in
  \emph{ICCVW}, 2019.

\bibitem{yu2020fas2}
Z.~{Yu}, J.~{Wan}, Y.~{Qin}, X.~{Li}, S.~Z. {Li}, and G.~{Zhao}, ``Nas-fas:
  Static-dynamic central difference network search for face anti-spoofing,''
  \emph{IEEE TPAMI}, pp. 1--1, 2020.

\bibitem{liu2019deep}
Y.~Liu, J.~Stehouwer, A.~Jourabloo, and X.~Liu, ``Deep tree learning for
  zero-shot face anti-spoofing,'' in \emph{CVPR}, 2019.

\bibitem{sun2020face}
W.~Sun, Y.~Song, C.~Chen, J.~Huang, and A.~C. Kot, ``Face spoofing detection
  based on local ternary label supervision in fully convolutional networks,''
  \emph{TIFS}, 2020.

\bibitem{li3dpc}
X.~Li, J.~Wan, Y.~Jin, A.~Liu, G.~Guo, and S.~Z. Li, ``3dpc-net: 3d point cloud
  network for face anti-spoofing,'' 2020.

\bibitem{feng2020learning}
H.~Feng, Z.~Hong, H.~Yue, Y.~Chen, K.~Wang, J.~Han, J.~Liu, and E.~Ding,
  ``Learning generalized spoof cues for face anti-spoofing,'' \emph{arXiv
  preprint arXiv:2005.03922}, 2020.

\bibitem{liu2020physics}
Y.~Liu and X.~Liu, ``Physics-guided spoof trace disentanglement for generic
  face anti-spoofing,'' \emph{arXiv preprint arXiv:2012.05185}, 2020.

\bibitem{qin2021meta}
Y.~Qin, Z.~Yu, L.~Yan, Z.~Wang, C.~Zhao, and Z.~Lei, ``Meta-teacher for face
  anti-spoofing,'' \emph{TPAMI}, 2021.

\bibitem{zhang2020celeba}
Y.~Zhang, Z.~Yin, Y.~Li, G.~Yin, J.~Yan, J.~Shao, and Z.~Liu, ``Celeba-spoof:
  Large-scale face anti-spoofing dataset with rich annotations,'' in
  \emph{ECCV}.\hskip 1em plus 0.5em minus 0.4em\relax Springer, 2020.

\bibitem{george2019biometric}
A.~George, Z.~Mostaani, D.~Geissenbuhler, O.~Nikisins, A.~Anjos, and S.~Marcel,
  ``Biometric face presentation attack detection with multi-channel
  convolutional neural network,'' \emph{TIFS}, 2019.

\bibitem{steiner2016reliable}
H.~Steiner, A.~Kolb, and N.~Jung, ``Reliable face anti-spoofing using
  multispectral swir imaging,'' in \emph{ICB}.\hskip 1em plus 0.5em minus
  0.4em\relax IEEE, 2016.

\bibitem{tian2020face}
Y.~Tian, K.~Zhang, L.~Wang, and Z.~Sun, ``Face anti-spoofing by learning
  polarization cues in a real-world scenario,'' \emph{arXiv preprint
  arXiv:2003.08024}, 2020.

\bibitem{shao2019multi}
R.~Shao, X.~Lan, J.~Li, and P.~C. Yuen, ``Multi-adversarial discriminative deep
  domain generalization for face presentation attack detection,'' in
  \emph{CVPR}, 2019.

\bibitem{shao2019regularized}
R.~Shao, X.~Lan, and P.~C. Yuen, ``Regularized fine-grained meta face
  anti-spoofing,'' in \emph{AAAI}, 2020.

\bibitem{wang2020cross}
G.~Wang, H.~Han, S.~Shan, and X.~Chen, ``Cross-domain face presentation attack
  detection via multi-domain disentangled representation learning,'' in
  \emph{CVPR}, 2020.

\bibitem{jia2020single}
Y.~Jia, J.~Zhang, S.~Shan, and X.~Chen, ``Single-side domain generalization for
  face anti-spoofing,'' in \emph{CVPR}, 2020.

\bibitem{arashloo2017anomaly}
S.~R. Arashloo, J.~Kittler, and W.~Christmas, ``An anomaly detection approach
  to face spoofing detection: A new formulation and evaluation protocol,''
  \emph{IEEE access}, 2017.

\bibitem{qin2019learning}
Y.~Qin, C.~Zhao, X.~Zhu, Z.~Wang, Z.~Yu, T.~Fu, F.~Zhou, J.~Shi, and Z.~Lei,
  ``Learning meta model for zero-and few-shot face anti-spoofing,'' in
  \emph{AAAI}, 2020.

\bibitem{qin2020one}
Y.~Qin, W.~Zhang, J.~Shi, Z.~Wang, and L.~Yan, ``One-class adaptation face
  anti-spoofing with loss function search,'' \emph{Neurocomputing}, vol. 417,
  pp. 384--395, 2020.

\bibitem{heusch2020deep}
G.~Heusch, A.~George, D.~Geissb{\"u}hler, Z.~Mostaani, and S.~Marcel, ``Deep
  models and shortwave infrared information to detect face presentation
  attacks,'' \emph{TBIOM}, 2020.

\bibitem{pereira2020rise}
L.~A. Pereira, A.~Pinto, F.~A. Andal{\'o}, A.~M. Ferreira, B.~Lavi,
  A.~Soriano-Vargas, M.~V. Cirne, and A.~Rocha, ``The rise of data-driven
  models in presentation attack detection,'' in \emph{Deep Biometrics}.\hskip
  1em plus 0.5em minus 0.4em\relax Springer, 2020, pp. 289--311.

\bibitem{jia2020survey}
S.~Jia, G.~Guo, and Z.~Xu, ``A survey on 3d mask presentation attack detection
  and countermeasures,'' \emph{Pattern Recognition}, 2020.

\bibitem{el2020deep}
Y.~S. El-Din, M.~N. Moustafa, and H.~Mahdi, ``Deep convolutional neural
  networks for face and iris presentation attack detection: survey and case
  study,'' \emph{IET Biometrics}, 2020.

\bibitem{liu2021contrastive}
A.~Liu, C.~Zhao, Z.~Yu, J.~Wan, A.~Su, X.~Liu, Z.~Tan, S.~Escalera, J.~Xing,
  Y.~Liang \emph{et~al.}, ``Contrastive context-aware learning for 3d
  high-fidelity mask face presentation attack detection,'' \emph{arXiv preprint
  arXiv:2104.06148}, 2021.

\bibitem{tolosana2020deepfakes}
R.~Tolosana, R.~Vera-Rodriguez, J.~Fierrez, A.~Morales, and J.~Ortega-Garcia,
  ``Deepfakes and beyond: A survey of face manipulation and fake detection,''
  \emph{Information Fusion}, 2020.

\bibitem{goswami2019detecting}
G.~Goswami, A.~Agarwal, N.~Ratha, R.~Singh, and M.~Vatsa, ``Detecting and
  mitigating adversarial perturbations for robust face recognition,''
  \emph{International Journal of Computer Vision}, 2019.

\bibitem{liu2021cross}
A.~Liu, X.~Li, J.~Wan, Y.~Liang, S.~Escalera, H.~J. Escalante, M.~Madadi,
  Y.~Jin, Z.~Wu, X.~Yu \emph{et~al.}, ``Cross-ethnicity face anti-spoofing
  recognition challenge: A review,'' \emph{IET Biometrics}, 2021.

\bibitem{hernandez2019introduction}
J.~Hernandez-Ortega, J.~Fierrez, A.~Morales, and J.~Galbally, ``Introduction to
  face presentation attack detection,'' in \emph{Handbook of Biometric
  Anti-Spoofing}.\hskip 1em plus 0.5em minus 0.4em\relax Springer, 2019, pp.
  187--206.

\bibitem{de2013can}
T.~de~Freitas~Pereira, A.~Anjos, J.~M. De~Martino, and S.~Marcel, ``Can face
  anti-spoofing countermeasures work in a real world scenario?'' in
  \emph{ICB}.\hskip 1em plus 0.5em minus 0.4em\relax IEEE, 2013, pp. 1--8.

\bibitem{li2018face}
L.~Li, P.~L. Correia, and A.~Hadid, ``Face recognition under spoofing attacks:
  countermeasures and research directions,'' \emph{IET Biometrics}, vol.~7,
  no.~1, pp. 3--14, 2018.

\bibitem{marcel2019handbook}
S.~Marcel, M.~S. Nixon, J.~Fierrez, and N.~Evans, \emph{Handbook of biometric
  anti-spoofing: Presentation attack detection}.\hskip 1em plus 0.5em minus
  0.4em\relax Springer, 2019.

\bibitem{ramachandra2017presentation}
R.~Ramachandra and C.~Busch, ``Presentation attack detection methods for face
  recognition systems: A comprehensive survey,'' \emph{ACM Computing Surveys
  (CSUR)}, vol.~50, no.~1, pp. 1--37, 2017.

\bibitem{peixoto2011face}
B.~Peixoto, C.~Michelassi, and A.~Rocha, ``Face liveness detection under bad
  illumination conditions,'' in \emph{ICIP}.\hskip 1em plus 0.5em minus
  0.4em\relax IEEE, 2011.

\bibitem{Zhang2012A}
Z.~Zhang, J.~Yan, S.~Liu, Z.~Lei, D.~Yi, and S.~Z. Li, ``A face antispoofing
  database with diverse attacks,'' in \emph{ICB}, 2012.

\bibitem{ReplayAttack}
I.~Chingovska, A.~Anjos, and S.~Marcel, ``On the effectiveness of local binary
  patterns in face anti-spoofing,'' in \emph{Biometrics Special Interest
  Group}, 2012.

\bibitem{kose2013shape}
N.~Kose and J.-L. Dugelay, ``Shape and texture based countermeasure to protect
  face recognition systems against mask attacks,'' in \emph{CVPRW}, 2013.

\bibitem{wen2015face}
D.~Wen, H.~Han, and A.~K. Jain, ``Face spoof detection with image distortion
  analysis,'' \emph{TIFS}, 2015.

\bibitem{pinto2015using}
A.~Pinto, W.~R. Schwartz, H.~Pedrini, and A.~de~Rezende~Rocha, ``Using visual
  rhythms for detecting video-based facial spoof attacks,'' \emph{TIFS}, 2015.

\bibitem{costa2016replay}
A.~Costa-Pazo, S.~Bhattacharjee, E.~Vazquez-Fernandez, and S.~Marcel, ``The
  replay-mobile face presentation-attack database,'' in \emph{BIOSIG}.\hskip
  1em plus 0.5em minus 0.4em\relax IEEE, 2016.

\bibitem{liu20163d}
S.~Liu, P.~C. Yuen, S.~Zhang, and G.~Zhao, ``3d mask face anti-spoofing with
  remote photoplethysmography,'' in \emph{ECCV}.\hskip 1em plus 0.5em minus
  0.4em\relax Springer, 2016.

\bibitem{manjani2017detecting}
I.~Manjani, S.~Tariyal, M.~Vatsa, R.~Singh, and A.~Majumdar, ``Detecting
  silicone mask-based presentation attack via deep dictionary learning,''
  \emph{TIFS}, 2017.

\bibitem{Boulkenafet2017OULU}
Z.~Boulkenafet, J.~Komulainen, L.~Li, X.~Feng, and A.~Hadid, ``Oulu-npu: A
  mobile face presentation attack database with real-world variations,'' in
  \emph{FGR}, 2017.

\bibitem{li2018unsupervised}
H.~Li, W.~Li, H.~Cao, S.~Wang, F.~Huang, and A.~C. Kot, ``Unsupervised domain
  adaptation for face anti-spoofing,'' \emph{TIFS}, 2018.

\bibitem{vareto2020swax}
R.~H. Vareto, A.~M. Saldanha, and W.~R. Schwartz, ``The swax benchmark:
  Attacking biometric systems with wax figures,'' in \emph{ICASSP}, 2020.

\bibitem{almeida2020detecting}
W.~R. Almeida, F.~A. Andal{\'o}, R.~Padilha, G.~Bertocco, W.~Dias, R.~d.~S.
  Torres, J.~Wainer, and A.~Rocha, ``Detecting face presentation attacks in
  mobile devices with a patch-based cnn and a sensor-aware loss function,''
  \emph{PloS one}, 2020.

\bibitem{erdogmus2014spoofing}
N.~Erdogmus and S.~Marcel, ``Spoofing face recognition with 3d masks,''
  \emph{TIFS}, 2014.

\bibitem{raghavendra2015presentation}
R.~Raghavendra, K.~B. Raja, and C.~Busch, ``Presentation attack detection for
  face recognition using light field camera,'' \emph{TIP}, vol.~24, no.~3, pp.
  1060--1075, 2015.

\bibitem{galbally2016three}
J.~Galbally and R.~Satta, ``Three-dimensional and two-and-a-half-dimensional
  face recognition spoofing using three-dimensional printed models,'' \emph{IET
  Biometrics}, 2016.

\bibitem{chingovska2016face}
I.~Chingovska, N.~Erdogmus, A.~Anjos, and S.~Marcel, ``Face recognition systems
  under spoofing attacks,'' in \emph{Face Recognition Across the Imaging
  Spectrum}.\hskip 1em plus 0.5em minus 0.4em\relax Springer, 2016, pp.
  165--194.

\bibitem{agarwal2017face}
A.~Agarwal, D.~Yadav, N.~Kohli, R.~Singh, M.~Vatsa, and A.~Noore, ``Face
  presentation attack with latex masks in multispectral videos,'' in
  \emph{CVPRW}, 2017.

\bibitem{bhattacharjee2017you}
S.~Bhattacharjee and S.~Marcel, ``What you can't see can help
  you-extended-range imaging for 3d-mask presentation attack detection,'' in
  \emph{BIOSIG}.\hskip 1em plus 0.5em minus 0.4em\relax IEEE, 2017.

\bibitem{liu2019light}
M.~Liu, H.~Fu, Y.~Wei, Y.~A.~U. Rehman, L.-m. Po, and W.~L. Lo, ``Light
  field-based face liveness detection with convolutional neural networks,''
  \emph{Journal of Electronic Imaging}, 2019.

\bibitem{bhattacharjee2018spoofing}
S.~Bhattacharjee, A.~Mohammadi, and S.~Marcel, ``Spoofing deep face recognition
  with custom silicone masks,'' in \emph{BTAS}, 2018.

\bibitem{xiao20193dma}
J.~Xiao, Y.~Tang, J.~Guo, Y.~Yang, X.~Zhu, Z.~Lei, and S.~Z. Li, ``3dma: A
  multi-modality 3d mask face anti-spoofing database,'' in \emph{AVSS}.\hskip
  1em plus 0.5em minus 0.4em\relax IEEE, 2019.

\bibitem{casiasurf}
S.~Zhang, X.~Wang, A.~Liu, C.~Zhao, J.~Wan, S.~Escalera, H.~Shi, Z.~Wang, and
  S.~Z. Li, ``A dataset and benchmark for large-scale multi-modal face
  anti-spoofing,'' in \emph{CVPR}, 2019.

\bibitem{li2020casia}
A.~Li, Z.~Tan, X.~Li, J.~Wan, S.~Escalera, G.~Guo, and S.~Z. Li, ``Casia-surf
  cefa: A benchmark for multi-modal cross-ethnicity face anti-spoofing,''
  \emph{WACV}, 2021.

\bibitem{rostami2021detection}
M.~Rostami, L.~Spinoulas, M.~Hussein, J.~Mathai, and W.~Abd-Almageed,
  ``Detection and continual learning of novel face presentation attacks,'' in
  \emph{ICCV}, 2021.

\bibitem{galbally2012high}
J.~Galbally, F.~Alonso-Fernandez, J.~Fierrez, and J.~Ortega-Garcia, ``A high
  performance fingerprint liveness detection method based on quality related
  features,'' \emph{Future Generation Computer Systems}, vol.~28, no.~1, pp.
  311--321, 2012.

\bibitem{chingovska2014biometrics}
I.~Chingovska, A.~R. Dos~Anjos, and S.~Marcel, ``Biometrics evaluation under
  spoofing attacks,'' \emph{IEEE TIFS}, 2014.

\bibitem{iso2017information}
I.~J.~S. Biometrics., ``Information technology--biometric presentation attack
  detection--part 3: testing and reporting,'' 2017.

\bibitem{wang2020deep}
Z.~Wang, Z.~Yu, C.~Zhao, X.~Zhu, Y.~Qin, Q.~Zhou, F.~Zhou, and Z.~Lei, ``Deep
  spatial gradient and temporal depth learning for face anti-spoofing,'' in
  \emph{CVPR}, 2020.

\bibitem{zhang2020face}
K.-Y. Zhang, T.~Yao, J.~Zhang, Y.~Tai, S.~Ding, J.~Li, F.~Huang, H.~Song, and
  L.~Ma, ``Face anti-spoofing via disentangled representation learning,'' in
  \emph{ECCV}.\hskip 1em plus 0.5em minus 0.4em\relax Springer, 2020.

\bibitem{yu2021dual}
Z.~Yu, Y.~Qin, H.~Zhao, X.~Li, and G.~Zhao, ``Dual-cross central difference
  network for face anti-spoofing,'' in \emph{IJCAI}, 2021.

\bibitem{liu2020disentangling}
Y.~Liu, J.~Stehouwer, and X.~Liu, ``On disentangling spoof trace for generic
  face anti-spoofing,'' in \emph{ECCV}.\hskip 1em plus 0.5em minus 0.4em\relax
  Springer, 2020.

\bibitem{xu2020improving}
X.~Xu, Y.~Xiong, and W.~Xia, ``On improving temporal consistency for online
  face liveness detection,'' in \emph{ICCVW}, 2021.

\bibitem{wang2022patchnet}
C.-Y. Wang, Y.-D. Lu, S.-T. Yang, and S.-H. Lai, ``Patchnet: A simple face
  anti-spoofing framework via fine-grained patch recognition,'' in \emph{CVPR},
  2022.

\bibitem{wang2022domain}
Z.~Wang, Z.~Wang, Z.~Yu, W.~Deng, J.~Li, S.~Li, and Z.~Wang, ``Domain
  generalization via shuffled style assembly for face anti-spoofing,'' in
  \emph{CVPR}, 2022.

\bibitem{jia2021unified}
Y.~Jia, J.~Zhang, S.~Shan, and X.~Chen, ``Unified unsupervised and
  semi-supervised domain adaptation network for cross-scenario face
  anti-spoofing,'' \emph{Pattern Recognition}, 2021.

\bibitem{He2015Deep}
K.~He, X.~Zhang, S.~Ren, and J.~Sun, ``Deep residual learning for image
  recognition,'' 2016.

\bibitem{huang2017densely}
G.~Huang, Z.~Liu, L.~Van Der~Maaten, and K.~Q. Weinberger, ``Densely connected
  convolutional networks,'' in \emph{CVPR}, 2017.

\bibitem{long2015fully}
J.~Long, E.~Shelhamer, and T.~Darrell, ``Fully convolutional networks for
  semantic segmentation,'' in \emph{CVPR}, 2015.

\bibitem{ren2016faster}
S.~Ren, K.~He, R.~Girshick, and J.~Sun, ``Faster r-cnn: towards real-time
  object detection with region proposal networks,'' \emph{IEEE TPAMI}, vol.~39,
  no.~6, pp. 1137--1149, 2016.

\bibitem{ahonen2006face}
T.~Ahonen, A.~Hadid, and M.~Pietikainen, ``Face description with local binary
  patterns: Application to face recognition,'' \emph{TPAMI}, no.~12, pp.
  2037--2041, 2006.

\bibitem{dalal2005histograms}
N.~Dalal and B.~Triggs, ``Histograms of oriented gradients for human
  detection,'' in \emph{CVPR}.\hskip 1em plus 0.5em minus 0.4em\relax IEEE,
  2005.

\bibitem{galbally2014face}
J.~Galbally and S.~Marcel, ``Face anti-spoofing based on general image quality
  assessment,'' in \emph{ICPR}, 2014.

\bibitem{brox2010large}
T.~Brox and J.~Malik, ``Large displacement optical flow: descriptor matching in
  variational motion estimation,'' \emph{TPAMI}, vol.~33, no.~3, pp. 500--513,
  2010.

\bibitem{niu2020video}
X.~Niu, Z.~Yu, H.~Han, X.~Li, S.~Shan, and G.~Zhao, ``Video-based remote
  physiological measurement via cross-verified feature disentangling,'' in
  \emph{ECCV}.\hskip 1em plus 0.5em minus 0.4em\relax Springer, 2020.

\bibitem{cai2019learning}
R.~Cai and C.~Chen, ``Learning deep forest with multi-scale local binary
  pattern features for face anti-spoofing,'' \emph{arXiv preprint
  arXiv:1910.03850}, 2019.

\bibitem{feng2016integration}
L.~Feng, L.-M. Po, Y.~Li, X.~Xu, F.~Yuan, T.~C.-H. Cheung, and K.-W. Cheung,
  ``Integration of image quality and motion cues for face anti-spoofing: A
  neural network approach,'' \emph{Journal of Visual Communication and Image
  Representation}, 2016.

\bibitem{yu2021transrppg}
Z.~Yu, X.~Li, P.~Wang, and G.~Zhao, ``Transrppg: Remote photoplethysmography
  transformer for 3d mask face presentation attack detection,'' \emph{IEEE
  SPL}, 2021.

\bibitem{li20203d}
L.~Li, Z.~Xia, X.~Jiang, Y.~Ma, F.~Roli, and X.~Feng, ``3d face mask
  presentation attack detection based on intrinsic image analysis,'' \emph{IET
  Biometrics}, 2020.

\bibitem{agarwal2019chif}
A.~Agarwal, M.~Vatsa, and R.~Singh, ``Chif: Convoluted histogram image features
  for detecting silicone mask based face presentation attack,'' in \emph{BTAS},
  2019.

\bibitem{shao2018joint}
R.~Shao, X.~Lan, and P.~C. Yuen, ``Joint discriminative learning of deep
  dynamic textures for 3d mask face anti-spoofing,'' \emph{TIFS}, 2018.

\bibitem{zhao2007dynamic}
G.~Zhao and M.~Pietikainen, ``Dynamic texture recognition using local binary
  patterns with an application to facial expressions,'' \emph{IEEE TPAMI},
  vol.~29, no.~6, pp. 915--928, 2007.

\bibitem{sharifi2019score}
O.~Sharifi, ``Score-level-based face anti-spoofing system using handcrafted and
  deep learned characteristics,'' \emph{International Journal of Image,
  Graphics and Signal Processing}, 2019.

\bibitem{rehman2019perturbing}
Y.~A.~U. Rehman, L.-M. Po, M.~Liu, Z.~Zou, and W.~Ou, ``Perturbing
  convolutional feature maps with histogram of oriented gradients for face
  liveness detection,'' in \emph{CISIS and ICEUTE 2019}.\hskip 1em plus 0.5em
  minus 0.4em\relax Springer, 2019.

\bibitem{li2019replayed}
L.~Li, Z.~Xia, A.~Hadid, X.~Jiang, H.~Zhang, and X.~Feng, ``Replayed video
  attack detection based on motion blur analysis,'' \emph{TIFS}, 2019.

\bibitem{ronneberger2015u}
O.~Ronneberger, P.~Fischer, and T.~Brox, ``U-net: Convolutional networks for
  biomedical image segmentation,'' in \emph{International Conference on Medical
  image computing and computer-assisted intervention}.\hskip 1em plus 0.5em
  minus 0.4em\relax Springer, 2015.

\bibitem{ioffe2015batch}
S.~Ioffe and C.~Szegedy, ``Batch normalization: Accelerating deep network
  training by reducing internal covariate shift,'' in \emph{International
  conference on machine learning}.\hskip 1em plus 0.5em minus 0.4em\relax PMLR,
  2015.

\bibitem{srivastava2014dropout}
N.~Srivastava, G.~Hinton, A.~Krizhevsky, I.~Sutskever, and R.~Salakhutdinov,
  ``Dropout: a simple way to prevent neural networks from overfitting,''
  \emph{The journal of machine learning research}, 2014.

\bibitem{hermans2017defense}
A.~Hermans, L.~Beyer, and B.~Leibe, ``In defense of the triplet loss for person
  re-identification,'' \emph{arXiv preprint arXiv:1703.07737}, 2017.

\bibitem{lucena2017transfer}
O.~Lucena, A.~Junior, V.~Moia, R.~Souza, E.~Valle, and R.~Lotufo, ``Transfer
  learning using convolutional neural networks for face anti-spoofing,'' in
  \emph{ICIAR}, 2017.

\bibitem{chen2019attention}
H.~Chen, G.~Hu, Z.~Lei, Y.~Chen, N.~M. Robertson, and S.~Z. Li,
  ``Attention-based two-stream convolutional networks for face spoofing
  detection,'' \emph{TIFS}, 2019.

\bibitem{george2020effectiveness}
A.~George and S.~Marcel, ``On the effectiveness of vision transformers for
  zero-shot face anti-spoofing,'' \emph{arXiv preprint arXiv:2011.08019}, 2020.

\bibitem{sandler2018mobilenetv2}
M.~Sandler, A.~Howard, M.~Zhu, A.~Zhmoginov, and L.-C. Chen, ``Mobilenetv2:
  Inverted residuals and linear bottlenecks,'' in \emph{CVPR}, 2018, pp.
  4510--4520.

\bibitem{deb2020look}
D.~Deb and A.~K. Jain, ``Look locally infer globally: A generalizable face
  anti-spoofing approach,'' \emph{TIFS}, 2020.

\bibitem{Xu2016Learning}
Z.~Xu, S.~Li, and W.~Deng, ``Learning temporal features using lstm-cnn
  architecture for face anti-spoofing,'' in \emph{ACPR}, 2015.

\bibitem{muhammad2019face}
U.~Muhammad, T.~Holmberg, W.~C. de~Melo, and A.~Hadid, ``Face anti-spoofing via
  sample learning based recurrent neural network (rnn).'' in \emph{BMVC}, 2019.

\bibitem{ge2020face}
H.~Ge, X.~Tu, W.~Ai, Y.~Luo, Z.~Ma, and M.~Xie, ``Face anti-spoofing by the
  enhancement of temporal motion,'' in \emph{CTISC}.\hskip 1em plus 0.5em minus
  0.4em\relax IEEE, 2020.

\bibitem{hochreiter1997long}
S.~Hochreiter and J.~Schmidhuber, ``Long short-term memory,'' \emph{Neural
  computation}, vol.~9, no.~8, pp. 1735--1780, 1997.

\bibitem{hao2019face}
H.~Hao, M.~Pei, and M.~Zhao, ``Face liveness detection based on client identity
  using siamese network,'' in \emph{PRCV}.\hskip 1em plus 0.5em minus
  0.4em\relax Springer, 2019.

\bibitem{chen2021camera}
B.~Chen, W.~Yang, H.~Li, S.~Wang, and S.~Kwong, ``Camera invariant feature
  learning for generalized face anti-spoofing,'' \emph{TIFS}, 2021.

\bibitem{mohammadi2020improving}
A.~Mohammadi, S.~Bhattacharjee, and S.~Marcel, ``Improving cross-dataset
  performance of face presentation attack detection systems using face
  recognition datasets,'' in \emph{ICASSP}, 2020.

\bibitem{peng2020ts}
D.~Peng, J.~Xiao, R.~Zhu, and G.~Gao, ``Ts-fen: Probing feature selection
  strategy for face anti-spoofing,'' in \emph{ICASS}.\hskip 1em plus 0.5em
  minus 0.4em\relax IEEE, 2020.

\bibitem{wu2021dual}
H.~Wu, D.~Zeng, Y.~Hu, H.~Shi, and T.~Mei, ``Dual spoof disentanglement
  generation for face anti-spoofing with depth uncertainty learning,''
  \emph{TCSVT}, 2021.

\bibitem{wang2022learning}
Z.~Wang, Q.~Wang, W.~Deng, and G.~Guo, ``Learning multi-granularity temporal
  characteristics for face anti-spoofing,'' \emph{TIFS}, 2022.

\bibitem{hossaindeeppixbis}
M.~S. Hossain, L.~Rupty, K.~Roy, M.~Hasan, S.~Sengupta, and N.~Mohammed,
  ``A-deeppixbis: Attentional angular margin for face anti-spoofing,'' 2020.

\bibitem{yu2020auto2}
Z.~Yu, Y.~Qin, X.~Xu, C.~Zhao, Z.~Wang, Z.~Lei, and G.~Zhao, ``Auto-fas:
  Searching lightweight networks for face anti-spoofing,'' in \emph{ICASSP},
  2020.

\bibitem{roy2021bi}
K.~Roy, M.~Hasan, L.~Rupty, M.~Hossain, S.~Sengupta, S.~N. Taus, N.~Mohammed
  \emph{et~al.}, ``Bi-fpnfas: Bi-directional feature pyramid network for
  pixel-wise face anti-spoofing by leveraging fourier spectra,''
  \emph{Sensors}, vol.~21, no.~8, p. 2799, 2021.

\bibitem{gretton2012kernel}
A.~Gretton, K.~M. Borgwardt, M.~J. Rasch, B.~Sch{\"o}lkopf, and A.~Smola, ``A
  kernel two-sample test,'' \emph{The Journal of Machine Learning Research},
  2012.

\bibitem{wang2019improving}
G.~Wang, H.~Han, S.~Shan, and X.~Chen, ``Improving cross-database face
  presentation attack detection via adversarial domain adaptation,'' in
  \emph{ICB}.\hskip 1em plus 0.5em minus 0.4em\relax IEEE, 2019.

\bibitem{wang2020unsupervised}
{Wang, Guoqing and Han, Hu and Shan, Shiguang and Chen, Xilin}, ``Unsupervised
  adversarial domain adaptation for cross-domain face presentation attack
  detection,'' \emph{TIFS}, 2020.

\bibitem{quan2021progressive}
R.~Quan, Y.~Wu, X.~Yu, and Y.~Yang, ``Progressive transfer learning for face
  anti-spoofing,'' \emph{TIP}, vol.~30, pp. 3946--3955, 2021.

\bibitem{zhou2019face}
F.~Zhou, C.~Gao, F.~Chen, C.~Li, X.~Li, F.~Yang, and Y.~Zhao, ``Face
  anti-spoofing based on multi-layer domain adaptation,'' in
  \emph{ICMEW}.\hskip 1em plus 0.5em minus 0.4em\relax IEEE, 2019.

\bibitem{mohammadi2020domain}
A.~Mohammadi, S.~Bhattacharjee, and S.~Marcel, ``Domain adaptation for
  generalization of face presentation attack detection in mobile settengs with
  minimal information,'' in \emph{ICASSP}, 2020.

\bibitem{li2020face2}
H.~Li, S.~Wang, P.~He, and A.~Rocha, ``Face anti-spoofing with deep neural
  network distillation,'' \emph{IEEE Journal of Selected Topics in Signal
  Processing}, 2020.

\bibitem{chen2021generalized}
Z.~Chen, T.~Yao, K.~Sheng, S.~Ding, Y.~Tai, J.~Li, F.~Huang, and X.~Jin,
  ``Generalizable representation learning for mixture domain face
  anti-spoofing,'' in \emph{AAAI}, 2021.

\bibitem{liu2021adaptive}
S.~Liu, K.-Y. Zhang, T.~Yao, M.~Bi, S.~Ding, J.~Li, F.~Huang, and L.~Ma,
  ``Adaptive normalized representation learning for generalizable face
  anti-spoofing,'' in \emph{ACM MM}, 2021.

\bibitem{wang2021self}
J.~Wang, J.~Zhang, Y.~Bian, Y.~Cai, C.~Wang, and S.~Pu, ``Self-domain
  adaptation for face anti-spoofing,'' in \emph{AAAI}, 2021.

\bibitem{perez2020learning}
D.~P{\'e}rez-Cabo, D.~Jim{\'e}nez-Cabello, A.~Costa-Pazo, and R.~J.
  L{\'o}pez-Sastre, ``Learning to learn face-pad: a lifelong learning
  approach,'' in \emph{IJCB}.\hskip 1em plus 0.5em minus 0.4em\relax IEEE,
  2020.

\bibitem{nikisins2018effectiveness}
O.~Nikisins, A.~Mohammadi, A.~Anjos, and S.~Marcel, ``On effectiveness of
  anomaly detection approaches against unseen presentation attacks in face
  anti-spoofing,'' in \emph{ICB}.\hskip 1em plus 0.5em minus 0.4em\relax IEEE,
  2018.

\bibitem{xiong2018unknown}
F.~Xiong and W.~AbdAlmageed, ``Unknown presentation attack detection with face
  rgb images,'' in \emph{BTAS}, 2018.

\bibitem{baweja2020anomaly}
Y.~Baweja, P.~Oza, P.~Perera, and V.~M. Patel, ``Anomaly detection-based
  unknown face presentation attack detection,'' \emph{arXiv preprint
  arXiv:2007.05856}, 2020.

\bibitem{perez2019deep}
D.~P{\'e}rez-Cabo, D.~Jim{\'e}nez-Cabello, A.~Costa-Pazo, and R.~J.
  L{\'o}pez-Sastre, ``Deep anomaly detection for generalized face
  anti-spoofing,'' in \emph{CVPRW}, 2019.

\bibitem{george2020learning}
A.~George and S.~Marcel, ``Learning one class representations for face
  presentation attack detection using multi-channel convolutional neural
  networks,'' \emph{TIFS}, 2020.

\bibitem{li2020unseen}
Z.~Li, H.~Li, K.-Y. Lam, and A.~C. Kot, ``Unseen face presentation attack
  detection with hypersphere loss,'' in \emph{ICASSP}, 2020.

\bibitem{rehman2020slnet}
Y.~A.~U. Rehman, L.-M. Po, and M.~Liu, ``Slnet: Stereo face liveness detection
  via dynamic disparity-maps and convolutional neural network,'' \emph{Expert
  Systems with Applications}, 2020.

\bibitem{liu2019AuroraGuard}
W.~Hu, G.~Te, J.~He, D.~Chen, and Z.~Guo, ``Aurora guard: Real-time face
  anti-spoofing via light reflection,'' \emph{arXiv preprint arXiv:
  1902.10311}, 2019.

\bibitem{wu2019review}
B.~Wu, M.~Pan, and Y.~Zhang, ``A review of face anti-spoofing and its
  applications in china,'' in \emph{International Conference on Harmony Search
  Algorithm}.\hskip 1em plus 0.5em minus 0.4em\relax Springer, 2019, pp.
  35--43.

\bibitem{connell2013fake}
J.~Connell, N.~Ratha, J.~Gentile, and R.~Bolle, ``Fake iris detection using
  structured light,'' in \emph{ICASSP}, 2013.

\bibitem{sun2016context}
X.~Sun, L.~Huang, and C.~Liu, ``Context based face spoofing detection using
  active near-infrared images,'' in \emph{ICPR}, 2016.

\bibitem{seo2019face}
J.~Seo and I.-J. Chung, ``Face liveness detection using thermal face-cnn with
  external knowledge,'' \emph{Symmetry}, 2019.

\bibitem{kang2021facial}
M.~Kang, J.~Choe, H.~Ha, H.-G. Jeon, S.~Im, and I.~S. Kweon, ``Facial depth and
  normal estimation using single dual-pixel camera,'' \emph{arXiv preprint
  arXiv:2111.12928}, 2021.

\bibitem{wu2020single}
X.~Wu, J.~Zhou, J.~Liu, F.~Ni, and H.~Fan, ``Single-shot face anti-spoofing for
  dual pixel camera,'' \emph{TIFS}, 2020.

\bibitem{farrukh2020facerevelio}
H.~Farrukh, R.~M. Aburas, S.~Cao, and H.~Wang, ``Facerevelio: a face liveness
  detection system for smartphones with a single front camera,'' in
  \emph{Proceedings of the 26th Annual International Conference on Mobile
  Computing and Networking}, 2020.

\bibitem{ebihara2019specular}
A.~F. Ebihara, K.~Sakurai, and H.~Imaoka, ``Specular-and
  diffuse-reflection-based face spoofing detection for mobile devices,'' in
  \emph{IJCB}.\hskip 1em plus 0.5em minus 0.4em\relax IEEE, 2020.

\bibitem{parkin2019recognizing}
A.~Parkin and O.~Grinchuk, ``Recognizing multi-modal face spoofing with face
  recognition networks,'' in \emph{CVPRW}, 2019.

\bibitem{kuang2019multi}
H.~Kuang, R.~Ji, H.~Liu, S.~Zhang, X.~Sun, F.~Huang, and B.~Zhang,
  ``Multi-modal multi-layer fusion network with average binary center loss for
  face anti-spoofing,'' in \emph{ACM MM}, 2019.

\bibitem{shen2019facebagnet}
T.~Shen, Y.~Huang, and Z.~Tong, ``Facebagnet: Bag-of-local-features model for
  multi-modal face anti-spoofing,'' in \emph{CVPRW}, 2019.

\bibitem{george2021cross}
A.~George and S.~Marcel, ``Cross modal focal loss for rgbd face
  anti-spoofing,'' in \emph{CVPR}, 2021.

\bibitem{nikisins2019domain}
O.~Nikisins, A.~George, and S.~Marcel, ``Domain adaptation in multi-channel
  autoencoder based features for robust face anti-spoofing,'' in
  \emph{ICB}.\hskip 1em plus 0.5em minus 0.4em\relax IEEE, 2019.

\bibitem{liu2021data}
W.~Liu, X.~Wei, T.~Lei, X.~Wang, H.~Meng, and A.~K. Nandi, ``Data fusion based
  two-stage cascade framework for multi-modality face anti-spoofing,''
  \emph{TCDS}, 2021.

\bibitem{zhang2019feathernets}
P.~Zhang, F.~Zou, Z.~Wu, N.~Dai, S.~Mark, M.~Fu, J.~Zhao, and K.~Li,
  ``Feathernets: Convolutional neural networks as light as feather for face
  anti-spoofing,'' in \emph{CVPRW}, 2019.

\bibitem{jiang2020face}
F.~Jiang, P.~Liu, X.~Shao, and X.~Zhou, ``Face anti-spoofing with generated
  near-infrared images,'' \emph{Multimedia Tools and Applications}, vol.~79,
  no.~29, pp. 21\,299--21\,323, 2020.

\bibitem{liu2021face}
A.~Liu, Z.~Tan, J.~Wan, Y.~Liang, Z.~Lei, G.~Guo, and S.~Z. Li, ``Face
  anti-spoofing via adversarial cross-modality translation,'' \emph{TIFS},
  2021.

\bibitem{mallatindirect2021}
K.~Mallat and J.-L. Dugelay, ``Indirect synthetic attack on thermal face
  biometric systems via visible-to-thermal spectrum conversion,'' in
  \emph{CVPRW}, 2021.

\bibitem{liu2022feature}
S.~Liu, S.~Lu, H.~Xu, J.~Yang, S.~Ding, and L.~Ma, ``Feature generation and
  hypothesis verification for reliable face anti-spoofing,'' in \emph{AAAI},
  2022.

\bibitem{sequeiraexploratory}
A.~F. Sequeira, T.~Gon{\c{c}}alves, W.~Silva, J.~R. Pinto, and J.~S. Cardoso,
  ``An exploratory study of interpretability for face presentation attack
  detection,'' \emph{IET Biometrics}, 2021.

\bibitem{selvaraju2017grad}
R.~R. Selvaraju, M.~Cogswell, A.~Das, R.~Vedantam, D.~Parikh, and D.~Batra,
  ``Grad-cam: Visual explanations from deep networks via gradient-based
  localization,'' in \emph{ICCV}, 2017.

\bibitem{mirzaalian2021explaining}
H.~Mirzaalian, M.~E. Hussein, L.~Spinoulas, J.~May, and W.~Abd-Almageed,
  ``Explaining face presentation attack detection using natural language,'' in
  \emph{FG}.\hskip 1em plus 0.5em minus 0.4em\relax IEEE, 2021.

\bibitem{liu2020cross}
A.~Liu, X.~Li, J.~Wan, Y.~Liang, S.~Escalera, H.~J. Escalante, M.~Madadi,
  Y.~Jin, Z.~Wu, X.~Yu \emph{et~al.}, ``Cross-ethnicity face anti-spoofing
  recognition challenge: A review,'' \emph{IET Biometrics}, vol.~10, no.~1, pp.
  24--43, 2021.

\bibitem{yu2022flexible}
Z.~Yu, C.~Zhao, K.~H. Cheng, X.~Cheng, and G.~Zhao, ``Flexible-modal face
  anti-spoofing: A benchmark,'' \emph{arXiv preprint arXiv:2202.08192}, 2022.

\bibitem{stehouwer2020noise}
J.~Stehouwer, A.~Jourabloo, Y.~Liu, and X.~Liu, ``Noise modeling, synthesis and
  classification for generic object anti-spoofing,'' in \emph{CVPR}, 2020.

\bibitem{sharif2019general}
M.~Sharif, S.~Bhagavatula, L.~Bauer, and M.~K. Reiter, ``A general framework
  for adversarial examples with objectives,'' \emph{ACM Transactions on Privacy
  and Security (TOPS)}, 2019.

\bibitem{komkov2021advhat}
S.~Komkov and A.~Petiushko, ``Advhat: Real-world adversarial attack on arcface
  face id system,'' in \emph{ICPR}, 2021.

\bibitem{yin2021adv}
B.~Yin, W.~Wang, T.~Yao, J.~Guo, Z.~Kong, S.~Ding, J.~Li, and C.~Liu,
  ``Adv-makeup: A new imperceptible and transferable attack on face
  recognition,'' in \emph{IJCAI}, 2021.

\bibitem{guo2021meaningful}
Y.~Guo, X.~Wei, G.~Wang, and B.~Zhang, ``Meaningful adversarial stickers for
  face recognition in physical world,'' \emph{arXiv preprint arXiv:2104.06728},
  2021.

\bibitem{zhang2019attacking}
B.~Zhang, B.~Tondi, and M.~Barni, ``Attacking cnn-based anti-spoofing face
  authentication in the physical domain,'' \emph{arXiv preprint
  arXiv:1910.00327}, 2019.

\bibitem{ciftci2020fakecatcher}
U.~A. Ciftci, I.~Demir, and L.~Yin, ``Fakecatcher: Detection of synthetic
  portrait videos using biological signals,'' \emph{TPAMI}, 2020.

\bibitem{sarkar2022gan}
E.~Sarkar, P.~Korshunov, L.~Colbois, and S.~Marcel, ``Are gan-based morphs
  threatening face recognition?'' in \emph{ICASSP}, 2022.

\bibitem{deb2021unified}
D.~Deb, X.~Liu, and A.~K. Jain, ``Unified detection of digital and physical
  face attacks,'' \emph{arXiv preprint arXiv:2104.02156}, 2021.

\bibitem{voigt2017eu}
P.~Voigt and A.~Von~dem Bussche, ``The eu general data protection regulation
  (gdpr),'' \emph{A Practical Guide, 1st Ed., Cham: Springer International
  Publishing}, 2017.

\bibitem{mcmahan2017communication}
B.~McMahan, E.~Moore, D.~Ramage, S.~Hampson, and B.~A. y~Arcas,
  ``Communication-efficient learning of deep networks from decentralized
  data,'' in \emph{Artificial Intelligence and Statistics}.\hskip 1em plus
  0.5em minus 0.4em\relax PMLR, 2017.

\bibitem{shao2021federated}
R.~Shao, B.~Zhang, P.~C. Yuen, and V.~M. Patel, ``Federated test-time adaptive
  face presentation attack detection with dual-phase privacy preservation,'' in
  \emph{FG}.\hskip 1em plus 0.5em minus 0.4em\relax IEEE, 2021.

\bibitem{kundu2020universal}
J.~N. Kundu, N.~Venkat, R.~V. Babu \emph{et~al.}, ``Universal source-free
  domain adaptation,'' in \emph{CVPR}, 2020.

\bibitem{lv2021combining}
L.~Lv, Y.~Xiang, X.~Li, H.~Huang, R.~Ruan, X.~Xu, and Y.~Fu, ``Combining
  dynamic image and prediction ensemble for cross-domain face anti-spoofing,''
  in \emph{ICASSP}, 2021.

\bibitem{kahm20122d}
O.~K{\"a}hm and N.~Damer, ``2d face liveness detection: An overview,'' in
  \emph{BIOSIG}.\hskip 1em plus 0.5em minus 0.4em\relax IEEE, 2012.

\bibitem{hadid2014face}
A.~Hadid, ``Face biometrics under spoofing attacks: Vulnerabilities,
  countermeasures, open issues, and research directions,'' in \emph{CVPRW},
  2014.

\bibitem{galbally2014biometric}
J.~Galbally, S.~Marcel, and J.~Fierrez, ``Biometric antispoofing methods: A
  survey in face recognition,'' \emph{IEEE Access}, 2014.

\bibitem{kumar2017comparative}
S.~Kumar, S.~Singh, and J.~Kumar, ``A comparative study on face spoofing
  attacks,'' in \emph{ICCCA}.\hskip 1em plus 0.5em minus 0.4em\relax IEEE,
  2017.

\bibitem{kisku2017face}
D.~R. Kisku and R.~D. Rakshit, ``Face spoofing and counter-spoofing: A survey
  of state-of-the-art algorithms,'' \emph{Transactions on Machine Learning and
  Artificial Intelligence}, vol.~5, no.~2, pp. 31--31, 2017.

\bibitem{souza2018far}
L.~Souza, L.~Oliveira, M.~Pamplona, and J.~Papa, ``How far did we get in face
  spoofing detection?'' \emph{Engineering Applications of Artificial
  Intelligence}, vol.~72, pp. 368--381, 2018.

\bibitem{raheem2019insight}
E.~A. Raheem, S.~M.~S. Ahmad, and W.~A.~W. Adnan, ``Insight on face liveness
  detection: A systematic literature review,'' \emph{International Journal of
  Electrical and Computer Engineering}, 2019.

\bibitem{ming2020survey}
Z.~Ming, M.~Visani, M.~M. Luqman, and J.-C. Burie, ``A survey on anti-spoofing
  methods for facial recognition with rgb cameras of generic consumer
  devices,'' \emph{Journal of Imaging}, 2020.

\bibitem{cai2020drl}
R.~Cai, H.~Li, S.~Wang, C.~Chen, and A.~C. Kot, ``Drl-fas: A novel framework
  based on deep reinforcement learning for face anti-spoofing,'' \emph{TIFS},
  vol.~16, pp. 937--951, 2020.

\bibitem{yu2021revisiting}
Z.~Yu, X.~Li, J.~Shi, Z.~Xia, and G.~Zhao, ``Revisiting pixel-wise supervision
  for face anti-spoofing,'' \emph{IEEE TBIOM}, 2021.

\bibitem{kim2020suppressing}
T.~Kim and Y.~Kim, ``Suppressing spoof-irrelevant factors for domain-agnostic
  face anti-spoofing,'' \emph{arXiv preprint arXiv:2012.01271}, 2020.

\bibitem{ojala2002multiresolution}
T.~Ojala, M.~Pietikainen, and T.~Maenpaa, ``Multiresolution gray-scale and
  rotation invariant texture classification with local binary patterns,''
  \emph{TPAMI}, vol.~24, no.~7, pp. 971--987, 2002.

\bibitem{li2019face}
L.~Li and X.~Feng, ``Face anti-spoofing via deep local binary pattern,'' in
  \emph{Deep Learning in Object Detection and Recognition}.\hskip 1em plus
  0.5em minus 0.4em\relax Springer, 2019, pp. 91--111.

\bibitem{chen2019cascade}
H.~Chen, Y.~Chen, X.~Tian, and R.~Jiang, ``A cascade face spoofing detector
  based on face anti-spoofing r-cnn and improved retinex lbp,'' \emph{IEEE
  Access}, 2019.

\bibitem{das2019new}
P.~K. Das, B.~Hu, C.~Liu, K.~Cui, P.~Ranjan, and G.~Xiong, ``A new approach for
  face anti-spoofing using handcrafted and deep network features,'' in
  \emph{SOLI}.\hskip 1em plus 0.5em minus 0.4em\relax IEEE, 2019.

\bibitem{menotti2015deep}
D.~Menotti, G.~Chiachia, A.~Pinto, W.~R. Schwartz, H.~Pedrini, A.~X. Falcao,
  and A.~Rocha, ``Deep representations for iris, face, and fingerprint spoofing
  detection,'' \emph{TIFS}, 2015.

\bibitem{li2017face}
L.~Li, Z.~Xia, L.~Li, X.~Jiang, X.~Feng, and F.~Roli, ``Face anti-spoofing via
  hybrid convolutional neural network,'' in \emph{FADS}.\hskip 1em plus 0.5em
  minus 0.4em\relax IEEE, 2017.

\bibitem{nagpal2019performance}
C.~Nagpal and S.~R. Dubey, ``A performance evaluation of convolutional neural
  networks for face anti spoofing,'' in \emph{IJCNN}.\hskip 1em plus 0.5em
  minus 0.4em\relax IEEE, 2019.

\bibitem{tu2017ultra}
X.~Tu and Y.~Fang, ``Ultra-deep neural network for face anti-spoofing,'' in
  \emph{International Conference on Neural Information Processing}.\hskip 1em
  plus 0.5em minus 0.4em\relax Springer, 2017.

\bibitem{rehman2017deep}
Y.~A.~U. Rehman, L.~M. Po, and M.~Liu, ``Deep learning for face anti-spoofing:
  An end-to-end approach,'' in \emph{SPA}.\hskip 1em plus 0.5em minus
  0.4em\relax IEEE, 2017, pp. 195--200.

\bibitem{lin2018live}
C.~Lin, Z.~Liao, P.~Zhou, J.~Hu, and B.~Ni, ``Live face verification with
  multiple instantialized local homographic parameterization.'' in
  \emph{IJCAI}, 2018.

\bibitem{de2018learning}
G.~B. de~Souza, J.~P. Papa, and A.~N. Marana, ``On the learning of deep local
  features for robust face spoofing detection,'' in \emph{SIBGRAPI}.\hskip 1em
  plus 0.5em minus 0.4em\relax IEEE, 2018.

\bibitem{rehman2018livenet}
Y.~A.~U. Rehman, L.~M. Po, and M.~Liu, ``Livenet: Improving features
  generalization for face liveness detection using convolution neural
  networks,'' \emph{Expert Systems with Applications}, vol. 108, pp. 159--169,
  2018.

\bibitem{luo2018face}
S.~Luo, M.~Kan, S.~Wu, X.~Chen, and S.~Shan, ``Face anti-spoofing with
  multi-scale information,'' in \emph{ICPR}, 2018.

\bibitem{larbi2018deepcolorfasd}
K.~Larbi, W.~Ouarda, H.~Drira, B.~B. Amor, and C.~B. Amar, ``Deepcolorfasd:
  Face anti spoofing solution using a multi channeled color spaces cnn,'' in
  \emph{SMC}.\hskip 1em plus 0.5em minus 0.4em\relax IEEE, 2018.

\bibitem{bresan2019facespoof}
R.~Bresan, A.~Pinto, A.~Rocha, C.~Beluzo, and T.~Carvalho, ``Facespoof buster:
  a presentation attack detector based on intrinsic image properties and deep
  learning,'' \emph{arXiv preprint arXiv:1902.02845}, 2019.

\bibitem{rehman2019face}
Y.~A.~U. Rehman, L.-M. Po, M.~Liu, Z.~Zou, W.~Ou, and Y.~Zhao, ``Face liveness
  detection using convolutional-features fusion of real and deep network
  generated face images,'' \emph{Journal of Visual Communication and Image
  Representation}, 2019.

\bibitem{laurensi2019style}
R.~Laurensi, A.~Israel, L.~T. Menon, N.~Penna, O.~Manoel~Camillo, A.~L.
  Koerich, and A.~S. Britto~Jr, ``Style transfer applied to face liveness
  detection with user-centered models,'' \emph{arXiv}, pp. arXiv--1907, 2019.

\bibitem{tu2020learning}
X.~Tu, Z.~Ma, J.~Zhao, G.~Du, M.~Xie, and J.~Feng, ``Learning generalizable and
  identity-discriminative representations for face anti-spoofing,'' \emph{ACM
  TIST}, vol.~11, no.~5, pp. 1--19, 2020.

\bibitem{8987415}
J.~{Guo}, X.~{Zhu}, J.~{Xiao}, Z.~{Lei}, G.~{Wan}, and S.~Z. {Li}, ``Improving
  face anti-spoofing by 3d virtual synthesis,'' in \emph{ICB}, 2019, pp. 1--8.

\bibitem{pinto2020leveraging}
A.~Pinto, S.~Goldenstein, A.~Ferreira, T.~Carvalho, H.~Pedrini, and A.~Rocha,
  ``Leveraging shape, reflectance and albedo from shading for face presentation
  attack detection,'' \emph{TIFS}, 2020.

\bibitem{zuo2020face}
Y.~Zuo, W.~Gao, and J.~Wang, ``Face liveness detection algorithm based on
  livenesslight network,'' in \emph{HPBD\&IS}.\hskip 1em plus 0.5em minus
  0.4em\relax IEEE, 2020.

\bibitem{chen2020face}
B.~Chen, W.~Yang, and S.~Wang, ``Face anti-spoofing by fusing high and low
  frequency features for advanced generalization capability,'' in
  \emph{MIPR}.\hskip 1em plus 0.5em minus 0.4em\relax IEEE, 2020, pp. 199--204.

\bibitem{parkin2020creating}
A.~Parkin and O.~Grinchuk, ``Creating artificial modalities to solve rgb
  liveness,'' \emph{arXiv preprint arXiv:2006.16028}, 2020.

\bibitem{huang2020deep}
Y.~Huang, W.~Zhang, and J.~Wang, ``Deep frequent spatial temporal learning for
  face anti-spoofing,'' \emph{arXiv preprint arXiv:2002.03723}, 2020.

\bibitem{ma2020novel}
Y.~Ma, L.~Wu, Z.~Li \emph{et~al.}, ``A novel face presentation attack detection
  scheme based on multi-regional convolutional neural networks,'' \emph{PR
  Letters}, vol. 131, pp. 261--267, 2020.

\bibitem{sun2020face2}
W.~Sun, Y.~Song, H.~Zhao, and Z.~Jin, ``A face spoofing detection method based
  on domain adaptation and lossless size adaptation,'' \emph{IEEE Access},
  2020.

\bibitem{zheng2021attention}
W.~Zheng, M.~Yue, S.~Zhao, and S.~Liu, ``Attention-based spatial-temporal
  multi-scale network for face anti-spoofing,'' \emph{TBIOM}, 2021.

\bibitem{chen2019towards}
Y.~Chen, T.~Wang, J.~Wang, P.~Shi, and H.~Snoussi, ``Towards good practices in
  face anti-spoofing: An image reconstruction based method,'' in
  \emph{CAC}.\hskip 1em plus 0.5em minus 0.4em\relax IEEE, 2019.

\bibitem{tu2019deep}
X.~Tu, H.~Zhang, M.~Xie, Y.~Luo, Y.~Zhang, and Z.~Ma, ``Deep transfer across
  domains for face antispoofing,'' \emph{Journal of Electronic Imaging}, 2019.

\bibitem{saha2020domain}
S.~Saha, W.~Xu, M.~Kanakis, S.~Georgoulis, Y.~Chen, D.~Pani~Paudel, and
  L.~Van~Gool, ``Domain agnostic feature learning for image and video based
  face anti-spoofing,'' in \emph{CVPRW}, 2020.

\bibitem{fatemifar2019spoofing}
S.~Fatemifar, S.~R. Arashloo, M.~Awais, and J.~Kittler, ``Spoofing attack
  detection by anomaly detection,'' in \emph{ICASSP}, 2019.

\bibitem{fatemifar2020stacking}
S.~Fatemifar, M.~Awais, A.~Akbari, and J.~Kittler, ``A stacking ensemble for
  anomaly based client-specific face spoofing detection,'' in
  \emph{ICIP}.\hskip 1em plus 0.5em minus 0.4em\relax IEEE, 2020.

\bibitem{fatemifar2020client}
S.~Fatemifar, S.~R. Arashloo, M.~Awais, and J.~Kittler, ``Client-specific
  anomaly detection for face presentation attack detection,'' \emph{Pattern
  Recognition}, 2020.

\bibitem{kowalski2020study}
M.~Kowalski, ``A study on presentation attack detection in thermal infrared,''
  \emph{Sensors}, vol.~20, no.~14, p. 3988, 2020.

\bibitem{wang2019multi}
G.~Wang, C.~Lan, H.~Han, S.~Shan, and X.~Chen, ``Multi-modal face presentation
  attack detection via spatial and channel attentions,'' in \emph{CVPRW}, 2019.

\bibitem{li2019dual}
L.~Li, Z.~Gao, L.~Huang, H.~Zhang, and M.~Lin, ``A dual-modal face
  anti-spoofing method via light-weight networks,'' in \emph{ASID}.\hskip 1em
  plus 0.5em minus 0.4em\relax IEEE, 2019.

\bibitem{li2020face}
X.~Li, W.~Wu, T.~Li, Y.~Su, and L.~Yang, ``Face liveness detection based on
  parallel cnn,'' in \emph{Journal of Physics: Conference Series}.\hskip 1em
  plus 0.5em minus 0.4em\relax IOP Publishing, 2020.

\bibitem{george2020can}
A.~George and S.~Marcel, ``Can your face detector do anti-spoofing? face
  presentation attack detection with a multi-channel face detector,''
  \emph{arXiv preprint arXiv:2006.16836}, 2020.

\bibitem{yang2020pipenet}
Q.~Yang, X.~Zhu, J.-K. Fwu, Y.~Ye, G.~You, and Y.~Zhu, ``Pipenet: Selective
  modal pipeline of fusion network for multi-modal face anti-spoofing,'' in
  \emph{CVPRW}, 2020.

\bibitem{te2020exploring}
G.~Te, W.~Hu, and Z.~Guo, ``Exploring hypergraph representation on face
  anti-spoofing beyond 2d attacks,'' in \emph{ICME}.\hskip 1em plus 0.5em minus
  0.4em\relax IEEE, 2020.

\end{thebibliography}

%

\begin{IEEEbiography}[{\includegraphics[width=1in,height=1.25in,clip,keepaspectratio]{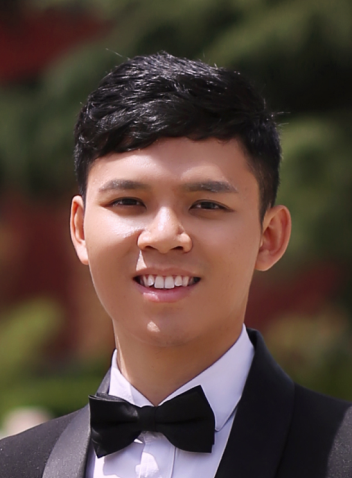}}]{Zitong Yu}
 received the M.S. degree in multimedia from University of Nantes, France, in 2016, and he received the Ph.D. degree in computer science from University of Oulu, Finland, in 2022. His research interests include face anti-spoofing, remote physiological measurement and video understanding. He led the team and won the 1st Place in the ChaLearn multi-modal face anti-spoofing attack detection challenge with CVPR 2020.
\end{IEEEbiography}

\begin{IEEEbiography}[{\includegraphics[width=1in,height=1.25in,clip,keepaspectratio]{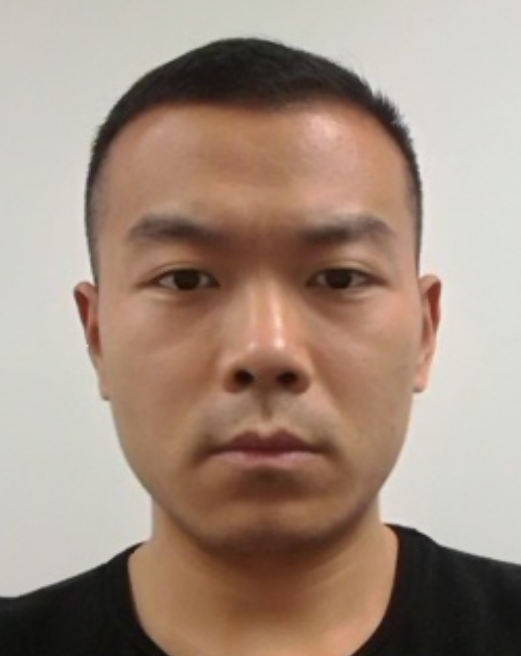}}]{Yunxiao Qin} received the M.S. degree in control theory and control engineering from the School of Marine Science and Technology, Northwestern Polytechnical University, Xi’an, China, in 2015, and he received the Ph.D. degree in control science and engineering from the School of Automation, Northwestern Polytechnical University, Xi’an, China, in 2021. His current research interests include meta-learning, face anti-spoofing, and deep reinforcement learning.
	\end{IEEEbiography}

\begin{IEEEbiography}[{\includegraphics[width=1in,height=1.25in,clip,keepaspectratio]{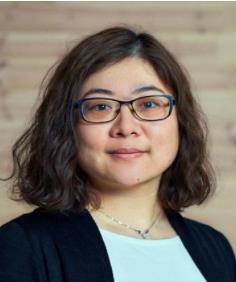}}]{Xiaobai Li}
received her B.Sc. degree in Psychology from Peking University, M.Sc. degree in Biophysics from the Chinese Academy of Science, and Ph.D. degree in Computer Science from University of Oulu. She is currently an assistant professor in the Center for Machine Vision and Signal Analysis of University of Oulu. Her research of interests includes facial expression recognition, micro-expression analysis, remote physiological signal measurement from facial videos, and related applications in affective computing, healthcare and biometrics. She is an associate editor for IEEE TCSVT, Frontiers in Psychology, and Image and Vision Computing. Dr. Li was a co-chair of several international workshops in CVPR, ICCV, FG and ACM Multimedia. 
\end{IEEEbiography}

\begin{IEEEbiography}[{\includegraphics[width=1in,height=1.25in,clip,keepaspectratio]{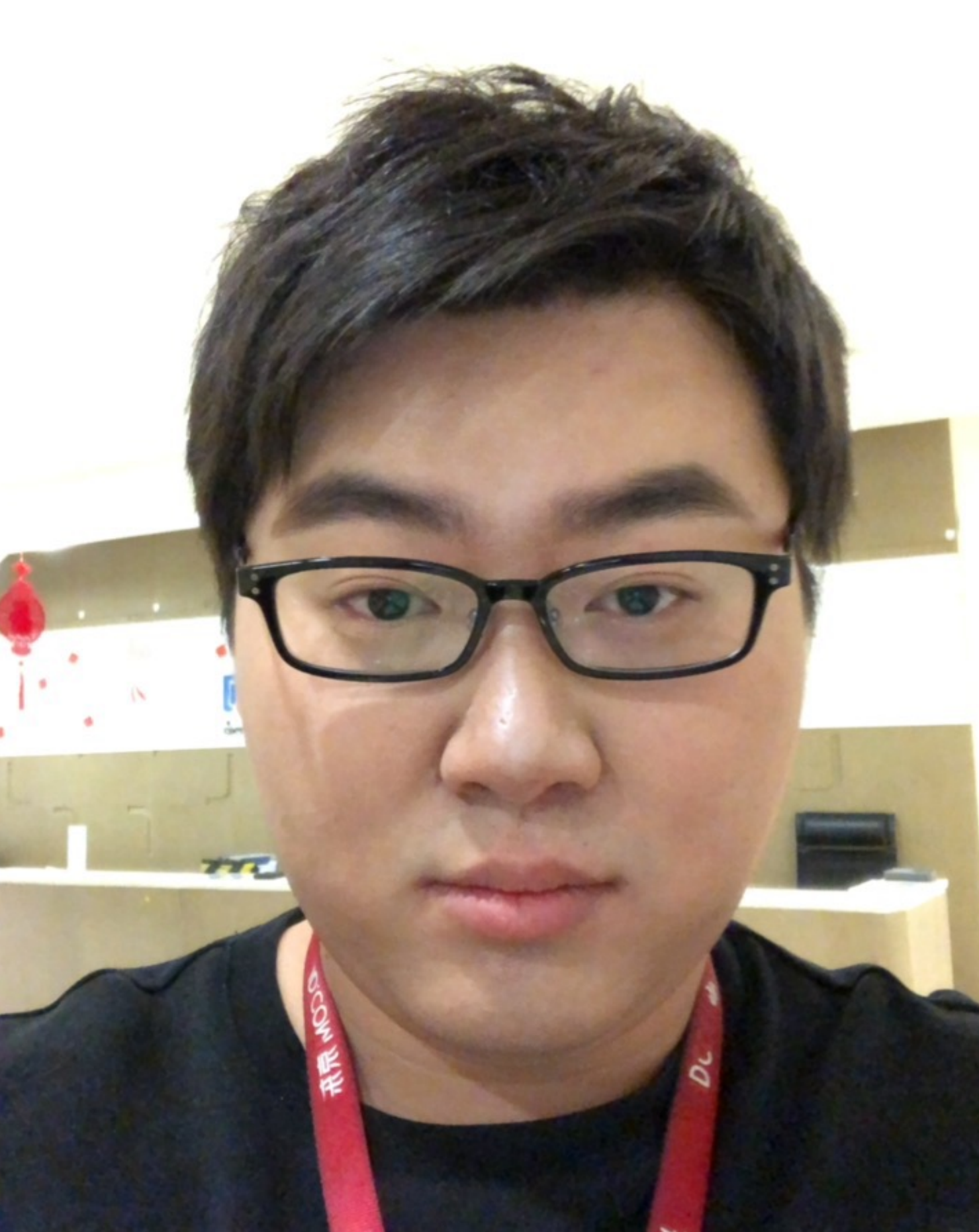}}]{Chenxu Zhao}
 received M.S. Degree from Beihang University, Beijing, China, in 2016, and was in the joint programme with National Laboratory of Pattern Recognition (NLPR) Laboratory of Institute of Automation, Chinese Academy of Sciences, from 2014 to 2016.
 He is currently served as a Co-Founder in SailYond Technology, Beijing, China.
 He served as a Research Director in MiningLamp Technology, Beijing, China. 
 His major research areas include face analysis, anomaly detection and meta-learning.
\end{IEEEbiography}



\begin{IEEEbiography}[{\includegraphics[width=1in,height=1.25in,clip,keepaspectratio]{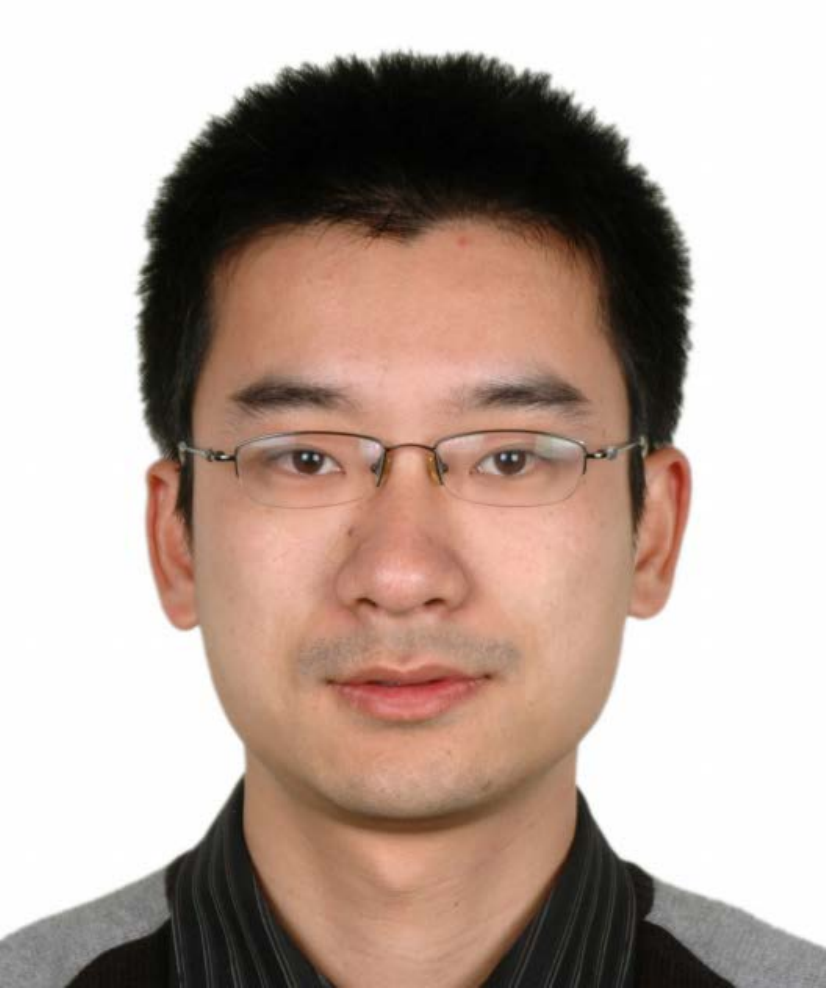}}]{Zhen Lei}
received the B.S. degree in automation from the University of Science and Technology of China, in 2005, and the Ph.D. degree from the Institute of Automation, Chinese Academy of Sciences, in 2010, where he is currently a Professor. He is IAPR Fellow and AAIA Fellow. He has published over 200 papers in international journals and conferences with 21000+ citations in Google Scholar and h-index 71. He was competition co-chair of IJCB2022 and has served as area chairs for server conferences and is associate editor for IEEE Trans. on Information Forensics and Security, Pattern Recognition, Neurocomputing and IET Computer Vision journals. His research interests are in computer vision, pattern recognition, image processing, and face recognition in particular. He is the winner of 2019 IAPR Young Biometrics Investigator Award.  

\end{IEEEbiography}

\begin{IEEEbiography}[{\includegraphics[width=1in,height=1.25in,clip,keepaspectratio]{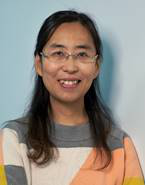}}]{Guoying Zhao}
(IEEE Fellow 2022) received the Ph.D. degree in computer science from the Chinese Academy of Sciences, Beijing, China, in 2005. She is currently an Academy Professor and full Professor (tenured in 2017) with University of Oulu. She is also a visiting professor with Aalto University. She is a member of Finnish Academy of Sciences and Letters, IAPR Fellow and AAIA Fellow. She has authored or co-authored more than 280 papers in journals and conferences with 20100+ citations in Google Scholar and h-index 66. She is panel chair for FG 2023, was co-program chair for ACM International Conference on Multimodal Interaction (ICMI 2021), co-publicity chair for FG2018, and has served as area chairs for several conferences and was/is associate editor for IEEE Trans. on Multimedia, Pattern Recognition, IEEE Trans. on Circuits and Systems for Video Technology, and Image and Vision Computing Journals. Her current research interests include image and video descriptors, facial-expression and micro-expression recognition, emotional gesture analysis, affective computing, and biometrics. Her research has been reported by Finnish TV programs, newspapers and MIT Technology Review.
\end{IEEEbiography}

\begin{table*}
\centering
\caption{A summary of existing surveys in FAS. Most of them focus on handcrafted feature based methods under single RGB modality, and investigate limited number of public datasets as well as evaluation protocols. `DL' and `M\&H' is short for `deep learning' and `modality \& hardware', respectively. `VIS', `NIR', `SWIR',`LF', and `Polarized' denotes using commercial visible RGB, near infrared, short-wave infrared, light field, and four-directional polarized camera, respectively.} \label{tab:surveys}
\resizebox{1.0\textwidth}{!} {\begin{tabular}{c c c c c c} 
 \toprule[1pt]
 Title \& Reference & Year & DL & M\&H & Datasets & Protocol \\
 \midrule
 2D Face Liveness Detection: an Overview~\cite{kahm20122d} & 2014 & No & VIS  & 2 & Intra-dataset intra-type\\

 \midrule
 \tabincell{c}{Face Biometrics under Spoofing Attacks: Vulnerabilities, Countermeasures, \\Open Issues and Research Directions}~\cite{hadid2014face} & 2014 & No & VIS  & 4 & Intra-dataset intra-type \\

 \midrule
 Biometric Anti-spoofing Methods: A Survey in Face Recognition~\cite{galbally2014biometric} & 2015 & No & VIS  & 6 & Intra-dataset intra-type\\
 

   \midrule
A Comparative Study on Face Spoofing Attacks~\cite{kumar2017comparative} & 2017 & No & VIS  & 9 & Intra-dataset intra-type\\ 
 
  \midrule
 Face Spoofing and Counter-Spoofing: A Survey of State-of-the-art Algorithms~\cite{kisku2017face} & 2017 & No & VIS  & 6 & \tabincell{c}{Intra-dataset intra-type,\\Cross-dataset intra-type}\\ 
 
  \midrule
 \tabincell{c}{Presentation Attack Detection Methods for Face Recognition Systems: \\A Comprehensive Survey}~\cite{ramachandra2017presentation} & 2017 & No & VIS  & 11 & Intra-dataset intra-type\\ 

  \midrule
 How Far Did We Get in Face Spoofing Detection?~\cite{souza2018far} & 2018 & \tabincell{c}{Yes\\(Few, $\textless$10)} & VIS  & 9 & \tabincell{c}{Intra-dataset intra-type,\\Cross-dataset intra-type}\\

  \midrule
 Insight on Face Liveness Detection: A Systematic Literature Review~\cite{raheem2019insight} & 2019 & No & VIS  & 14 & Intra-dataset intra-type\\

  \midrule
 The Rise of Data-driven Models in Presentation Attack Detection~\cite{pereira2020rise} & 2019 & \tabincell{c}{Yes\\(Few, $\textless$10)} & VIS  & 7 & \tabincell{c}{Intra-dataset intra-type,\\Cross-dataset intra-type}\\ 

  \midrule
 A Survey on 3D Mask Presentation Attack Detection and Countermeasures~\cite{jia2020survey} & 2020 & \tabincell{c}{Yes\\(Few, $\textless$10)} & VIS  & 10 & Intra-dataset intra-type\\

  \midrule
 \tabincell{c}{Deep Convolutional Neural Networks for Face and Iris Presentation Attack \\Detection: Survey and Case Study}~\cite{el2020deep} & 2020 & \tabincell{c}{Yes\\(Few, $\textless$30)} & VIS  & 8 & \tabincell{c}{Intra-dataset intra-type,\\Cross-dataset intra-type}\\

 \midrule
 \tabincell{c}{A Survey On Anti-Spoofing Methods For Face Recognition with\\ RGB Cameras of Generic Consumer Devices}~\cite{ming2020survey} & 2020 & \tabincell{c}{Yes\\(Few, $\textless$50)} & VIS  & 12 & \tabincell{c}{Intra-dataset intra-type,\\Cross-dataset intra-type}\\

  \midrule
 \textbf{Deep Learning for Face Anti-spoofing: A Survey (Ours)} & 2022 & \tabincell{c}{Yes\\(Full, $\textgreater$100)} & \tabincell{c}{VIS, Flash,\\NIR, Thermal, \\Depth, SWIR, \\LF, Polarized}  & 36 & \tabincell{c}{Intra-dataset intra-type,\\Cross-dataset intra-type,\\Intra-dataset cross-type,\\Cross-dataset cross-type}\\ 

 \bottomrule[1pt]
 \end{tabular}}
\end{table*}

\begin{table*}[!htb]
	\centering	
	\small
	
\caption{ ACER (\%) results of the intra-dataset intra-type testings on OULU-NPU (4 sub-protocols) and SiW (3 sub-protocols) datasets for common deep learning methods with binary cross-entropy
supervision and pixel-wise supervision.} 
\vspace{-0.8em}	
	\scalebox{0.95}{\begin{tabular}{l|c|c|c|c|c|c|c|c|c}
		\toprule[1pt]
		 & \multirow{2}{*}{Method } & \multirow{2}{*}{Venue } & \multicolumn{4}{c|}{OULU-NPU} & \multicolumn{3}{c}{SiW} 
		\\ \cline{4-10} 		
	 &  &	& Prot. 1   & Prot. 2  & Prot. 3  & Prot. 4    & Prot. 1   & Prot. 2   & Prot. 3                             
	\\ \cline{1-10}

   & STASN~\cite{yang2019face} & CVPR'19 & 1.9  & 2.2  & 2.8$\pm$1.6  & 7.5$\pm$4.7  & 1.00  & 0.28$\pm$0.05  & 12.10$\pm$ 1.50
   \\ \cline{2-10}

   & TSCNN~\cite{chen2019attention} & TIFS'19 & 5.9  & 4.9  & 5.6$\pm$1.6  & 9.8$\pm$4.2  & -  & -  & -
   \\ \cline{2-10}
   
   \quad Binary 
    & CIFL~\cite{chen2021camera} & TIFS'21 & 3.4  & 2.4  & 2.5$\pm$0.8  & 6.1$\pm$4.1  & -  & -  & -
   \\ \cline{2-10}
   
    Cross-entropy 
   & DRL-FAS~\cite{cai2020drl} & TIFS'20 & 4.7  & 1.9  & 3.0$\pm$1.5  & 7.2$\pm$3.9  & \textbf{0.00}  & \textbf{0.00$\pm$0.00}  & 4.51$\pm$ 0.00
   \\ \cline{2-10}
   
   Supervision
   & SSR-FCN~\cite{deb2020look} & TIFS'20 &  4.6 & 3.4  & 2.8 $\pm$2.2  & 10:8$\pm$ 5:1  & -  & -  & -
   \\ \cline{2-10}

    & FasTCo~\cite{xu2020improving} & ICCVW'21 & 0.8  &  \textbf{1.1} & \textbf{1.1$\pm$0.8}  & \textbf{1.5$\pm$1.2}  & 0.0003  & 0.01$\pm$0.01  & 2.00$\pm$0.56
   \\ \cline{2-10}
    
   & PatchNet~\cite{wang2022patchnet} & CVPR'22 & \textbf{0.0}  & 1.2  & 1.18$\pm$1.26  & 2.9$\pm$3.0  & \textbf{0.00}  & \textbf{0.00$\pm$0.00}  & 2.45$\pm$ 0.45    
     \\ \cline{1-10}
   
   & Auxiliary~\cite{Liu2018Learning} & CVPR'18  & 1.6  & 2.7  & 2.9$\pm$1.5  & 9.5$\pm$6.0  & 3.58  & 0.57$\pm$0.69  & 8.31$\pm$3.81
    \\ \cline{2-10}
    
    & PixBiS~\cite{george2019deep} & IJCB'19  & 0.4  & 6.0  & 11.1$\pm$9.4  & 25.0$\pm$12.7  & -  & - & -
    \\ \cline{2-10}
    
    & FAS-SGTD~\cite{wang2020deep} & CVPR'20  & 1.0  & 1.9  & 2.7$\pm$0.6  & 5.0$\pm$2.2  & 0.40  & 0.02$\pm$ 0.04 & 2.78$\pm$ 3.57
    \\ \cline{2-10}

   Pixel-wise
   & De-Spoof~\cite{jourabloo2018face} & ECCV'20  & 1.5  & 4.3  & 3.6$\pm$1.6  & 5.6$\pm$5.7  & -  & -  & -
    \\ \cline{2-10}
    
    Supervision
   & Disentangled~\cite{zhang2020face} & ECCV'20  & 1.3  & 2.4  & 2.2$\pm$ 2.2  & 4.4$\pm$ 3.0  & 0.28  & 0.10$\pm$ 0.04  & 5.59$\pm$ 4.37
    \\ \cline{2-10}
    
   & STDN~\cite{liu2020disentangling} & ECCV'20  & 1.1  & 1.9  & 2.8$\pm$ 3.3  & 3.8$\pm$ 4.2  & \textbf{0.00}  & \textbf{0.00$\pm$ 0.00}  & 7.9$\pm$ 3.3
    \\ \cline{2-10}
    
    & BCN~\cite{yu2020face} & ECCV'20  & 0.8  & 1.7  & 2.5$\pm$ 1.1  & 5.2$\pm$ 3.7  & 0.36  & 0.11$\pm$0.08  & 2.45$\pm$ 0.68
    \\ \cline{2-10}

    & CDCN~\cite{yu2020searching} & CVPR'20  & 1.0  & 1.5  & 2.3$\pm$ 1.4  & 6.9$\pm$ 2.9  & 0.12  & 0.06$\pm$ 0.04  & 1.71$\pm$ 0.11
    \\ \cline{2-10}
    
    & DC-CDN~\cite{yu2021dual} & IJCAI'21  & 0.4  & 1.3  & 1.9$\pm$ 1.1  & 4.0$\pm$ 3.1  & -  & -  & -
    \\ \cline{2-10}
   
   & NAS-FAS~\cite{yu2020fas2} & PAMI'21 & 0.2  & 1.2  & 1.7$\pm$0.6  & 2.9$\pm$2.8  & 0.12  & 0.04$\pm$0.05  & \textbf{1.52$\pm$0.13}
   \\ \cline{2-10}
   
      & MT-FAS~\cite{qin2021meta} & PAMI'21  &  0.4 & 1.4  & 2.1$\pm$ 1.7  & 3.7$\pm$ 2.9  & -  & - & -

    \\ 
    
	\bottomrule[1pt]	
	\end{tabular}}
	
    \label{tab:semi_DG_intra}
    \vspace{-0.2em}
\end{table*}

\begin{table*}[!htb]
	\centering	
	\small
	
\caption{ HTER (\%) results of the cross-dataset intra-type testings among OULU-NPU (O), CASIA-MFSD (C), Replay-Attack (I), and MSU-MFSD (M) datasets with different numbers of source domains for training. For example, `C to I' means training on CASIA-MFSD and then testing on Replay-Attack.} 
\vspace{-0.8em}	
	\scalebox{0.83}{\begin{tabular}{l|c|c|c|c|c|c|c|c|c|c}
		\toprule[1pt]
		 & \multirow{2}{*}{Method } & \multirow{2}{*}{Venue } & \multicolumn{2}{c|}{1 source domain} & \multicolumn{2}{c|}{2 source domains} & \multicolumn{4}{c}{3 source domains} 
		\\ \cline{4-11} 		
	 &  &	& C to I   & I to C  & M\&I to C  & M\&I to O   & O\&C\&I to M   & O\&M\&I to C  & O\&C\&M to I  & I\&C\&M to O                             
	\\ \cline{1-11}

   & Auxiliary~\cite{Liu2018Learning} & CVPR'18 &  27.6 & 28.4  &  - & -  & - & 28.4  & 27.6 & -
   \\ \cline{2-11}
   
   & CDCN~\cite{yu2020searching} & CVPR'20 &  15.5  & 32.6  &  - & -  & - & -  & - & -
   \\ \cline{2-11}
    Traditional
   & FAS-SGTD~\cite{wang2020deep} & CVPR'20 &  17.0 & \textbf{22.8}  &  - & -  & - & -  & - & -
   \\ \cline{2-11}
   
    \quad Deep 
  & BCN~\cite{yu2020face} & ECCV'20 &   16.6  &  36.4 & -  & - & 19.81 & 25.12 & 22.75 & 21.24
   \\ \cline{2-11}    
    
    Learning
   & NAS-FAS~\cite{yu2020fas2} & PAMI'21 &  - & -  &  - & -  & 16.85 & 15.21  & 11.63 & 13.16
   \\ \cline{2-11}

    & MT-FAS~\cite{qin2021meta} & PAMI'21 &  - & -  &  - & -  & 11.67 & 18.44  & 11.93 & 16.23
   \\ \cline{2-11}

   & PS~\cite{yu2021revisiting} & TBIOM'21 & 13.8 & 31.3  &  - & -  & 20.42 & 18.25  & 19.55 & 15.76
   \\ \cline{2-11}
   
   & DC-CDN~\cite{yu2021dual} & IJCAI'21 & \textbf{6.0} & 30.1  &  - & -  & 25.31 & 15.00  & 15.88 & 18.82
   \\ \cline{1-11}

   & MADDG~\cite{shao2019multi} & CVPR'19  &  - & -  & 41.02 & 39.35 & 17.69 & 24.50  & 22.19 & 27.98
    \\ \cline{2-11}
    
    & PAD-GAN~\cite{wang2020cross} & CVPR'20  &  - & -  & 31.67 & 34.02 & 17.02 & 19.68  & 20.87 & 25.02
    \\ \cline{2-11}

   & RF-Meta~\cite{shao2019regularized} & AAAI'20  &  - & -  &  - & -  & 13.89 & 20.27  & 17.30 & 16.45
    \\ \cline{2-11}

   & SSDG~\cite{jia2020single} & CVPR'20  &  - & -  & 31.89 & 36.01 & 7.38 &  10.44  & 11.71 & 15.61
    \\ \cline{2-11}
    
   Generalized
   & SDA~\cite{wang2021self} & AAAI'21  &   - & -  &  - & -  & 15.40 & 24.50  & 15.60 & 23.10
    \\ \cline{2-11}
    
    \quad Deep
   & D$^2$AM~\cite{chen2021generalized} & AAAI'21  &  - & -  &  - & -  & 12.70 & 20.98  & 15.43 & 15.27
    \\ \cline{2-11}
    
    Learning
      & DASN~\cite{kim2020suppressing} & Access'21  &   - & -  &  - & -  & 8.33 & 12.04  & 13.38 & \textbf{11.77}
    \\ \cline{2-11}

    & DRDG~\cite{liu2021dual} & IJCAI'21  &  - & -  &  31.28 & 33.35 & 12.43 & 19.05  & 15.56 & 15.63
    \\ \cline{2-11}
    
    & ANRL~\cite{liu2021adaptive} & MM'21  &  - & -  & 31.06 & 30.73 & 10.83 & 17.83  & 16.03 & 15.67
     \\ \cline{2-11}
    
    & FGHV~\cite{liu2022feature} & AAAI'22  &  - & -  & - & - & 9.17 & 12.47  & 16.29 & 13.58
    
    \\ \cline{2-11}
    
    & SSAN~\cite{wang2022domain} & CVPR'22  &  - & -  &  \textbf{30.00} & \textbf{29.44} & \textbf{6.67} & \textbf{10.00}  & \textbf{8.88} & 13.72

    \\ 
    
	\bottomrule[1pt]	
	\end{tabular}}
	
    \label{tab:semi_DG_intra}
    \vspace{-0.2em}
\end{table*}

\begin{table*}
\centering

\caption{EER (\%) results of the Intra-dataset cross-type testings on SiW-M with the leave-one-type-out setting.}

\scalebox{0.78}{\begin{tabular}{l|c|c|c|c|c|c|c|c|c|c|c|c|c|c|c|c}
\toprule[1pt]

&\multirow{2}{*}{Method} &\multirow{2}{*}{Venue} &\multirow{2}{*}{Replay} &\multirow{2}{*}{Print} &\multicolumn{5}{c|}{Mask Attacks} &\multicolumn{3}{c|}{Makeup Attacks}&\multicolumn{3}{c|}{Partial Attacks} &\multirow{2}{*}{Average} \\
\cline{6-16} &  &  &  & &  \tabincell{c}{Half} &\tabincell{c}{Silicone} &\tabincell{c}{Trans.} &\tabincell{c}{Paper}&\tabincell{c}{Manne.}&\tabincell{c}{Obfusc.}&\tabincell{c}{Im.}&\tabincell{c}{Cos.}&\tabincell{c}{Fun.} & \tabincell{c}{Glasses} &\tabincell{c}{Partial} & \\



\cline{1-17}

& Auxiliary~\cite{Liu2018Learning} & CVPR'18 & 14.0 & 4.3 & 11.6  & 12.4 & 24.6 & 7.8 & 10.0 & 72.3 & 10.1 & \textbf{9.4} & 21.4 & 18.6 & 4.0 & 17.0$\pm$17.7 \\

\cline{2-17}

& CDCN~\cite{yu2020searching} & CVPR'20 & 8.2 & 7.8 & 8.3 & \textbf{7.4} & 20.5 & 5.9 & 5.0 & 47.8 & 1.6 & 14.0 & 24.5 & 18.3 & 1.1 & 13.1$\pm$ 12.6 \\

\cline{2-17}

& STDN ~\cite{liu2020disentangling} & ECCV'20 & 7.6 & 3.8 & 8.4 & 13.8 & 14.5 & \textbf{5.3} & 4.4 & 35.4 & \textbf{0.0} & 19.3 & 21.0 & 20.8 & 1.6 & 12.0$\pm$ 10.0 \\

\cline{2-17}
Traditional
& BCN ~\cite{yu2020face} & ECCV'20 & 13.4 & 5.2 & 8.3 & 9.7 & 13.6 & 5.8 & 2.5 & 33.8 & \textbf{0.0} & 14.0 & 23.3 & 16.6 & 1.2 & 11.3$\pm$ 9.5 \\

\cline{2-17}
\quad Deep
& SSR-FCN~\cite{deb2020look} & TIFS'20 & \textbf{6.8} & 11.2 & \textbf{2.8} & 6.3 & 28.5 & 0.4 & 3.3 & 17.8 & 3.9 & 11.7 & 21.6 & 13.5 & 3.6 & 10.1$\pm$ 8.4 \\

\cline{2-17}
Learning
& PS~\cite{yu2021revisiting} & TBIOM'21 & 10.3 & 7.8 & 8.3 & 7.4 & 10.2 & 5.9 & 5.0 & 43.4 & \textbf{0.0} & 12.0 & 23.9 & 15.9 & \textbf{0.0} & 11.5$\pm$11.4 \\

\cline{2-17}
& NAS-FAS~\cite{yu2021revisiting} & PAMI'21 & 10.3 & 7.8 & 8.3 & 7.4 & 10.2 & 5.9 & 5.0 & 43.4 & \textbf{0.0} & 12.0 & 23.9 & 15.9 & \textbf{0.0} & 11.5$\pm$11.4 \\

\cline{2-17}

& DC-CDN~\cite{yu2021dual} & IJCAI'21 & 10.3 & 8.7 & 11.1  & 7.4 & 12.5 & 5.9 & \textbf{0.0} & 39.1 & \textbf{0.0} & 12.0 & 18.9 & 13.5 & 1.2 & 10.8$\pm$10.1 \\

\cline{2-17}

& MT-FAS~\cite{qin2021meta} & PAMI'21 & 7.8 & 4.4 & 11.2  & 5.8 & 11.2 & 2.8 & 2.7 & 38.9 & 0.2 & 10.1 & 20.5 & 18.9 & 1.3 & 10.4$\pm$10.2 \\

\cline{2-17}

& ViTranZFAS~\cite{george2020effectiveness} & IJCB'21 & 15.2 & 5.8 & 5.8  & \textbf{4.9} & \textbf{5.9} & \textbf{0.1} & 3.2 & \textbf{9.8} & 0.4 & 10.7 & 20.1 & \textbf{2.9} & 1.9 & \textbf{6.7$\pm$5.6} \\

\cline{1-17}

& DTN~\cite{liu2019deep}
 & CVPR'19 & 10.0 & \textbf{2.1} & 14.4 & 18.6 & 26.5 & 5.7 & 9.6 & 50.2 & 10.1 & 13.2 & 19.8 & 20.5 & 8.8 & 16.1$\pm$ 12.2 \\

\cline{2-17}
Generalized
& Hypersphere~\cite{li2020unseen}
 & ICASSP'20 & 13.2 & 14.0 & 18.1  & 24.0 & 12.4 & 3.1 & 6.2 & 34.8 & 3.1 & 16.3 & 21.4 & 21.7 & 9.3 & 15.2$\pm$ 9.0 \\

\cline{2-17}
Learning
& FGHV~\cite{wang2022domain} & AAAI'22 & 9.0 & 8.0 & 5.9 & 9.9 & 14.3 & 3.7 & 4.8 & 19.3 & 2.0 & 9.2 & 18.9 & 8.5 & 4.7 & 9.1$\pm$ 5.4 \\

\bottomrule[1pt]

\end{tabular}
}

\label{tab:SiW-M}
\end{table*}

\begin{table*}
\centering
\caption{Summary of the \textbf{hybrid (handcraft+deep learning)} FAS methods with \textbf{binary cross-entropy supervision}. `S/D', `CE', `OF', `OFM', `NN', `HOG', `LBP' are short for `Static/Dynamic', `cross-entropy', `optical flow', `optical flow magnitude', `nearest neighbor', `histogram of oriented gradients~\cite{dalal2005histograms}' and `local binary pattern~\cite{ojala2002multiresolution}', respectively.} \label{tab:handcrafted}
\resizebox{1.0\textwidth}{!} {\begin{tabular}{l c c c c c c} 
 \toprule[1pt]
 Method & Year & Backbone & Loss & Input & S/D & Description \\

   \midrule
DPCNN~\cite{Li2017An} & 2016 & VGG-Face & Trained with SVM & RGB & S &  deep partial features with Blocks PCA  \\  

   \midrule
Multi-cues+NN~\cite{feng2016integration} & 2016 & MLP & Binary CE loss & \tabincell{c}{RGB+OFM} & D & fused features from image quality cues and motion cues  \\

   \midrule
CNN LBP-TOP~\cite{asim2017cnn} & 2017 & 5-layer CNN & \tabincell{c}{Binary CE loss\\SVM} & RGB & D & \tabincell{c}{cascading LBP-TOP with CNN to extract \\discriminative spatio-temporal features}  \\  

   \midrule
DF-MSLBP~\cite{cai2019learning} & 2018 & Deep forest & Binary CE loss & HSV+YCbCr & S & multi-scale LBP based Tree-Ensembled features  \\

   \midrule
SPMT+SSD~\cite{song2019discriminative} & 2018 & VGG16 & \tabincell{c}{Binary CE loss\\SVM\\bbox regression} & \tabincell{c}{RGB\\Landmarks} & S & \tabincell{c}{hand-crafted texture\&depth features cascaded \\with fast deep face spoofing detector}  \\

   \midrule
CHIF~\cite{agarwal2019chif} & 2019 & VGG-Face & Trained with SVM & RGB & S & \tabincell{c}{convoluted histogram image features for fine-\\grained mask texture representation}  \\  

   \midrule
DeepLBP~\cite{li2019face} & 2019 & VGG-Face & \tabincell{c}{Binary CE loss\\SVM} & \tabincell{c}{RGB,HSV,\\YCbCr} & S & \tabincell{c}{extracted the handcrafted features from the\\ convolutional responses of the fine-tuned CNN model}  \\

   \midrule
\tabincell{c}{CNN+LBP\\+WLD}~\cite{khammari2019robust} & 2019 & CaffeNet & Binary CE loss & RGB & S & \tabincell{c}{combined CNN features with LBP/WLD for preserving \\both semantic feature and local information}  \\  

   \midrule
Intrinsic~\cite{li20203d} & 2019 & 1D-CNN & Trained with SVM & Reflection & D & deep temporal cues from reflection intensity histogram  \\  

   \midrule
FARCNN~\cite{chen2019cascade} & 2019 & \tabincell{c}{Multi-scale \\attentional CNN}  & \tabincell{c}{Regression loss\\Crystal loss\\ Center loss} & RGB & S & \tabincell{c}{cascade detector features with \\improved Retinex based LBP}  \\  

   \midrule
CNN-LSP~\cite{li2019replayed} & 2019 & 1D-CNN & Trained with SVM & RGB & D &  \tabincell{c}{joint learned temporal features with attentional \\spatial regions and channels from magnified videos}  \\  

   \midrule
DT-Mask~\cite{shao2018joint} & 2019 & VGG16  & \tabincell{c}{Binary CE loss\\Channel\&Spatial-\\ discriminability} & RGB+OF & D & \tabincell{c}{joint learned discriminative features with \\attentional spatial regions and channels}  \\  

   \midrule
VGG+LBP~\cite{das2019new} & 2019 & VGG16 & Binary CE loss & RGB & S &  \tabincell{c}{combining deep CNN features, and LBP features\\ from brightness and chrominance channels}  \\  

   \midrule
CNN+OVLBP~\cite{sharifi2019score} & 2019 & VGG16 & \tabincell{c}{Binary CE loss\\NN classifier} & RGB & S &  \tabincell{c}{hybrid decisions using majority vote of CNN, over-\\lapped histograms of LBP and their fused vector}  \\  

   \midrule
HOG-Pert.~\cite{rehman2019perturbing} & 2019 & Multi-scale CNN & Binary CE loss & \tabincell{c}{RGB+HOG} & S & hybrid convolutional features and HOG features \\

   \midrule
LBP-Pert.~\cite{rehman2020enhancing} & 2020 & Multi-scale CNN & Binary CE loss & \tabincell{c}{RGB+LBP} & S & discriminative features enhanced by LBP perturbation \\ 

   \midrule
TransRPPG~\cite{yu2021transrppg} & 2021 & Vision Transformer & Binary CE loss & \tabincell{c}{rPPG map} & D & intrinsic liveness features via fully attentional transformer \\

 \bottomrule[1pt]
 \end{tabular}}
\end{table*}

\begin{table*}
\centering
\caption{Summary of the representative \textbf{traditional} deep learning} based FAS methods with \textbf{binary cross-entropy supervision}. `S/D' and `CE' are short for `Static/Dynamic' and `cross-entropy', respectively. `Reflect.', `OF', and `RP' denote the generated reflection map, optical flow, and rank pooling, respectively. \label{tab:binaryloss}
\resizebox{1.0\textwidth}{!} {\begin{tabular}{l c c c c c c} 
 \toprule[1pt]
 Method & Year & Backbone & Loss & Input & S/D & Description \\

   \midrule
CNN1~\cite{yang2014learn} & 2014 & 8-layer CNN & Trained with SVM & RGB & S & deep features from different spatial scales  \\

   \midrule
LSTM-CNN~\cite{Xu2016Learning} & 2015 & CNN+LSTM  & Binary CE loss & RGB & D & long-range local and dense features from sequence \\

   \midrule
SpoofNet~\cite{menotti2015deep} & 2015 & 2-layer CNN  & Binary CE loss & RGB & S & deep representation with architecture optimization \\ 

   \midrule
HybridCNN~\cite{li2017face} & 2017 & VGG-Face  & Trained with SVM & RGB & S & hybrid CNN for both global face and facial patches \\

   \midrule
CNN2~\cite{nagpal2019performance} & 2017 & VGG11 & Binary CE loss & RGB & S & model trained with continuous data-randomization  \\

   \midrule
Ultra-Deep~\cite{tu2017ultra} & 2017 & ResNet50+LSTM & Binary CE loss & RGB & D & ultra-deep features with rich long-range temporal context \\ 

    \midrule
FASNet~\cite{lucena2017transfer} & 2017 & VGG16  & Binary CE loss & RGB & S & transfer learned features based on a pre-trained CNN \\

   \midrule
CNN3~\cite{rehman2017deep} & 2018 & Inception, ResNet & Binary CE loss & RGB & S & transferred deep feature  \\

  \midrule
 MILHP~\cite{lin2018live} & 2018 & ResNet+STN  & Multiple Instances CE loss & RGB & D & underlying subtle motion features\\

   \midrule
LSCNN~\cite{de2018learning} & 2018 & 9 PatchNets & Binary CE loss & RGB & S & global feature via aggregating 9 deep local features \\

   \midrule
LiveNet~\cite{rehman2018livenet} & 2018 & VGG11 & Binary CE loss & RGB & S & model trained with continuous data-randomization  \\

   \midrule
MS-FANS~\cite{luo2018face} & 2018 & AlexNet+LSTM & Binary CE loss & RGB & S & multi-scale deep feature with rich spatial context \\

   \midrule
DeepColorFAS~\cite{larbi2018deepcolorfasd} & 2018 & 5-layer CNN  & Binary CE loss & \tabincell{c}{RGB,\\HSV,\\YCbCr} & S &  \tabincell{c}{investigates the effect of multi-channel space colors\\ on CNN architectures and proposes a fusion based\\ voting method for FAS} \\

   \midrule
Siamese~\cite{hao2019face} & 2019 & AlexNet & Contrastive loss & RGB & S & deep features guided by client identity information  \\ 

   \midrule
FSBuster~\cite{bresan2019facespoof} & 2019 & ResNet50 & Trained with SVM & RGB & S & fused deep features from Intrinsic Image Properties  \\

   \midrule
FuseDNG~\cite{rehman2019face} & 2019 & 7-layer CNN  & \tabincell{c}{Binary CE loss\\Reconstruction loss} & RGB & S & \tabincell{c}{adaptive fusion of deep features learned from\\ real-world face and deep autoencoder generated face }  \\ 
 
 \midrule
 STASN~\cite{yang2019face} & 2019 & ResNet50+LSTM & Binary CE loss & RGB & D & deep spatio-temporal feature from local salient regions\\

 \midrule
TSCNN~\cite{chen2019attention} & 2019 & ResNet18  & Binary CE loss & \tabincell{c}{RGB\\MSR} & S & \tabincell{c}{ attentional illumination-invariant features\\ with discriminative high-frequency information}\\

   \midrule
FAS-UCM~\cite{laurensi2019style} & 2019 & \tabincell{c}{MobileNetV2\\VGG19}  & \tabincell{c}{Binary CE loss\\Style loss} & RGB & S & \tabincell{c}{deep features trained from generated \\style transferred images}  \\

   \midrule
 SLRNN~\cite{muhammad2019face} & 2019 & ResNet50+LSTM  & Binary CE loss & RGB & D & augmented temporal features via sparse filtering\\ 

   \midrule
GFA-CNN~\cite{tu2020learning} & 2019 & VGG16  & Binary CE loss & RGB & S & generalizable features via multitask and metric learning  \\ 

   \midrule
3DSynthesis~\cite{8987415} & 2019 & ResNet15  & Binary CE loss & RGB & S & trained on synthesized virtual data of print attacks \\

 \midrule
 CompactNet~\cite{li2020compactnet} & 2020 & VGG19  & Points-to-Center triplet loss & RGB & S & deep features on the learned color-liked compact space\\ 

 \midrule
 SSR-FCN~\cite{deb2020look} & 2020 & FCN with 6 layers  & Binary CE loss & RGB & S & local discriminative features from Self-Regional Supervision\\

    \midrule
 DRL-FAS~\cite{cai2020drl} & 2020 & ResNet18+GRU & Binary CE loss & RGB & S & fused local(sub-patches) \& global(entire face) features\\ 
 
   \midrule
SfSNet~\cite{pinto2020leveraging} & 2020 & 6-layer CNN  & Binary CE loss & \tabincell{c}{Albedo,\\ Depth,\\ Reflect.} & S & \tabincell{c}{ intrinsic features from shape-from-shading generated\\ pseduo albedo, depth, and reflectance maps}\\
 
   \midrule
LivenesSlight~\cite{zuo2020face} & 2020 & 6-layer CNN  & Binary CE loss & RGB & S & lightweight model and takes less training time\\ 
 
   \midrule
\tabincell{c}{Motion-\\Enhancement}~\cite{ge2020face} & 2020 & VGGface+LSTM  & Binary CE loss & RGB & D & \tabincell{c}{deep temporal dynamics features with eulerian motion\\
magnification and temporal attention mechanism}  \\  
 
   \midrule
CFSA-FAS~\cite{chen2020face} & 2020 & ResNet18  & Binary CE loss & RGB & S & \tabincell{c}{fuse high and low frequency information with cross-\\frequency spatial and self-channel attention modules}  \\

   \midrule
MC-FBC~\cite{jia20203d} & 2020 & \tabincell{c}{VGG16\\ResNet50}  & Binary CE loss & RGB & S & \tabincell{c}{fine-grained features via factorizing bilinear\\ coding of multiple color channels}  \\

   \midrule
SimpleNet~\cite{parkin2020creating} & 2020 & \tabincell{c}{Multi-stream \\5-layer CNN}  & Binary CE loss & \tabincell{c}{RGB, \\OF, RP} & D & \tabincell{c}{using intermediate representations from RankPooling\\ and optical flow to increase model's robustness}  \\

   \midrule
PatchCNN~\cite{almeida2020detecting} & 2020 & SqueezeNet v1.1  & \tabincell{c}{Binary CE loss\\Triplet loss} & RGB & S & \tabincell{c}{trained with multi-resolution patches and \\a multi-objective loss function}  \\ 

   \midrule
\tabincell{c}{FreqSpatial-\\TempNet}~\cite{huang2020deep} & 2020 & ResNet18  & Binary CE loss & \tabincell{c}{RGB,\\HSV,\\Spectral} & D & \tabincell{c}{discriminative fused features of  frequent, \\spatial and temporal information}  \\

   \midrule
ViTranZFAS~\cite{george2020effectiveness} & 2021 & \tabincell{c}{Vision\\ Transformer}  & Binary CE loss & RGB & S & \tabincell{c}{transfer learning from the pre-trained \\vision transformer model }  \\  

   \midrule
CIFL~\cite{chen2021camera} & 2021 & \tabincell{c}{ResNet18}  & \tabincell{c}{Binary focal loss\\camear type loss} & RGB & S & \tabincell{c}{camera-invariant spoofing features in the high-\\frequency domain and enhanced image }  \\ 

  \midrule
FasTCo~\cite{xu2020improving} & 2021 & \tabincell{c}{ResNet50\\MobileNetV2}  & \tabincell{c}{Multi-class CE loss\\Temporal Consistency loss\\Class Consistency loss} & RGB & D & \tabincell{c}{ temporal consistent features as well as \\temporal smooothed predictions}\\

   \midrule
PatchNet~\cite{wang2022patchnet} & 2022 & \tabincell{c}{ResNet18}  & \tabincell{c}{Asymmetric AM-Softmax loss\\self-supervised similarity loss} & \tabincell{c}{RGB\\patch} & S & \tabincell{c}{fine-grained patch-type live/spoof recognition with\\ strong patch embedding space regularization}  \\

 \bottomrule[1pt]
 \end{tabular}}
\end{table*}

\begin{table*}
\centering
\caption{Summary of the representative \textbf{traditional deep learning} based FAS methods with \textbf{pixel-wise supervision}. Most methods (in the upper part) are supervised with \textit{auxiliary} tasks while the methods in the last eight rows are based on the \textit{generative} models. `S/D' is short for Static/Dynamic. `NAS' denotes neural searched architecture. `TSM' and `FPN' denote temporal shift module and feature pyramid network, respectively. `Info-VAE' means information maximizing variational autoencoder. Note that some methods also consider classification loss (e.g., binary cross entropy loss, triplet loss, and adversarial loss), which are not listed in the `Supervision' column.} \label{tab:auxiliary}
\resizebox{1.0\textwidth}{!} {\begin{tabular}{l c c c c c c} 
 \toprule[1pt]
 Method & Year & Supervision & Backbone & Input & S/D & Description \\
 \midrule
 Depth\&Patch~\cite{Atoum2018Face} & 2017 & Depth & \tabincell{c}{PatchNet\\DepthNet} & \tabincell{c}{YCbCr\\HSV} & S & \tabincell{c}{ local patch features and holistic\\ depth maps extracted by two-stream CNNs}\\

 \midrule
 Auxiliary~\cite{Liu2018Learning} & 2018 & \tabincell{c}{Depth\\rPPG spectrum} & DepthNet & \tabincell{c}{RGB\\HSV} & D & \tabincell{c}{local temporal features learned from CNN-RNN model \\with pixel-wise depth and sequence-wise rPPG supervision}\\

  \midrule
 BASN~\cite{kim2019basn} & 2019 & \tabincell{c}{Depth\\Reflection} & \tabincell{c}{DepthNet\\Enrichment} & \tabincell{c}{RGB\\HSV} & S & \tabincell{c}{generalizable features via bipartite auxiliary supervision}\\

  \midrule
DTN~\cite{liu2019deep} & 2019 & BinaryMask & Tree Network  & \tabincell{c}{RGB\\HSV} & S & \tabincell{c}{partition the spoof samples into semantic \\sub-groups in an unsupervised fashion}\\

  \midrule
 PixBiS~\cite{george2019deep} & 2019 & BinaryMask & DenseNet161  & RGB & S & \tabincell{c}{deep pixel-wise binary supervision without trivial depth synthesis}\\

 \midrule
 A-PixBiS~\cite{hossaindeeppixbis} & 2020 & BinaryMask & DenseNet161  & RGB & S & \tabincell{c}{incorporate a variant of binary cross entropy that  enforces\\ a margin in angular space for attentive pixel wise supervision}\\

 \midrule
Auto-FAS~\cite{yu2020auto2} & 2020 & BinaryMask  & NAS  & RGB & S & \tabincell{c}{well-suitable lightweight networks searched for mobile-level FAS}\\

 \midrule
MRCNN~\cite{ma2020novel} & 2020 &  BinaryMask  & Shallow CNN  & RGB & S & \tabincell{c}{introducing local losses to patches, and constraints the \\ entire face region to avoid over-emphasizing certain local areas}\\

 \midrule
FCN-LSA~\cite{sun2020face2} & 2020 &  BinaryMask  & DepthNet  & RGB & S & high frequent spoof cues from  lossless size adaptation module\\

 \midrule
CDCN~\cite{yu2020searching} & 2020 &  Depth  & DepthNet  & RGB & S & \tabincell{c}{intrinsic detailed patterns via aggregating both intensity and\\ gradient information from stacked central difference convolutions. }\\


\midrule
FAS-SGTD~\cite{wang2020deep} & 2020 &  Depth  &  \tabincell{c}{DepthNet\\STPM}  & RGB & D & \tabincell{c}{detailed discriminative dynamics cues from stacked Residual \\Spatial Gradient Block and Spatio-Temporal Propagation Module}\\

 \midrule
TS-FEN~\cite{peng2020ts} & 2020 &  Depth   & \tabincell{c}{ResNet34\\FCN}  & \tabincell{c}{RGB\\YCbCr\\HSV} & S & \tabincell{c}{discriminative fused features from \\depth-stream and chroma-stream networks}\\

 \midrule
SAPLC~\cite{sun2020face} & 2020 &  TernaryMap   & DepthNet  & \tabincell{c}{RGB\\HSV} & S & \tabincell{c}{accurate image-level decision via spatial aggregation of \\pixel-level local classifiers even with insufficient training samples}\\

 \midrule
BCN~\cite{yu2020face} & 2020 &  \tabincell{c}{BinaryMask\\Depth\\Reflection}   & DepthNet  & RGB & S & \tabincell{c}{intrinsic material-based patterns captured via \\aggregating
multi-level bilateral macro- and micro- information}\\

 \midrule
Disentangled~\cite{zhang2020face} & 2020 &  \tabincell{c}{Depth\\TextureMap}   & DepthNet  & RGB & S & \tabincell{c}{liveness and content features via disentangled representation learning}\\

 \midrule
AENet~\cite{zhang2020celeba} & 2020 &  \tabincell{c}{Depth\\Reflection}   & ResNet18  & RGB & S & \tabincell{c}{rich semantic features using Auxiliary Information,\\ Embedding Network with multi-task learning framework}\\

 \midrule
3DPC-Net~\cite{li3dpc} & 2020 &  3D Point Cloud   & ResNet18  & RGB & S & discriminative features via fine-grained 3D Point Cloud supervision\\

 \midrule
PS~\cite{yu2021revisiting} & 2020 &  \tabincell{c}{BinaryMask\\Depth}   & \tabincell{c}{ResNet50\\CDCN}  & RGB & S & \tabincell{c}{ pyramid supervision guides models to learn both local\\ details and global semantics from multi-scale spatial context} \\

 \midrule
NAS-FAS~\cite{yu2020fas2} & 2020 &  \tabincell{c}{BinaryMask\\Depth}   & NAS  & RGB & D & \tabincell{c}{leveraging cross-domain/type knowledge and static-dynamic\\ representation for central difference network search} \\

 \midrule
DAM~\cite{zheng2021attention} & 2021 &  \tabincell{c}{Depth}   & \tabincell{c}{VGG16\\ TSM}  & RGB & D & \tabincell{c}{attentional fused depth and multi-scale temporal clues using a \\two-stream network as well as a self-supervised symmetry loss} \\

 \midrule
Bi-FPNFAS~\cite{roy2021bi} & 2021 &  \tabincell{c}{Fourier spectra}   & \tabincell{c}{EfficientNetB0\\FPN}  & RGB & S & \tabincell{c}{multiscale bidirectional propagated features with\\ self-generated frequency spectra supervision} \\

 \midrule
DC-CDN~\cite{yu2021dual} & 2021 &  \tabincell{c}{Depth}   & CDCN  & RGB & S & \tabincell{c}{efficient feature learning on dual-cross central difference\\ network with Cross Feature Interaction Modules} \\

 \midrule[1pt]
  \midrule[1pt]
 De-Spoof~\cite{jourabloo2018face} & 2018 & \tabincell{c}{Depth\\BinaryMask\\FourierMap} & \tabincell{c}{DSNet\\DepthNet} & \tabincell{c}{RGB\\HSV} & S & \tabincell{c}{inversely decomposing a spoof face
into a spoof\\ noise and a live face, and estimating subtle\\ spoof noise with proper supervisions}\\

 \midrule
Reconstruction~\cite{chen2019towards} & 2019 &  \tabincell{c}{RGB Input (live)\\ZeroMap (spoof)}   & U-Net  & RGB & S & multi-level semantic features from autoencoder\\

 \midrule
LGSC~\cite{feng2020learning} & 2020 &  ZeroMap (live)   & \tabincell{c}{U-Net\\ResNet18}  & RGB & S & \tabincell{c}{discriminative live-spoof differences learned within a residual-\\learning framework with the perspective of anomaly detection }\\

 \midrule
\tabincell{c}{TAE}~\cite{mohammadi2020improving} & 2020  & \tabincell{c}{Binary CE loss\\Reconstruction loss} & \tabincell{c}{Info-VAE+\\DenseNet161 } & RGB & S &  \tabincell{c}{self-pretrained autoencoder in large-scale face recognition\\ datasets to obtain the reconstruction-error images for FAS}  \\

 \midrule
STDN~\cite{liu2020disentangling} & 2020 &  \tabincell{c}{BinaryMask\\RGB Input (live)}   & \tabincell{c}{U-Net\\PatchGAN}  & RGB & S & \tabincell{c}{disentangled spoof trace via adversarial learning and\\ hierarchical combination of patterns at multiple scales}\\

 \midrule
GOGen~\cite{stehouwer2020noise} & 2020 &  RGB input   & \tabincell{c}{DepthNet}  & \tabincell{c}{RGB+one-\\hot vector} & S &  \tabincell{c}{GAN-based architecture to synthesize and identify the\\ spoof noise patterns from medium/sensor combinations}\\

 \midrule
PhySTD~\cite{liu2020physics} & 2021 &  \tabincell{c}{Depth\\RGB Input (live)}   & \tabincell{c}{U-Net\\PatchGAN}  & \tabincell{c}{Frequency\\ Trace} & S & \tabincell{c}{disentangling spoof faces into the spoof traces and\\ live counterparts guided by physical properties} \\

 \midrule
MT-FAS~\cite{qin2021meta} & 2021 &  \tabincell{c}{ZeroMap (live)\\LearnableMap (Spoof)}   & DepthNet  & RGB & S & \tabincell{c}{ train a meta-teacher to generate optimal pixel-wise\\ signals for supervising the spoofing detector} \\

 \bottomrule[1pt]
 \end{tabular}}
\end{table*}

\begin{table*}
\centering
\caption{Summary of the representative \textbf{generalized deep learning} FAS methods to \textbf{unseen domain (domain adaptation and domain generalization)}. `MMD' is short for `Maximum Mean Discrepancy'.}
\label{tab:domain}
\resizebox{1.0\textwidth}{!} {\begin{tabular}{l| c c c c c c} 
 \toprule[1pt]
 & Method & Year & Backbone & Loss  & S/D & Description \\
 
\midrule

& \tabincell{c}{OR-DA}~\cite{li2018unsupervised} & 2018 & \tabincell{c}{AlexNet} & \tabincell{c}{Binary CE loss\\MMD loss} & S &  \tabincell{c}{learned classifier for target domain, and embedding space\\ with similar distribution for source and target domains}  \\

\cmidrule{2-7}
& \tabincell{c}{DTCNN}~\cite{tu2019deep} & 2019 & \tabincell{c}{AlexNet} & \tabincell{c}{Binary CE loss\\MMD loss} & S &  \tabincell{c}{domain invariant features using a few\\ labeled samples
from the target domain}  \\ 

\cmidrule{2-7}
 & \tabincell{c}{Adversarial}~\cite{wang2019improving} & 2019 & \tabincell{c}{ResNet18} & \tabincell{c}{Triplet loss\\Adversarial loss} & S &  \tabincell{c}{learn a shared embedding space by both source and\\ target domain models via adversarial domain adaptation }  \\ 

\cmidrule{2-7}
 & \tabincell{c}{ML-MMD}~\cite{zhou2019face} & 2019 & \tabincell{c}{Multi-scale\\ FCN} & \tabincell{c}{CE loss\\MMD loss} & S &  \tabincell{c}{adapt in both representation and classifier\\ layers to bridge for the domain discrepancy}  \\ 

\cmidrule{2-7}
& \tabincell{c}{OCA-FAS}~\cite{qin2020one} & 2020 & \tabincell{c}{DepthNet} & \tabincell{c}{Binary CE loss\\Pixel-wise binary loss} & \tabincell{c}{S} &  \tabincell{c}{train a meta-learner with loss function search on\\ one-class adaptation FAS tasks with only live samples}  \\ 

\cmidrule{2-7}
\tabincell{c}{Domain\\Adaptation}
& \tabincell{c}{DR-UDA}~\cite{wang2020unsupervised} & 2020 & \tabincell{c}{ResNet18} & \tabincell{c}{Center\&Triplet loss\\Adversarial loss\\Disentangled loss} & S &  \tabincell{c}{disentangles the features
irrelevant to specific\\ domains, and learn a shared embedding\\ space by both source and target domains}  \\

\cmidrule{2-7}
& \tabincell{c}{DGP}~\cite{mohammadi2020domain} & 2020 & \tabincell{c}{DenseNet161} & \tabincell{c}{Feature divergence measure \\BinaryMask} & S & \tabincell{c}{prune the filters specific to the source dataset\\ for performance improvement on target dataset}  \\

\cmidrule{2-7}
& \tabincell{c}{Distillation}~\cite{li2020face2} & 2020 & \tabincell{c}{AlexNet} & \tabincell{c}{Binary CE loss\\MMD loss\\Paired Similarity} & S &  \tabincell{c}{spoofing-specific information captured by distilled\\ deep network on the application-specific domain}  \\

\cmidrule{2-7}
& \tabincell{c}{S-CNN\\+PL+TC}~\cite{quan2021progressive} & 2021 & \tabincell{c}{ResNet18} & \tabincell{c}{CE Loss in labeled \\ and unlabeled sets} & D &  \tabincell{c}{semi-supervised learning framework with only a few\\ labeled training data, and progressively adopt the\\ unlabeled data with reliable pseudo labels.  }  \\

\cmidrule{2-7}
& \tabincell{c}{USDAN}~\cite{jia2021unified} & 2021 & \tabincell{c}{ResNet18} & \tabincell{c}{Adaptive binary CE loss\\Entropy loss\\Adversarial loss} & S &  \tabincell{c}{design different distribution alignment operations to\\ enhance generalization for un- \& semi-supervised\\ domain adaptation to address cross-scenario problem}  \\

\midrule[1pt]
& \tabincell{c}{MADDG}~\cite{shao2019multi} & 2019 & \tabincell{c}{DepthNet} & \tabincell{c}{Binary CE \& Depth loss\\Multi-adversarial loss\\Dual-force Triplet loss} & S &  \tabincell{c}{leverage the large variability present in FR datasets\\ to induce invariance to factors that cause domain-shift}  \\

\cmidrule{2-7}
& \tabincell{c}{PAD-GAN}~\cite{wang2020cross} & 2020 & \tabincell{c}{ResNet18} & \tabincell{c}{Binary CE \& GAN loss\\Reconstruction loss} & S &  \tabincell{c}{disentangled and domain-independent features rather\\ than subject discriminative and domain related features}  \\

\cmidrule{2-7}
& \tabincell{c}{SSDG}~\cite{jia2020single} & 2020 & \tabincell{c}{ResNet18} & \tabincell{c}{Binary CE loss\\Single-Side adversarial loss\\Asymmetric Triplet loss} & S &  \tabincell{c}{learn a generalized space where the feature distribution\\ of real faces is compact while that of fake ones is disper-\\sed among domains but compact within each domain}  \\ 

\cmidrule{2-7}
& \tabincell{c}{RF-Meta}~\cite{shao2019regularized} & 2020 & \tabincell{c}{DepthNet} & \tabincell{c}{Binary CE loss\\Depth loss} & S &  \tabincell{c}{meta-learned generalized features across multiple\\ source domains with auxiliary regularization}  \\ 

\cmidrule{2-7}
& \tabincell{c}{CCDD}~\cite{saha2020domain} & 2020 & \tabincell{c}{ResNet50\\+LSTM} & \tabincell{c}{Binary CE loss\\Class-conditional loss} & D &  \tabincell{c}{learn discriminative but domain-robust features with\\ class-conditional domain discriminator module and GRL}  \\

\cmidrule{2-7}
& \tabincell{c}{DASN}~\cite{kim2020suppressing} & 2021 & \tabincell{c}{ResNet18} & \tabincell{c}{Binary CE \& Spoof-\\irrelevant factor loss} & S &  \tabincell{c}{adopt doubly adversarial learning to suppress the\\ spoof-irrelevant
factors, and intensify spoof factors.}  \\

\cmidrule{2-7}
\tabincell{c}{Domain\\ Generalization}
& \tabincell{c}{SDA}~\cite{wang2021self} & 2021 & \tabincell{c}{DepthNet} & \tabincell{c}{Binary CE \& Depth loss\\Reconstruction loss\\Orthogonality regularization} & S &  \tabincell{c}{use meta-learning based adaptor learning for better\\ adaptor initialization, and an unsupervised adaptor\\ loss for appropriate adaptor optimization}  \\

\cmidrule{2-7}
& \tabincell{c}{D$^{2}$AM}~\cite{chen2021generalized} & 2021 & \tabincell{c}{DepthNet} & \tabincell{c}{Binary CE loss\\ Depth loss\\MMD loss} & S &  \tabincell{c}{iteratively divide mixture domains via discriminative\\ domain representation and train generalizable models\\ with meta-learning without using domain labels}  \\ 

\cmidrule{2-7}
& \tabincell{c}{DRDG~\cite{liu2021dual}} & 2021 & \tabincell{c}{DepthNet} & \tabincell{c}{Binary CE \& Depth loss\\domain loss} & S &  \tabincell{c}{iteratively
reweight the relative importance between samples\\ and features to extract domain-irrelevant features}  \\ 

\cmidrule{2-7}
& \tabincell{c}{ANRL~\cite{liu2021adaptive}} & 2021 & \tabincell{c}{DepthNet} & \tabincell{c}{Binary CE \& Depth loss\\inter-domain compatible loss\\inter-class separable loss} & S &  \tabincell{c}{adaptively select feature normalization methods to learn\\ domain-agnostic and discriminative representation}   \\

\cmidrule{2-7}
& \tabincell{c}{FGHV~\cite{wang2022domain}} & 2022 & \tabincell{c}{DepthNet} & \tabincell{c}{Variance, relative correlation\\ distribution discrimination constraints} & S &  \tabincell{c}{feature generation networks generate hypotheses of real\\ faces and known attacks, and two hypothesis verification\\ modules are applied to judge real/generative distributions}  \\ 

\cmidrule{2-7}
& \tabincell{c}{SSAN~\cite{wang2022domain}} & 2022 & \tabincell{c}{DepthNet\\ResNet} & \tabincell{c}{Binary CE loss\\ domain adversarial loss\\contrastive loss} &S &  \tabincell{c}{extract and reassemble different content and style features\\ for a stylized feature space, and emphasize liveness-related \\ style information while suppress the domain-specific one }  \\



 \bottomrule[1pt]
 \end{tabular}}
\end{table*}

\begin{table*}
\centering
\caption{Summary of the \textbf{generalized deep learning} FAS methods to \textbf{unknown attack types (zero/few-shot learning and anomaly detection)}. `OCSVM', `MD', `GMM', and `OCCL' are short for `One-Class Support Vector Machine', `Mahalanobis-distance', `Gaussian Mixture Model', and `One-Class Contrastive Loss', respectively.} \label{tab:type}
\resizebox{1.0\textwidth}{!} {\begin{tabular}{l| c c c c c c} 
 \toprule[1pt]
 & Method & Year & Backbone & Loss  & Input & Description \\
 
\midrule

& \tabincell{c}{DTN}~\cite{liu2019deep} & 2019 & \tabincell{c}{Deep Tree\\Network} & \tabincell{c}{Binary CE loss\\Pixel-wise binary loss\\Unsupervised Tree loss} & \tabincell{c}{RGB\\HSV} &  \tabincell{c}{adaptively routing the attacks to the most similar\\ spoof cluster, and makes the binary decision}  \\ 

\cmidrule{2-7}
\tabincell{c}{Zero/Few-\\Shot}
& \tabincell{c}{AIM-FAS}~\cite{qin2019learning} & 2020 & \tabincell{c}{DepthNet} & \tabincell{c}{Depth loss\\Contrastive Depth loss} & RGB &  \tabincell{c}{adaptive inner-updated meta features generalized \\ to unseen spoof types from predefined PAs}  \\

\cmidrule{2-7}
& \tabincell{c}{CM-PAD}~\cite{perez2020learning} & 2021 & \tabincell{c}{DepthNet\\ResNet} & \tabincell{c}{Binary CE loss\\Depth loss\\Gradient alignment} & RGB &  \tabincell{c}{continual meta-learning PAD solution that\\ can be trained on unseen attack scenarios\\ catastrophic seen attack
forgetting }  \\

\midrule[1pt]
& \tabincell{c}{AE+LBP}~\cite{xiong2018unknown} & 2018 & \tabincell{c}{AutoEncoder} & \tabincell{c}{Reconstruction loss} & RGB &  \tabincell{c}{embedding features (cascaded with LBP) from outlier\\ detection based neural network autoencoder}  \\

\cmidrule{2-7}
& \tabincell{c}{Anomaly}~\cite{perez2019deep} & 2019 & \tabincell{c}{ResNet50} & \tabincell{c}{Triplet focal loss\\Metric-Softmax loss} & \tabincell{c}{RGB} &  \tabincell{c}{deep anomaly detection via introducing a\\ few-shot posteriori probability estimation}  \\

\cmidrule{2-7}
& \tabincell{c}{Anomaly2}~\cite{fatemifar2019spoofing} & 2019 & \tabincell{c}{GoogLeNet\\ResNet50} & \tabincell{c}{MD} & \tabincell{c}{RGB} &  \tabincell{c}{subject specific anomaly detector is trained on\\ genuine accesses only using one-class classifiers }  \\ 

\cmidrule{2-7}
& \tabincell{c}{Hypersphere}~\cite{li2020unseen} & 2020 & \tabincell{c}{ResNet18} & \tabincell{c}{Hypersphere loss} & \tabincell{c}{RGB\\HSV} &  \tabincell{c}{deep anomaly detection supervised by hypersphere\\ loss, and detects PAs directly on learned feature space}  \\ 


\cmidrule{2-7}
\tabincell{c}{Anomaly-\\Detection}& \tabincell{c}{Ensemble-\\Anomaly}~\cite{fatemifar2020stacking} & 2020 & \tabincell{c}{GoogLeNet\\ResNet50} & \tabincell{c}{GMM \\ (not end-to-end)} & \tabincell{c}{RGB\\ patches} &  \tabincell{c}{ensemble of one-class classifiers from different\\ facial regions, CNNs, and anomaly detectors}  \\

\cmidrule{2-7}
& \tabincell{c}{MCCNN}~\cite{george2020learning} & 2020 & \tabincell{c}{LightCNN} & \tabincell{c}{Binary CE loss\\Contrastive loss} & \tabincell{c}{Grayscale,\\ IR, Depth,\\ Thermal } &  \tabincell{c}{learn a compact embedding for bonafide while\\ being far from the representation of PAs via\\ OCCL, and cascaded with a one-class GMM}  \\

\cmidrule{2-7}
& \tabincell{c}{End2End-\\Anomaly}~\cite{baweja2020anomaly} & 2020 & \tabincell{c}{VGG-Face} & \tabincell{c}{Binary CE loss\\Pairwise confusion} & \tabincell{c}{RGB} &  \tabincell{c}{both classifier and representations are learned\\ end-to-end with pseudo negative class}  \\

\cmidrule{2-7}
& \tabincell{c}{ClientAnomaly}~\cite{fatemifar2020client} & 2020 & \tabincell{c}{ResNet50\\GoogLeNet\\VGG16} & \tabincell{c}{OCSVM\\GMM\\MD} & \tabincell{c}{RGB} &  \tabincell{c}{client-specific knowledge are leveraged for\\ anomaly-based spoofing detectors as well as\\ determination thresholds}  \\

 \bottomrule[1pt]
 \end{tabular}}
\end{table*}

\begin{table*}
\centering
\caption{Summary of the representative \textbf{deep learning} FAS methods with \textbf{specialized sensor/hardware inputs}. `S/D', `SD', `AD', `FM', `APD', `LFC', 'DP' and `DOLP' are short for `Static/Dynamic', `Square Disparity', `Absolute Disparity', `Feature Multiplication', `Approximate Disparity', `Light Field Camera', `Dual Pixel' and `Degree of Linear Polarization', respectively.} \label{tab:sensors}
\resizebox{1.0\textwidth}{!} {\begin{tabular}{l c c c c c c} 
 \toprule[1pt]
 Method & Year & Backbone & Loss & Input & S/D & Description \\

   \midrule
\tabincell{c}{Thermal-\\FaceCNN}~\cite{seo2019face} & 2019 & \tabincell{c}{AlexNet} & Regression loss & \tabincell{c}{ Thermal infrared\\ face image} & S &  \tabincell{c}{temperature related features based on the fact that\\ real face temperature is 36$\sim$37 degrees on average}  \\ 


   \midrule
\tabincell{c}{SLNet}~\cite{rehman2020slnet} & 2019 & \tabincell{c}{17-layer CNN} & Binary CE loss & \tabincell{c}{Stereo (left\&right)\\ face images} & S &  \tabincell{c}{disparities between deep features are\\ learned using SD, AD, FM, and APD operations}  \\

   \midrule
\tabincell{c}{Aurora\\Guard}~\cite{liu2019AuroraGuard} & 2019 & U-Net & \tabincell{c}{Binary CE loss\\Depth regression\\Light Regression} & \tabincell{c}{Casted face with dynamic \\changing light specified by\\ random light CAPTCHA} & D &  \tabincell{c}{based on the normal cues extracted from\\ the light reflection, multi-tasck CNN recovers\\ both subjects' depth maps and light CAPTCHA }  \\

   \midrule
LFC~\cite{liu2019light} & 2019 & AlexNet & Binary CE loss & \tabincell{c}{Ray difference/microlens\\ images from LFC} & S &  \tabincell{c}{meaningful features extracted from single-shot\\ LFC images with rich depth information of objects}  \\  

   \midrule
PAAS~\cite{tian2020face} & 2020 & MobileNetV2 & \tabincell{c}{Contrastive loss\\SVM} & \tabincell{c}{Four-directional polarized\\ face image} & S &  \tabincell{c}{learned discriminative and robust features from DOLP\\ as polarization reveals the intrinsic attributes}  \\  

   \midrule
\tabincell{c}{Face-\\Revelio}~\cite{farrukh2020facerevelio} & 2020 & \tabincell{c}{Siamese-\\AlexNet} & L1 distance & \tabincell{c}{Four flash lights displayed\\ on four quarters of a screen} & D &  \tabincell{c}{ varying illumination enables the recovery\\ of the face surface normals via photometric
stereo}  \\

   \midrule
\tabincell{c}{SpecDiff}~\cite{ebihara2019specular} & 2020 & \tabincell{c}{ResNet4} & Binary CE loss &  \tabincell{c}{Concatenated face images\\ w/ and w/o flash} & S &  \tabincell{c}{a novel descriptor based on specular and diffuse\\ reflections, with a flash-based deep FAS baseline}  \\

   \midrule
\tabincell{c}{MC-\\PixBiS}~\cite{heusch2020deep} & 2020 & \tabincell{c}{DenseNet161} & Binary mask loss &  SWIR images differences & S &  \tabincell{c}{discriminative features for Impersonation attacks as \\water is very absorbing in some SWIR wavelengths}  \\

   \midrule
\tabincell{c}{Thermal-\\ization}~\cite{kowalski2020study} & 2020 & \tabincell{c}{YOLO V3+\\ GoogLeNet} & Binary CE loss & \tabincell{c}{ Thermal infrared\\ face image} & S &  \tabincell{c}{learned specific physical features \\from Thermal infrared imaging of PAs}  \\ 

   \midrule
\tabincell{c}{DP Bin-\\-Cls-Net}~\cite{wu2020single} & 2021 & \tabincell{c}{Shallow U-Net\\ + Xception} & \tabincell{c}{Transformation consistency\\Relative disparity loss\\ Binary CE loss} & \tabincell{c}{ DP image pair} & S &  \tabincell{c}{reconstructed depth based on the DP pair with\\ self-supervised loss for planar attack detection}  \\ 

 \bottomrule[1pt]
 \end{tabular}}
\end{table*}

\begin{table*}
\centering
\caption{Summary of the \textbf{multi-modal deep learning} FAS methods. `MFEM', `SPM', `LFV' and `MLP' are short for `Modal Feature Erasing Module', `Selective Modal Pipeline', `Limited Frame Vote' and `Multilayer Perceptron', respectively.} \label{tab:multimodal}
\resizebox{1.0\textwidth}{!} {\begin{tabular}{l c c c c c c} 
 \toprule[1pt]
 Method & Year & Backbone & Loss & Input & Fusion    & Description \\

   \midrule
FaceBagNet~\cite{shen2019facebagnet} & 2019 & \tabincell{c}{Multi-stream\\CNN} & Binary CE loss & \tabincell{c}{RGB, Depth, NIR\\ face patches} & \tabincell{c}{Feature-\\level} &  \tabincell{c}{spoof-specific features from patch CNN, and\\ MFEM to prevent overfitting and better fusion}  \\  

   \midrule
FeatherNets~\cite{zhang2019feathernets} & 2019 & \tabincell{c}{Ensemble-\\FeatherNet} & Binary CE loss & \tabincell{c}{Depth,
NIR} & \tabincell{c}{Decision-\\level} &  \tabincell{c}{single compact FeatherNet trained by depth image,\\ then fused with “ensemble + cascade” structure}  \\  

   \midrule
Attention~\cite{wang2019multi} & 2019 & ResNet18 & \tabincell{c}{Binary CE loss\\Center loss} & \tabincell{c}{RGB, Depth, NIR} & \tabincell{c}{Feature-\\level} &  \tabincell{c}{using channel and spatial attention
module\\ to refine the multimodal features}  \\  

   \midrule
mmfCNN~\cite{kuang2019multi} & 2019 & ResNet34  & \tabincell{c}{Binary CE loss\\Binary Center Loss} & \tabincell{c}{RGB, NIR, Depth,\\ HSV, YCbCr} & \tabincell{c}{Feature-\\level} &  \tabincell{c}{fuses multi-level features among modalities in\\ a unified framework with weight-adaptation}  \\

   \midrule
MM-FAS~\cite{parkin2019recognizing} & 2019 & ResNet18/50  & \tabincell{c}{Binary CE loss} & \tabincell{c}{RGB, NIR, Depth} & \tabincell{c}{Feature-\\level} &  \tabincell{c}{leverages multimodal data and aggregates intra-
\\channel features at multiple network layers}  \\

   \midrule
AEs+MLP~\cite{nikisins2019domain} & 2019 & \tabincell{c}{Autoencoder\\MLP}  & \tabincell{c}{Binary CE loss\\Reconstruction loss} & \tabincell{c}{Grayscale-Depth-\\Infrared composition} & \tabincell{c}{Input-\\level} &  \tabincell{c}{trasfer learning within facial patches from the\\ facial RGB appearance to multi-channel modalities }  \\

   \midrule
SD-Net~\cite{zhang2020casia} & 2019 & ResNet18  & \tabincell{c}{Binary CE loss} & \tabincell{c}{RGB, NIR,\\ Depth} & \tabincell{c}{Feature-\\level} &  \tabincell{c}{multimodal fusion via feature re-weighting to\\ select more informative channels for modalities}  \\  

   \midrule
Dual-modal~\cite{li2019dual} & 2019 & MoblienetV3  & \tabincell{c}{Binary CE loss} & \tabincell{c}{RGB, IR} & \tabincell{c}{Feature-\\level} &  \tabincell{c}{light-weight networks to extract and merge\\ embedding features from NIR-VIS image pairs}  \\

   \midrule
\tabincell{c}{Parallel-CNN}~\cite{li2020face} & 2020 & \tabincell{c}{Attentional-\\CNN} & Binary CE loss & \tabincell{c}{Depth, NIR} & \tabincell{c}{Feature-\\level} &  \tabincell{c}{fused deep depth and IR features from paralleled\\ attentional CNN with spatial pyramid pooling}  \\  

   \midrule
\tabincell{c}{Multi-Channel\\ Detector}~\cite{george2020can} & 2020 & \tabincell{c}{RetinaNet\\(FPN+ResNet18)} & \tabincell{c}{Landmark regression\\Focal loss} & \tabincell{c}{Grayscale-Depth-\\Infrared composition} & \tabincell{c}{Input-\\level} &  \tabincell{c}{learned joint face detection-based and PAD-based\\ representation from fused 3 channel images}  \\

   \midrule
PSMM-Net~\cite{li2020casia} & 2020 & ResNet18  & \tabincell{c}{Binary CE loss\\ for each stream} & \tabincell{c}{RGB, Depth, NIR} & \tabincell{c}{Feature-\\level} &  \tabincell{c}{static-dynamic fusion mechanism with part-\\ially shared fusion strategy is proposed}  \\ 

   \midrule
PipeNet~\cite{yang2020pipenet} & 2020 & SENet154 & \tabincell{c}{Binary CE loss} & \tabincell{c}{RGB, Depth, NIR\\
 face patches} & \tabincell{c}{Feature-\\level} &  \tabincell{c}{SMP takes full advantage of multi-modal data.\\ LFV ensures stable video-level prediction}  \\  

   \midrule
MM-CDCN~\cite{yu2020multi} & 2020 & CDCN  & \tabincell{c}{Pixel-wise binary\\ loss, Contrastive\\ depth loss} & \tabincell{c}{RGB, Depth, NIR} & \tabincell{c}{Feature\&\\Decision\\level} &  \tabincell{c}{capture central-difference-based intrinsic\\ spoofing patterns among three modalities}  \\

   \midrule
HGCNN~\cite{te2020exploring} & 2020 & \tabincell{c}{Hypergraph-
\\CNN + MLP}  & \tabincell{c}{Binary CE loss} & \tabincell{c}{RGB, Depth} & \tabincell{c}{Feature-\\level} &  \tabincell{c}{auxiliary depth fused with texture in the feature\\ domain from hypergraph convolution}  \\

   \midrule
MCT-GAN~\cite{jiang2020face} & 2020 & \tabincell{c}{CycleGAN
\\ResNet50}  & \tabincell{c}{GAN loss\\Binary CE loss} & \tabincell{c}{RGB, NIR} & \tabincell{c}{Input-\\level} &  \tabincell{c}{generate NIR counterpart for VIS inputs\\ via GAN, and learn fusing features}  \\ 

   \midrule
D-M-Net~\cite{liu2021data} & 2021 & \tabincell{c}{ResNeXt }  & \tabincell{c}{Binary CE loss} & \tabincell{c}{Multi-preprocessed\\ Depth,  RGB-NIR\\ composition} & \tabincell{c}{Input\&\\Feature-\\level} &  \tabincell{c}{two-stage cascade architecture to fuse depth features\\ with multi-scale RGB-NIR composite features}  \\

   \midrule
CMFL~\cite{george2021cross} & 2021 & \tabincell{c}{DenseNet161 }  & \tabincell{c}{Cross modal focal loss\\Binary CE loss} & \tabincell{c}{RGB, Depth} & \tabincell{c}{Feature-\\level} &  \tabincell{c}{modulate the loss contribution and  comple-\\mentary information from the two modalities}  \\

   \midrule
MA-Net~\cite{liu2021face} & 2021 & \tabincell{c}{CycleGAN\\ResNet18 }  & \tabincell{c}{GAN loss\\Binary CE loss} & \tabincell{c}{RGB, NIR} & \tabincell{c}{Feature-\\level} &  \tabincell{c}{translate the visible inputs into NIR \\images, and then extract VIS-NIR features}  \\ 

   \midrule
\tabincell{c}{FlexModal\\-FAS~\cite{yu2022flexible}} & 2022 & \tabincell{c}{CDCN\\ ResNet50, ViT }  & \tabincell{c}{Binary CE loss\\Pixel-wise binary loss} & \tabincell{c}{RGB, Depth, NIR} & \tabincell{c}{Feature-\\level} &  \tabincell{c}{cross-attention fusion to efficiently mine\\ cross-modal clues for flexible-modal deployment}  \\ 


 \bottomrule[1pt]
 \end{tabular}}
\end{table*}

\end{document}